\RequirePackage[T1]{fontenc}
\documentclass[lettersize,journal]{IEEEtran}
\usepackage{amsmath,amsfonts}
\usepackage{array}
\usepackage[caption=false,font=normalsize,labelfont=sf,textfont=sf]{subfig}
\usepackage{textcomp}
\usepackage{stfloats}

\usepackage{url}
\usepackage{verbatim}
\usepackage{graphicx}
\usepackage{cite}
\usepackage[dvipsnames]{xcolor}
\usepackage{wrapfig}
\usepackage{booktabs}
\usepackage{iftex}
\ifPDFTeX\else
  \newcount\pdfcompresslevel
  \newcount\pdfoptionpdfminorversion
  \newcount\pdfgentounicode
  \providecommand{\pdfcatalog}[1]{}
  \providecommand{\pdfglyphtounicode}[2]{}
\fi

\usepackage[accsupp]{axessibility}
\usepackage{pifont}
\usepackage{float}
\usepackage{algorithm}
\usepackage{multirow}
\usepackage{makecell,colortbl}
\usepackage{algorithmicx}
\usepackage{algpseudocode}
\usepackage{placeins}
\usepackage{etoc}
\usepackage[most]{tcolorbox}
\usepackage{caption}
\usepackage{xcolor}

\definecolor{ForestGreen}{RGB}{34,139,34}

\usepackage{hyperref}
\usepackage{cleveref}

\crefname{figure}{Fig.}{Figs.}
\Crefname{figure}{Fig.}{Figs.}

\crefname{section}{Section}{Sections}
\Crefname{section}{Section}{Sections}

\crefname{equation}{Eq.}{Eqs.}
\Crefname{equation}{Eq.}{Eqs.}

\crefname{table}{Table}{Tables}
\Crefname{table}{Table}{Tables}

\PassOptionsToPackage{no-math}{fontspec}
\usepackage[
  enable,
  labname={CMLab},
  leftlogo={assets/cmlab_logo.png},
  logo={assets/princeton.png},
  titleicon={assets/icon.png},
  logowidth={40mm},
  rightlogowidth={30mm},
  projectpage={https://cmlab-korea.github.io/CHIMERA/}
]{cmlab}

\normalfont

\begin{document}

\title{%
  CHIMERA: Adaptive Cache Injection and Semantic Anchor Prompting\\
  for Zero-shot Image Morphing with Morphing-oriented Metrics%
}
\author{Dahyeon~Kye$^{*}$,~Jeahun~Sung$^{*}$,~Minkyu~Jeon,~and~Jihyong~Oh$^{\dagger}$%
\thanks{$^{*}$ Equal contribution.}%
\thanks{$^{\dagger}$ Corresponding author.}%
\thanks{D. Kye, J. Sung, and J. Oh are with CMLab, Chung-Ang University (e-mail: rpekgus@cau.ac.kr; jhseong@cau.ac.kr; jihyongoh@cau.ac.kr).}%
\thanks{M. Jeon is with Princeton University (e-mail: mj7341@princeton.edu).}%
}

\markboth{Journal of \LaTeX\ Class Files,~Vol.~14, No.~8, August~2021}%
{Shell \MakeLowercase{\textit{et al.}}: A Sample Article Using IEEEtran.cls for IEEE Journals}

\IEEEpubid{0000--0000/00\$00.00~\copyright~2021 IEEE}

\cmlabAuthors{Dahyeon Kye$^{*,1}$ \qquad Jeahun Sung$^{*,1}$ \qquad Minkyu Jeon$^{2}$ \qquad Jihyong Oh$^{\dagger,1}$}
\cmlabAffiliations{$^{1}$ CMLab, Chung-Ang University \qquad $^{2}$ Princeton University}
\cmlabAuthorEmail{\{rpekgus,jhseong,jihyongoh\}@cau.ac.kr \qquad mj7341@princeton.edu}
\cmlabProjectPage{https://cmlab-korea.github.io/CHIMERA/}

{%
  \hypersetup{hidelinks}%
  \maketitle
}
{%
  \renewcommand{\thefootnote}{}%
  \footnotetext{$^{*}$ Equal contribution.}%
  \footnotetext{$^{\dagger}$ Corresponding author.}%
}

\begin{abstract}
Recent diffusion-based image morphing methods typically interpolate inverted latents and reuse limited conditioning signals, often yielding unstable intermediates for heterogeneous endpoint pairs. In particular, (i) feature reuse is usually partial or non-adaptive, leading to abrupt structural changes or over-smoothing, and (ii) text conditions are commonly obtained independently per endpoint and then interpolated, which can introduce incompatible semantics. We present CHIMERA, a novel zero-shot diffusion morphing framework that addresses both issues via inversion-guided denoising with complementary feature reuse and text conditioning. 
Adaptive Cache Injection (ACI) caches a broader set of multi-scale diffusion features beyond Key--Value-only reuse during DDIM inversion and re-injects them with layer- and timestep-aware scheduling to stabilize denoising and enable gradual fusion.
Semantic Anchor Prompting (SAP) uses a VLM to generate a shared anchor-prompt and anchor-conditioned endpoint prompts and injects the anchor into cross-attention to improve intermediate semantic coherence. Finally, we propose the Global-Local Consistency Score (GLCS), a morphing-oriented metric that jointly captures global domain harmonization and local transition smoothness. Extensive experiments and a user study show that CHIMERA produces smoother and more semantically consistent morphing results than prior methods while remaining efficient and applicable across diverse diffusion backbones without retraining.
\end{abstract}

{\footnotesize This work has been submitted to the IEEE for possible publication. Copyright may be transferred without notice, after which this version may no longer be accessible.\par}
\section{Introduction}
\label{sec:intro}

\IEEEPARstart{I}{mage} morphing generates perceptually and semantically smooth transitions between two images. It is widely used in animation, film, and design, where intermediates must remain visually plausible and semantically meaningful~\cite{beier2023feature, liao2014automating, wolberg1998image}. Classical pipelines typically rely on handcrafted correspondences or optical-flow-based warping~\cite{wolberg1998image, beier2023feature, liao2014automating, brox2004high, darabi2012image}, which often fail for input pairs with large structural or domain gaps. Recent extensions to 3D morphing~\cite{liu2026interp3d, sun2026morphany3d, eisenberger2021neuromorph} further highlight the growing importance of robust morphing across modalities.

Recent diffusion-based morphing~\cite{wang2023infodiffusion, wang2023interpolating, yang2024impus, zhang2024diffmorpher, cao2025freemorph} improves fidelity via latent-space interpolation of pre-trained diffusion models~\cite{rombach2022high, song2020denoising} without explicit correspondence estimation. However, they remain unstable for heterogeneous or cross-domain inputs (e.g., photographs vs.~illustrations), often exhibiting structural inconsistency, semantic drift, or over-saturated artifacts. We argue that a successful morphing should

\begin{figure}[!t]
    \centering
    \includegraphics[width=\linewidth]{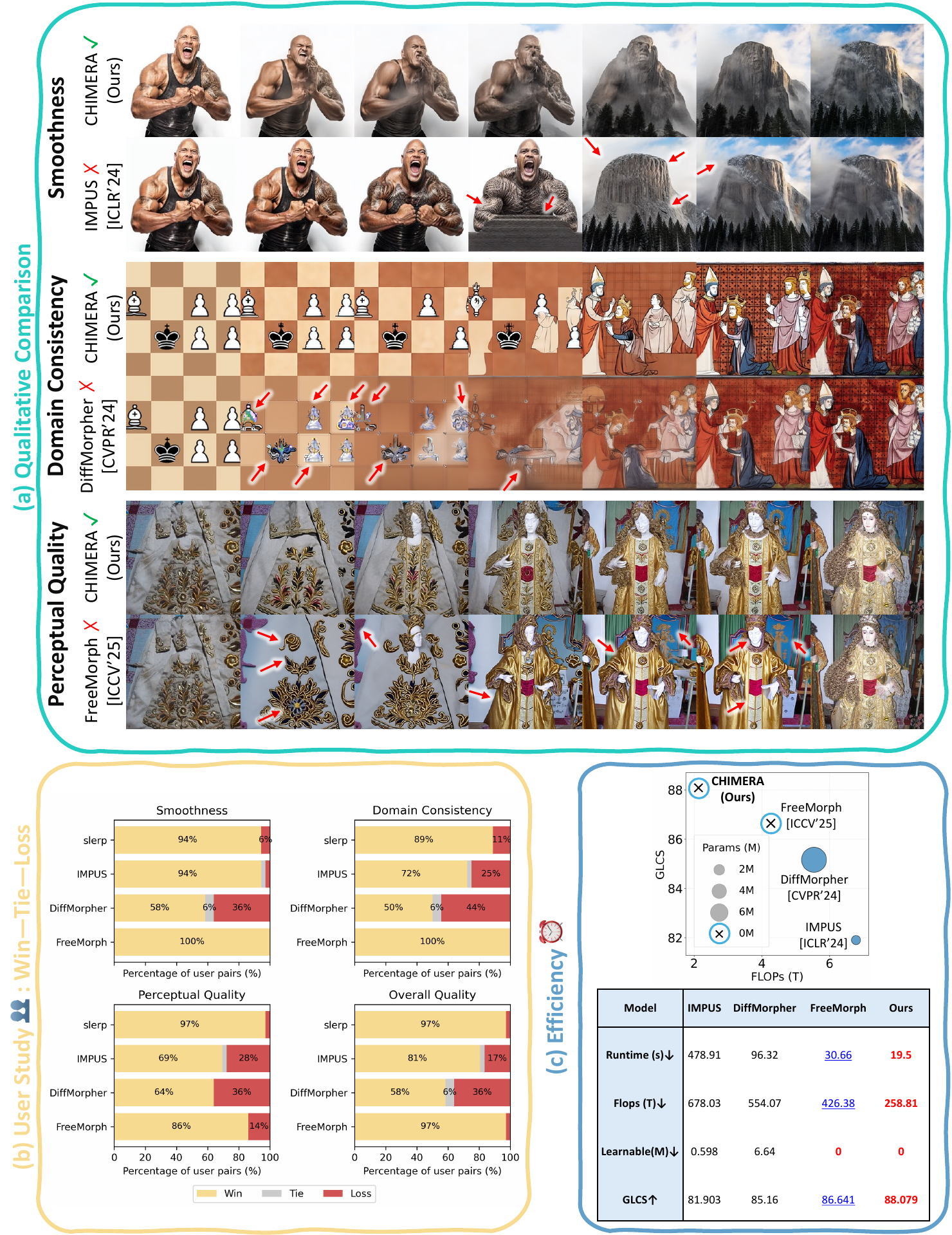}
    \caption{Given two input images, CHIMERA generates perceptually smooth transitions while jointly satisfying three key criteria: Smoothness, Domain Consistency, and Perceptual Quality (left; red arrows highlight failures of prior methods). Our user study further shows that CHIMERA is consistently preferred over baselines across all criteria, and this preference trend aligns with our morphing-oriented metric, GLCS. Finally, CHIMERA is training-free and efficient, achieving favorable runtime and computational cost compared to tuning-based approaches (right).}
    \label{fig:teaser1}
\end{figure}
\IEEEpubidadjcol

\noindent satisfy three criteria simultaneously: (i) \textbf{Smoothness}, (ii) \textbf{Domain Consistency}, and (iii) \textbf{Perceptual Quality}. As shown in \cref{fig:teaser1}, prior methods typically satisfy only a subset.


A key limitation of prior methods is insufficient transfer of input pair information to the denoising process. Tuning-based methods (e.g., IMPUS~\cite{yang2024impus}) improve semantics but incur substantial per-pair optimization cost, while diffusion feature reuse methods (e.g., DiffMorpher~\cite{zhang2024diffmorpher} and FreeMorph~\cite{cao2025freemorph}) primarily rely on Key-Value features from self-attention layers. We show that this KV-only reuse is a key contributing factor in their failure to satisfy both smoothness and domain consistency simultaneously. To support this, we analyze the similarity between diffusion features of the input pair and those of generated morphing images (\cref{Sec:PreNObs}), and observe that reusing only specific features is insufficient for stable morphing.

To preserve the efficiency of training-free morphing while overcoming these limitations, we propose \textbf{CHIMERA} (\textbf{C}ac\textbf{H}e \textbf{I}njection and Se\textbf{M}antic Anchor Prompting for Z\textbf{ER}o-shot Im\textbf{A}ge Morphing with Morphing-oriented Metrics), a novel zero-shot diffusion-based image morphing framework. CHIMERA is built on two complementary components: \textbf{Adaptive Cache Injection (ACI)} and \textbf{Semantic Anchor Prompting (SAP)}.

ACI improves feature reuse by leveraging multi-scale diffusion features beyond the KV-only used in prior methods~\cite{zhang2024diffmorpher,cao2025freemorph}.
Rather than naively injecting all cached features, ACI performs frequency-aware layer-timestep alignment via \textbf{Layer- and Timestep-wise Frequency Matching (LTM)}, enabling appropriate feature reuse at appropriate denoising steps. This design stabilizes structure and appearance transitions, improving domain consistency and smoothness while preserving fidelity.

Morphing remains difficult when the input pair shares little semantic or structural overlap. To address this, SAP uses a VLM~\cite{bai2025qwen2, liu2023visual} to identify the shared semantic or structural intersection (i.e., an \emph{anchor}) of the input pair and to construct anchor-correlated prompts. In contrast to prior methods that generate endpoint prompts independently, SAP constructs morphing-aware prompts, improving semantic coherence and reducing ambiguity in text conditioning.

Finally, we extend our analysis method (\cref{fig:observation}) to define a morphing-oriented metric, \textbf{Global-Local Consistency Score (GLCS)}, which jointly quantifies global domain harmonization and local transition smoothness. Unlike conventional metrics that capture only partial aspects of morphing quality, GLCS offers a unified measure that better reflects perceptual morphing quality.
Our main contributions are summarized as follows:
\vspace{-0.1cm}
\begin{itemize}
    \item[$\bullet$] We propose \textbf{CHIMERA}, a novel zero-shot diffusion morphing framework combining adaptive inversion-cache reuse and semantic anchor prompting for smooth and semantically coherent transitions.
    \item[$\bullet$] We propose \textbf{Adaptive Cache Injection (ACI)}, a layer- and timestep-adaptive diffusion feature reuse strategy that enables stable sampling and gradual, coherent fusion of structure and appearance.
    \item[$\bullet$] We propose \textbf{Semantic Anchor Prompting (SAP)}, which derives anchor-correlated prompts from a VLM to improve intermediate coherence and reduce semantic drift.
    \item[$\bullet$] We show that CHIMERA is training-free, efficient, and backbone-agnostic across diffusion models, while remaining competitive with tuning-based methods.
    \item[$\bullet$] We propose \textbf{Global-Local Consistency Score (GLCS)}, a morphing-oriented metric that jointly measures global domain harmonization and local transition smoothness, and better aligns with human preference than conventional metrics in our user study.
\end{itemize}
\section{Related Work}
\label{sec:rel}

\subsection{Image Morphing}
Image morphing is a long-standing problem in computer vision and graphics\cite{wolberg1998image, lee1996image, rajkovic2023geodesics, liao2014automating}, aiming for perceptually smooth transitions between two images.
Early approaches based on geometric correspondences (e.g., feature-line interpolation or flow-based warping)~\cite{brox2004high, darabi2012image, liao2014automating} often break when the input pair exhibits large appearance or semantic gaps, yielding ghosting or distorted intermediates~\cite{lee1996image, schaefer2006image}.
Tuning-based methods improve morphing quality but can suffer from limited generalization due to task- or class-specific training data~\cite{yang2024impus, zhang2024diffmorpher}.
Recent diffusion-based methods exploit generative priors for latent-space morphing~\cite{yang2024impus, zhang2024diffmorpher, cao2025freemorph, wang2023interpolating}, yet remain unreliable on highly heterogeneous pairs, partly due to limited adaptability and architecture-specific reuse designs.
To address this, we propose a backbone-agnostic framework that stabilizes denoising via timestep- and layer-wise adaptive inversion-cache reuse and improves semantic coherence with an anchor-correlated prompt triplet.

\subsection{Diffusion Latents and Feature Reuse}
Diffusion models~\cite{ho2020denoising, song2020score, dhariwal2021diffusion} produce hierarchical U-Net features during denoising~\cite{ronneberger2015u}. These intermediate representations encode geometric and semantic cues~\cite{kwon2022diffusion} and are widely used for generation control and stabilization via feature-level guidance or modulation~\cite{ho2022classifier, hong2023improving, zhang2023controlnet, mou2024t2i, si2024freeu, hertz2022prompt}. More specifically, previous morphing methods such as DiffMorpher~\cite{zhang2024diffmorpher} and FreeMorph~\cite{cao2025freemorph} reuse attention-layer Key and Value features to preserve structure, texture, or source identity. In contrast, we argue that broader diffusion feature reuse across layers yields more stable morphing in terms of domain consistency and smoothness (\cref{Sec:PreNObs}), which motivates ACI. 

\subsection{Text-guided Diffusion Models}
Text-conditioned diffusion models~\cite{rombach2022high, nichol2022glide, saharia2022imagen} provide a strong prior for controllable generation, where text token embeddings modulate features via cross-attention. VLMs~\cite{ bai2025qwen2, liu2023visual} align visual and textual semantics, enabling captioning and semantic control widely used in text-guided editing and attention-based manipulation~\cite{patashnik2021styleclip, he2024aid, cao2025freemorph, zhang2024diffmorpher}.
In diffusion-based morphing, endpoint prompts are often obtained independently and then interpolated during sampling~\cite{cao2025freemorph, zhang2024diffmorpher, yang2024impus}. For heterogeneous pairs, such independently formed prompts can yield conflicting conditioning cues and cause semantic drift in intermediate states. To overcome this, we use a VLM to extract a shared anchor and produce anchor-conditioned endpoint prompts, forming an anchor-correlated prompt triplet.
\section{Observations}
\label{Sec:PreNObs}

\begin{figure*}[!t]
    \centering
    \includegraphics[width=0.8\textwidth]{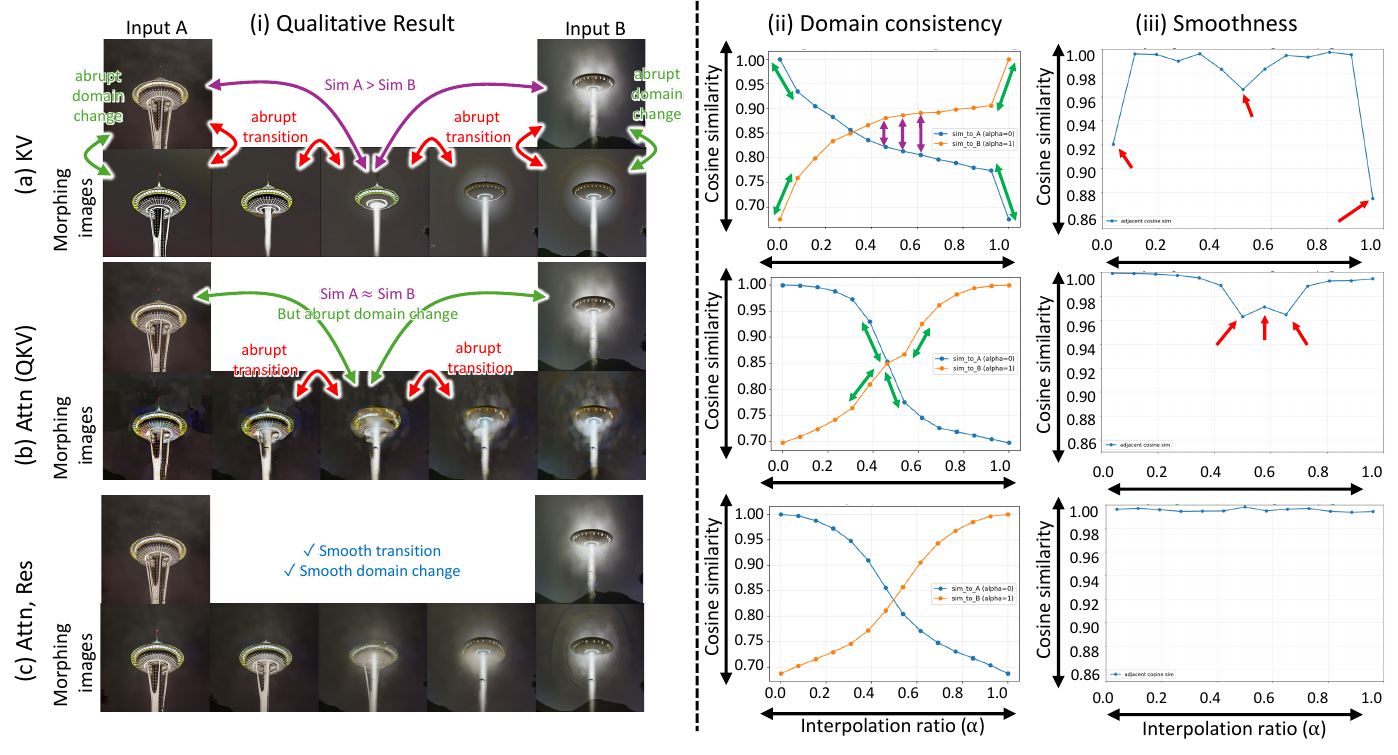}
    \caption{
    Observation on diffusion feature reuse.
    (a) reuses only attention-layer Key/Value features (KV),
    (b) reuses attention features including Query (Attn; QKV),
    and (c) further includes ResNet features (Attn, Res).
    (i) shows qualitative results, (ii) reports domain consistency, and (iii) reports smoothness.
    The results in (ii) and (iii) are averaged over the Morph4Data~\cite{cao2025freemorph} dataset.
    }
    \label{fig:observation}
    \vspace{-1em}
\end{figure*}
\subsubsection{Why We Reuse a Broader Set of Diffusion Features}
Before introducing Adaptive Cache Injection (ACI), we analyze whether reusing only partial diffusion features (such as the attention-layer KV features used in DiffMorpher~\cite{zhang2024diffmorpher} and FreeMorph~\cite{cao2025freemorph}) is sufficient, or whether broader diffusion feature reuse is more desirable. 
\cref{fig:observation} compares three settings: (a) KV-only reuse, (b) reuse of the full attention features (including Query), and (c) extended diffusion feature reuse including additional ResNet features~\cite{he2016deep}.
We analyze domain consistency and smoothness at the feature level using DiffSim~\cite{song2025diffsim}. 
Specifically, \cref{fig:observation}(ii) measures the similarity between each morphing feature and the endpoint features (\(A\), \(B\)) to evaluate domain consistency, while \cref{fig:observation}(iii) measures the similarity between adjacent morphing features to evaluate smoothness. 
Ideally, the similarity curves to the two endpoints intersect near $\alpha=0.5$ and vary smoothly along the interpolation trajectory, while the similarity between adjacent images changes gradually without large oscillations.
As shown in \cref{fig:observation}(a), KV-only reuse leads to overly strong endpoint transitions (green arrow), a bias of the middle image toward one domain (purple arrow), and significant drops in smoothness at certain points (red arrow). 
Including the entire attention feature in \cref{fig:observation}(b) alleviates these issues, and further incorporating ResNet features in \cref{fig:observation}(c) stabilizes both curves. 
These trends are also reflected in the qualitative results in \cref{fig:observation}(i). 
Overall, these observations suggest that broader diffusion feature reuse provides more stable morphing guidance than KV-only reuse, motivating the design of ACI. Furthermore, this analysis method provides a unified perspective for understanding domain consistency and smoothness. 
Based on this observation, we extend it to define a single quantitative metric, GLCS (\cref{pm:glcs}). 

\begin{figure}[t]
\centering
\includegraphics[width=0.9\columnwidth]{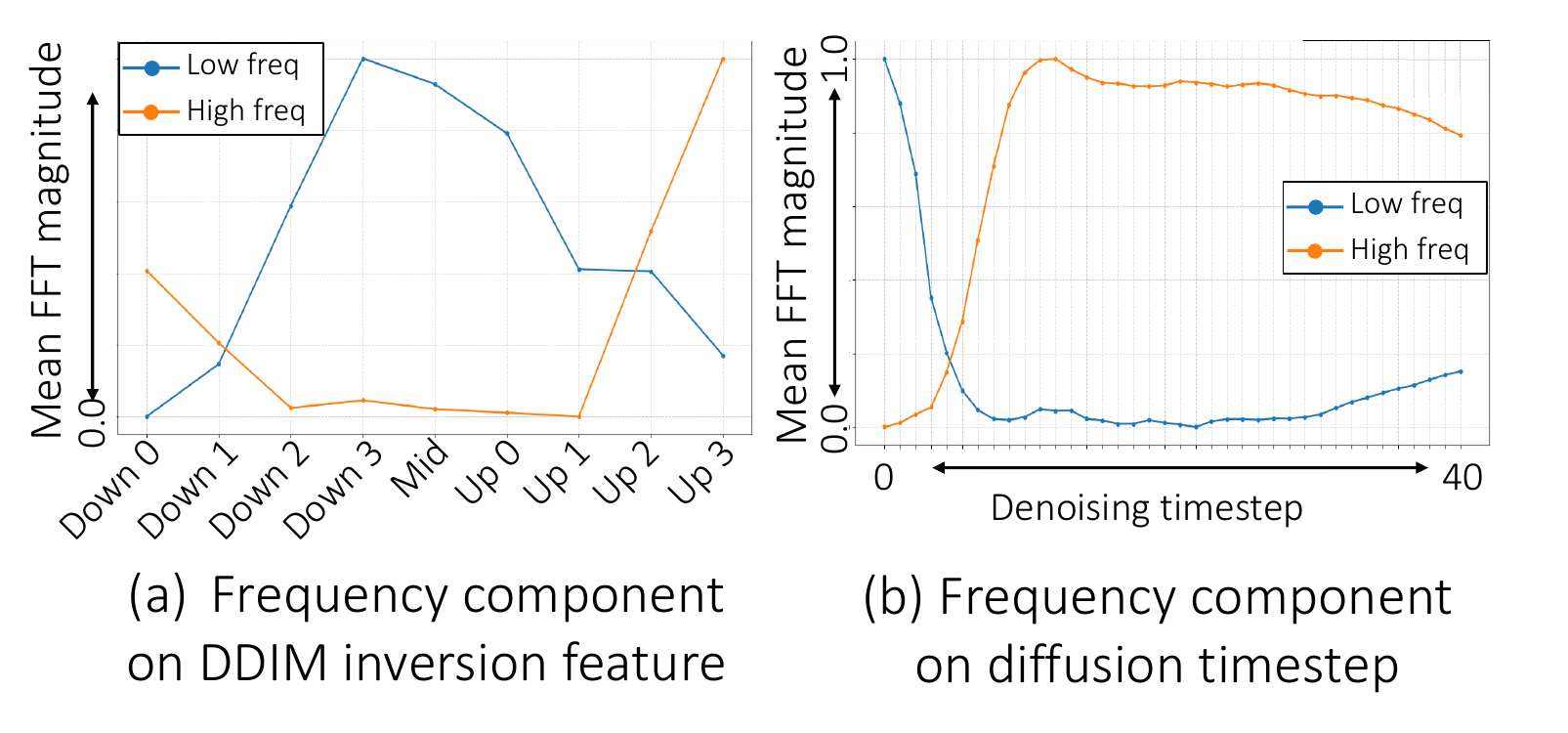}
\caption{Frequency analysis of diffusion features and denoising timesteps.
Low- (blue) and high-frequency (orange) components across (a) U-Net feature layers and (b) DDIM denoising timesteps are measured for the base model without CHIMERA’s ACI and SAP on Morph4Data~\cite{cao2025freemorph}. Values are obtained by applying FFT with masked frequency bands.}
\label{fig:freq_analysis}
\vspace{-0.7cm}
\end{figure}

\subsubsection{How We Reuse Diffusion Features}
Broader diffusion feature reuse is beneficial, but injecting all layers naively is unstable. To address this limitation, we analyze DDIM inversion features and denoising timesteps in Stable Diffusion 2.1~\cite{rombach2022high} from a frequency-domain perspective (Fourier domain).

As shown in \cref{fig:freq_analysis}, mid-block inversion features are dominated by low-frequency components, whereas up-block features contain stronger high-frequency components; likewise, early timesteps emphasize low-frequency structures while later timesteps focus on high-frequency details.
Accordingly, we inject mid-block features at early timesteps and up-block features at later timesteps.
We implement this as Layer- and Timestep-wise Frequency Matching (LTM) in ACI, which jointly considers layer-wise feature spectra and timestep-wise denoising properties.
While \cref{fig:freq_analysis} reports SD 2.1, we find the same coarse-to-fine frequency shift across timesteps to be consistent across backbones (see \cref{exp:further_analysis} and \textit{Suppl}.).

\begin{figure*}[t]
    \centering
    \includegraphics[width=1.0\textwidth]{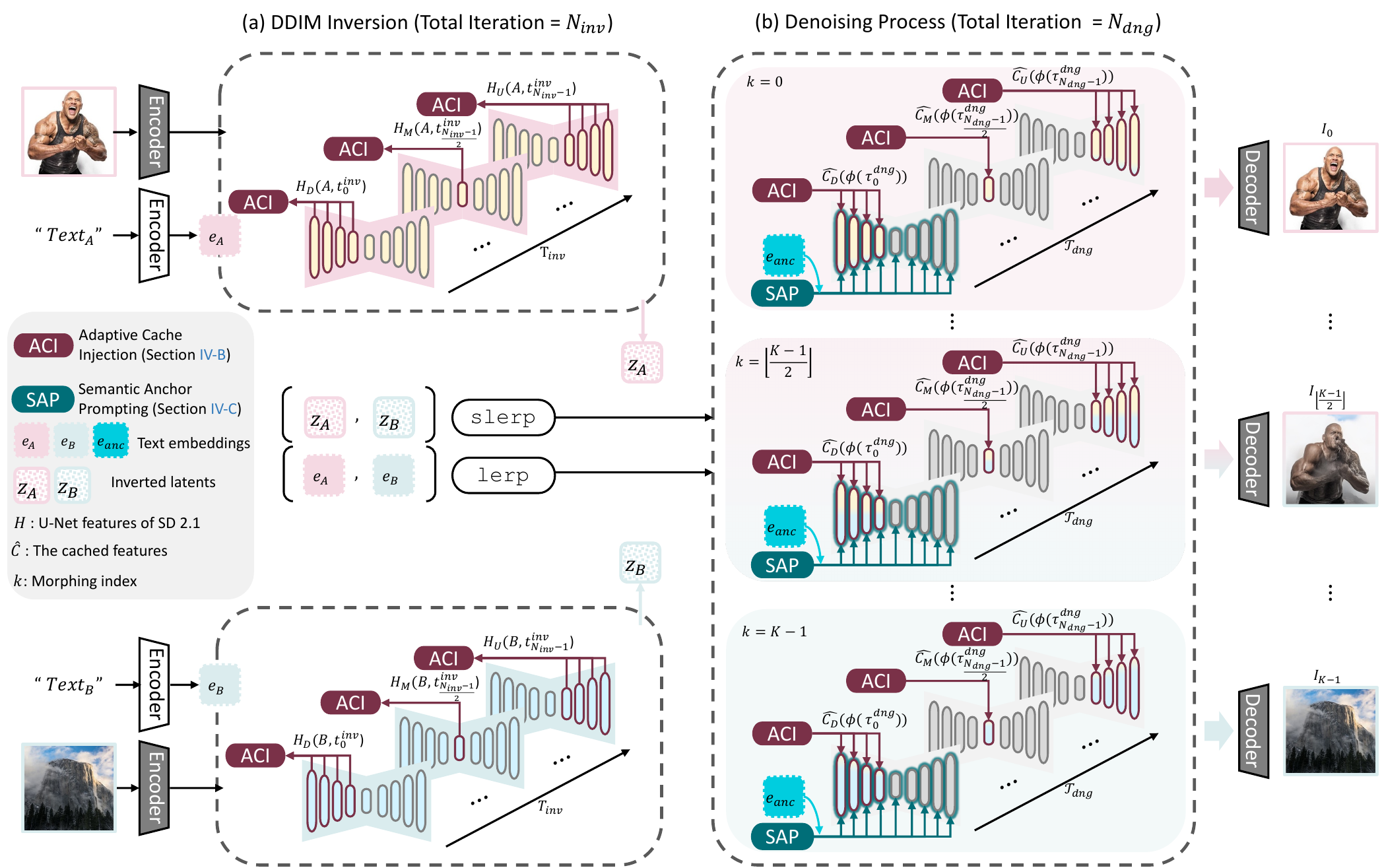}
    \caption{Overview of CHIMERA. (a) DDIM Inversion: Inputs $A$ and $B$ are inverted while caching multi-scale U-Net features from the down, mid, and up blocks. The cached features are interpolated via \texttt{slerp}, forming morphing-aligned latents. (b) Denoising: The interpolated caches are re-injected through ACI. 
    ACI adaptively injects cached features by applying LTM based on the observations in \cref{fig:freq_analysis}. In addition, SAP introduces a VLM-derived anchor-prompt into early cross-attention layers, stabilizing semantics. The full algorithm is provided in Algorithm~1 of the \textit{Suppl}.}
    \label{fig:framework}
\end{figure*}

\section{Proposed Method: CHIMERA}
\label{sec:pm}

\subsection{Overall Pipeline}
\label{subsec:pm_overall}
As shown in \cref{fig:framework}, given input images \(A\) and \(B\), DDIM inversion (\texttt{DDIM})~\cite{song2020denoising} projects them into the latent space to obtain \(z_A = \texttt{DDIM}(A)\) and \(z_B = \texttt{DDIM}(B)\).
Spherical interpolation (\texttt{slerp})~\cite{shoemake1985animating} then yields the $K$-morphing latents $z_k = \texttt{slerp}(z_A,\, z_B;\, \alpha_k), \ k = 0,\dots,K-1$, where \(\alpha_k\) denotes the interpolation weight, \(K\) is the number of intermediate morphing latents, and \(k\) denotes the index of \texttt{slerp}.

During \texttt{DDIM} of \(A\) and \(B\), 
we cache multi\hbox{-}scale U\hbox{-}Net features. Specifically, we record the down, mid, and up features as:
\begin{equation}
H_S(X, t),
\qquad
S \in \{\mathbf{D}, \mathbf{M}, \mathbf{U}\},\;
X \in \{A, B\},\;
t \in \mathrm{T}_{\mathrm{inv}},
\label{eq:cache_layer}
\end{equation}
where $H_S$ denotes the multi-scale U-Net features and $\mathbf{D}$, $\mathbf{M}$, $\mathbf{U}$ represent downsampling, mid, and upsampling blocks.
Here, \(\mathrm{T}_{\mathrm{inv}}=\bigl(t^{\mathrm{inv}}_0,\dots,t^{\mathrm{inv}}_{N_{\mathrm{inv}}-1}\bigr)\) denotes the set of inversion timesteps, where \(N_{\mathrm{inv}}\) is the number of inversion timesteps.
To align the characteristics of layers with those of timesteps, we introduce Layer- and Timestep-wise Frequency Matching (LTM). 
The aligned cached multi-scale feature is denoted as $H_{S^*}(X, t)$.
We then apply \texttt{slerp} to the cached features:
\begin{equation}
\widehat{C}_{S^*}(k, t)
= \texttt{slerp}\bigl(H_{S^*}(A,t),\, H_{S^*}(B,t);\alpha_{k}\bigr),
\label{eq:cache_layer_inversion}
\end{equation}
where $\widehat{C}_{S^*}(k,t)$ denotes interpolated cached U-Net feature, $k$ denotes the \texttt{slerp} index and $t$ denotes the \texttt{DDIM} timestep. These features are then injected into the U-Net during the denoising process according to their matched timesteps.

\begin{figure*}[t]
    \centering
    \includegraphics[width=2.0\columnwidth]{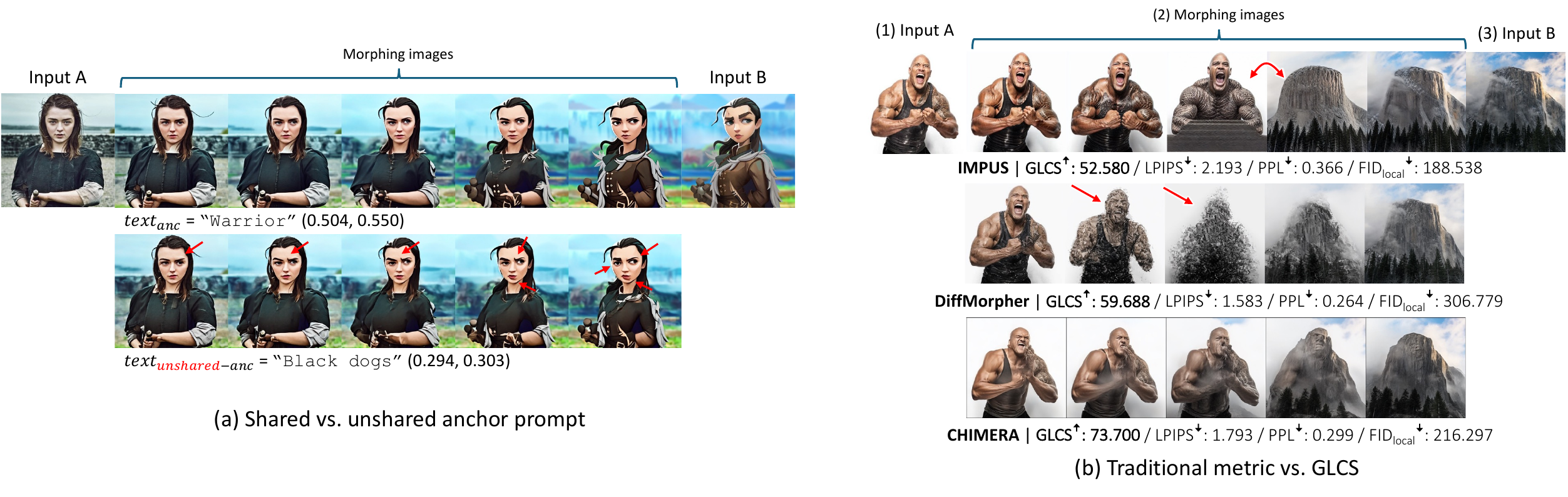}
    \caption{
    (a): Effect of shared vs.\ unshared anchor-prompts.
    The second row uses a shared anchor-prompt, while the third row uses an intentionally unshared anchor-prompt. The two numbers in parentheses denote \textit{cossim} between the anchor-prompt and each endpoint prompt, respectively. Red arrows highlight artifacts under unshared anchors.
    (b): Traditional metric vs. GLCS.
    Given input image pairs (1) and (3), different methods produce morphing sequences shown in (2), highlighting cases where GLCS successfully reflects differences in global--local consistency that are not fully captured by conventional metrics.
    }
    \label{fig:anchor_glcs}
\end{figure*}

\subsection{Adaptive Cache Injection (ACI)}
\label{pm:aci}

As discussed in \cref{Sec:PreNObs}, reusing only the Key and Value features from attention layers cannot guarantee both domain consistency and smoothness. 
To overcome this limitation, we propose Adaptive Cache Injection (ACI), which adaptively leverages all diffusion layers.

As described in \cref{subsec:pm_overall}, during \texttt{DDIM}, we cache down ($\mathbf{D}$), mid ($\mathbf{M}$), and up ($\mathbf{U}$) features for each input (see \cref{eq:cache_layer}), and blend the cached features from $A$ and $B$ via \cref{eq:cache_layer_inversion} for each morphing index $k$.
However, naively reusing all cached layers at all timesteps does not guarantee stable performance.
To address this issue, we introduce \textbf{Layer- and Timestep-wise Frequency Matching (LTM)}, which constructs a frequency-based layer-timestep correspondence \emph{during the caching stage} based on the observations in \cref{Sec:PreNObs}.

Specifically, for a feature tensor $Z \in \mathbb{R}^{C\times H\times W}$, we quantify its spectral property using an FFT-magnitude descriptor $\mathbf{r}(Z)=\mathrm{Pool}\!\left(\frac{1}{C}\sum_{c=1}^{C}\left|\mathcal{F}(Z_c)\right|\right)$, where $\mathcal{F}(\cdot)$ denotes 2D FFT and $\mathrm{Pool}(\cdot)$ is radial (band-wise) pooling. 
Using cached inversion features $H_S(X,t)$, we construct frequency prototypes $\bar{\mathbf{r}}_S$ and $\bar{\mathbf{r}}_t$ for each layer group $S\!\in\!\{\mathbf{D},\mathbf{M},\mathbf{U}\}$ and inversion timestep $t\!\in\!\mathcal{T}_{\mathrm{inv}}$, computed offline as dataset-level averages of the FFT-magnitude descriptor rather than from any specific pair $(A,B)$, so the resulting correspondence is pair-invariant.
We then measure their frequency discrepancy by $D_{S,t}=d\!\left(\bar{\mathbf{r}}_S,\bar{\mathbf{r}}_t\right)$, where $d(\cdot,\cdot)$ is an $\ell_1$ distance. Based on $D_{S,t}$, LTM builds a layer-timestep correspondence by matching each timestep to the most frequency-consistent layer group $S^*(t)=\arg\min_{S\in\{\mathbf{D},\mathbf{M},\mathbf{U}\}} D_{S,t}.$ This precomputed correspondence is used to organize and select cache entries for later ACI injection, enabling frequency-aware feature reuse that better preserves domain consistency and transition smoothness. We then sequentially inject the cached features obtained above into the denoising process. 
However, in our experimental setting, the number of timesteps in DDIM inversion differs from that of the denoising process. 
To align the cached features from specific DDIM inversion timesteps with the denoising timesteps, we introduce \textbf{Inversion-Denoising Timestep Mapping (IDM)}, which linearly maps each denoising step $\tau$ to an inversion timestep $t = \phi(\tau)$. We denote the mapped cached features of images $A$ and $B$ as $H_{S^*}(A,\phi(\tau))$ and $H_{S^*}(B,\phi(\tau))$, respectively. 
By applying \texttt{slerp} to these two features, the final cached feature injected into the denoising process is expressed as follows:
\begin{equation}
\widehat{C}_{S^*}(k,\tau)
=
\texttt{slerp}\!\bigl(H_{S^*}(A,\phi(\tau)),\,H_{S^*}(B,\phi(\tau));\,\alpha_k\bigr).
\label{eq:cache_layer_interp}
\end{equation}
The cached feature $\widehat{C}_{S^*}$ is multiplied by the blending weight $\lambda_S$ and then added as a residual $\tilde{F}_S^{(\tau)} = F_S^{(\tau)} + \lambda_S \cdot \hat{C}_{S^*}(k,\phi(\tau)),$
where $F_S$ denotes the denoising feature at layer $S$, and $\tilde{F}_S$ represents the feature after adding the cached feature as a residual. 
Through this process, ACI provides \textit{layer-wise} and \textit{timestep-wise} guidance, enabling smooth and coherent morphing.

\subsection{Semantic Anchor Prompting (SAP)}
\label{sec:sap}

ACI stabilizes morphing by feature reuse, but semantic coherence can still degrade when the inputs are \textit{heterogeneous}.
In particular, diffusion-based morphing~\cite{zhang2024diffmorpher, cao2025freemorph, yang2024impus} often relies on independently obtained endpoint prompts and interpolates them during sampling, which may yield incompatible or unbalanced textual conditions for intermediates.
To address this, we propose \textbf{Semantic Anchor Prompting (SAP)}, which constructs an anchor-correlated prompt triplet for the input pair and injects the semantically shared anchor-prompt as an auxiliary conditioning in cross-attention to provide a stable semantic guidance.

\subsubsection{Anchor-correlated Prompt Triplet}
Given an input pair, we query a VLM~\cite{bai2025qwen2} with a structured instruction to generate an anchor-correlated prompt triplet $(text_{\mathrm{anc}}, text_A, text_B)$.
The anchor-prompt $text_{\mathrm{anc}}$ captures the shared semantic or structural concept of the pair, while the endpoint prompts $text_A$ and $text_B$ describe each input while preserving the anchor. Unlike prior pipelines that obtain endpoint prompts independently~\cite{cao2025freemorph, yang2024impus, zhang2024diffmorpher}, this anchor-conditioned prompt construction explicitly encourages semantic overlap between the endpoint prompts, making the subsequent interpolation and text conditioning more reliable.
We encode $(text_{\mathrm{anc}}, text_A, text_B)$ using the CLIP text encoder~\cite{radford2021clip} to obtain embeddings $(\mathbf{e}_{\mathrm{anc}}, \mathbf{e}_A, \mathbf{e}_B)$, and denote the cosine similarity as \textit{cossim}.
Following prior findings on the local linearity of CLIP’s embedding space~\cite{patashnik2021styleclip, yang2024impus, brack2023sega, liang2022mind, kawar2023imagic}, semantically related prompts tend to yield closer embeddings, which supports more stable endpoint prompt interpolation. Consistent with this, \cref{fig:anchor_glcs}(a) shows that an unshared anchor-prompt (low $\textit{cossim}$ to the endpoints) leads to degraded morphing quality, e.g., facial distortions. 
\cref{tab:anchor_similarity} further confirms that shared anchors exhibit higher similarity to both endpoints than unshared anchors.
Accordingly, we re-query the VLM until the anchor satisfies an anchor-reliability criterion (e.g., \textit{cossim}$\ge$ 0.45). Full VLM instructions, the anchor-reliability criterion, the activation schedule, and SAP cost or robustness analyses are in the \textit{Suppl}.

\begin{table}
    \centering
    \caption{Average cosine similarity between anchor and endpoint prompts on Morph4Data~\cite{cao2025freemorph}.}
    \label{tab:anchor_similarity}
    \small
    \setlength{\tabcolsep}{10pt}
    \begin{tabular}{lcc}
    \toprule
    {\shortstack{Text\\Prompt}} &
    {\shortstack{Shared\\Anchor-prompt}} &
    {\shortstack{Unshared\\Anchor-prompt}} \\
    \midrule
    $text_A$ & 0.561 & 0.405 \\
    $text_B$ & 0.557 & 0.398 \\
    \bottomrule
    \end{tabular}
\end{table}
\subsubsection{Anchor-guided Cross-Attention}
SAP injects the anchor-prompt as auxiliary conditioning into the cross-attention, where text tokens directly modulate spatial features.
At a denoising step $\tau$, the endpoint and anchor embeddings are projected to $(K_A,V_A)$, $(K_B,V_B)$, and $(K_{\mathrm{anc}},V_{\mathrm{anc}})$, and the anchor Key--Value is concatenated to each endpoint branch:
\begin{equation}
\texttt{Attn}_X=\mathrm{softmax}\!\left(\tfrac{Q[K_X\Vert K_{anc}]^\top}{\sqrt d}\right)[V_X\Vert V_{anc}],\; X \in\{A,B\},
\label{eq:sap1}
\end{equation}
where $Q$ is the Query from the current diffusion feature and $d$ is the attention dimension.
This anchor-guided design encourages intermediate features to remain compatible with the semantics shared by both endpoints, reducing drift toward either side.
We activate SAP only at early denoising timesteps to avoid over-constraining fine details, as validated by the ablation in the \textit{Suppl.}

\section{Proposed Metric: GLCS}
\label{pm:glcs}

\subsection{Motivation} 
FID\(_{\text{local}}\)~\cite{heusel2017fid}, FID\(_{\text{global}}\)~\cite{heusel2017fid}, 
LPIPS~\cite{zhang2018lpips}, and PPL~\cite{karras2020analyzing} are widely used for quantitative morphing evaluation (see \cref{sec:experiment}). 
However, they often misalign with perceived morphing quality. 
LPIPS and PPL only measure adjacent-image similarity, and thus sequences that deviate from the input images \(A\) and \(B\) can still score well if local changes are small.
\noindent For example, in \cref{fig:anchor_glcs}(b)(2), the \textit{second} row yields lower LPIPS and PPL than the \textit{third}, even though the latter is visually superior. 
FID\(_{\text{local}}\) also fails to reflect perceptual domain consistency, as it averages the distribution gap between \(A,B\) and all intermediates without considering the interpolation ratio. 
Consequently, it often favors images that resemble both inputs simultaneously rather than those forming a natural transition. 
As shown in \cref{fig:anchor_glcs}(b)(2), FID\(_{\text{local}}\) incorrectly prefers the \textit{first} row over the \textit{third}, even though the third better preserves the domain characteristics (e.g., stone texture and facial identity). 
To address these issues, we propose the \textbf{Global-Local Consistency Score (GLCS)}, motivated by the observations in \cref{Sec:PreNObs}, where the domain-consistency and smoothness analysis align well with human perception. GLCS further shows 80\% agreement with user study preferences (\cref{exp:eval}), providing a morphing-specific metric that jointly measures domain consistency and smoothness.

GLCS consists of two complementary components. First, the \textbf{Global Consistency Score (GCS)} measures \emph{domain consistency}. 
It checks whether each image follows the expected global trend between the two input images \(A\) and \(B\). 
We obtain this trend by interpolating the endpoint similarities with \texttt{slerp}, so the sequence should change in a balanced way from \(A\) to \(B\). Second, the \textbf{Local Consistency Score (LCS)} measures \textit{smoothness}. 
It checks whether the similarity of each image changes smoothly with respect to its neighbors. 
Thus, LCS captures local continuity along the morphing transition.
We use a DiffSim-based \cite{song2025diffsim} bounded similarity \(s(\cdot,\cdot)\), which is sensitive to low-level structure and also reflects style and semantic similarity. 
Both GCS and LCS are clamped to \([0,1]\) for stability and interpretability.

\subsection{Detailed Description of GLCS}

Let \(A\) and \(B\) be the endpoint images, and let \(\{I_k\}_{k=0}^{K-1}\) be the predicted morphing images ordered from \(A\) to \(B\).
We adopt a DiffSim-based bounded similarity~\cite{song2025diffsim}, denoted by:
\begin{equation}
s(X,Y)\in[-1,1],
\label{eq:sim_def}
\end{equation}

For each index \(k\), we define the normalized interpolation weight:
\begin{equation}
\alpha_k = \frac{k+1}{K+1}, \qquad k=0,\dots,K-1,
\label{eq:alpha_def}
\end{equation}
where \(\alpha_k\) encodes the ideal mixing ratio between the two endpoints \(A\) and \(B\).

For convenience, we denote the similarities between each image and the endpoints as:
\begin{equation}
s_X(k) = s(X, I_k), \qquad X\in\{A,B\},
\label{eq:sxk_def}
\end{equation}
and introduce a clamping operator to the unit interval,
\begin{equation}
[x]_0^1 = \min\bigl(1,\max(0,x)\bigr),
\label{eq:clamp_def}
\end{equation}
so that all per-image consistency terms are normalized to \([0,1]\).

\vspace{4pt}
\subsubsection{Global Consistency Score (GCS)}
We first model the global expected trend of similarities along the morphing sequence. 
Given the four endpoint similarities:
\begin{equation}
s(A,A),\; s(A,B),\; s(B,A),\; s(B,B),
\label{eq:endpoint_sims}
\end{equation}
we define the expected similarity of image \(I_k\) to each endpoint \(X\in\{A,B\}\) using spherical interpolation (\texttt{slerp}) in similarity space:
\begin{equation}
\bar{s}_X(k)
= \texttt{slerp}\bigl(s(X,A),\,s(X,B);\,\alpha_k\bigr).
\label{eq:gcs_slerp}
\end{equation}

Using this expected trend, we define the per-image global consistency term as:
\begin{equation}
g_k =
\bigl[\,1 - |s_A(k) - \bar{s}_A(k)|\,\bigr]_0^1
\cdot
\bigl[\,1 - |s_B(k) - \bar{s}_B(k)|\,\bigr]_0^1,
\label{eq:gk_def}
\end{equation}
where each factor evaluates how well the measured similarity \(s_X(k)\) matches the expected similarity \(\bar{s}_X(k)\) for \(X\in\{A,B\}\).
Finally, we define the Global Consistency Score (GCS) as:
\begin{equation}
\mathrm{GCS}
= \frac{1}{K} \sum_{k=0}^{K-1} {g}_k.
\label{eq:gcs_def}
\end{equation}

\vspace{4pt}
\subsubsection{Local Consistency Score (LCS)}
To capture local smoothness along the morphing trajectory, we define a local expectation that relates each image to its temporal neighbors. 
For each \(X\in\{A,B\}\), we first estimate the locally expected similarity at index \(k\) as:
\begin{equation}
\tilde{s}_X(k) =
\begin{cases}
s_X(1), & k = 0, \\
\frac{1}{2}\bigl(s_X(k-1) + s_X(k+1)\bigr), & 0 < k < K-1, \\
s_X(K-2), & k = K-1,
\end{cases}
\label{eq:local_expectation}
\end{equation}
where boundary images use their single temporal neighbor and interior images use the average of the preceding and succeeding images. Using \(\tilde{s}_X(k)\), we define the per-image local consistency term as:
\begin{equation}
\ell_k =
\bigl[\,1 - |s_A(k) - \tilde{s}_A(k)|\,\bigr]_0^1
\cdot
\bigl[\,1 - |s_B(k) - \tilde{s}_B(k)|\,\bigr]_0^1,
\label{eq:lk_def}
\end{equation}
which measures whether the similarity to each endpoint evolves smoothly when compared to neighboring images. The resulting Local Consistency Score (LCS) is given as:
\begin{equation}
\mathrm{LCS}
= \frac{1}{K} \sum_{k=0}^{K-1} \ell_k.
\label{eq:lcs_def}
\end{equation}

\vspace{4pt}
\subsubsection{Global-Local Consistency Score (GLCS)}
Finally, we combine these two complementary components into our morphing-oriented metric, the \textbf{G}lobal--\textbf{L}ocal \textbf{C}onsistency \textbf{S}core (GLCS), defined as:
\begin{equation}
\mathrm{GLCS} = \sqrt{\mathrm{GCS} \cdot \mathrm{LCS}}.
\label{eq:glcs_def}
\end{equation}

For detailed explanations of the algorithm, edge-case analyses of LCS and GCS, and additional qualitative results, please refer to the \textit{Suppl}.
\begin{table*}[!t]
\centering
\caption{Quantitative results on Morph4Data and MorphBench datasets. The best scores are marked in bold, while the second best are underlined.}
\begingroup
\small
\setlength{\tabcolsep}{6pt}

\begin{minipage}{0.49\textwidth}
\centering
\resizebox{\textwidth}{!}{%
\begin{tabular}{l c c c c c}
\toprule
\multicolumn{6}{c}{\textbf{Morph4Data}} \\
\midrule
\textbf{Model name}
& FID$_{\text{local}}\downarrow$
& FID$_{\text{global}}\downarrow$
& LPIPS$\downarrow$
& PPL$\downarrow$
& GLCS$\uparrow$ \\
\midrule
IMPUS
& \textcolor{red}{\textbf{150.133}}
& \textcolor{red}{\textbf{70.231}}
& 1.913
& 0.319
& 81.903 \\

DiffMorpher
& 181.992
& 92.548
& \textcolor{blue}{\underline{1.638}}
& \textcolor{blue}{\underline{0.273}}
& 85.160 \\

FreeMorph
& 191.349
& 98.444
& 1.973
& 0.329
& \textcolor{blue}{\underline{86.641}} \\

\textbf{CHIMERA (Ours)}
& \textcolor{blue}{\underline{161.054}}
& \textcolor{blue}{\underline{82.308}}
& \textcolor{red}{\textbf{1.576}}
& \textcolor{red}{\textbf{0.263}}
& \textcolor{red}{\textbf{88.079}} \\
\bottomrule
\end{tabular}%
}
\end{minipage}
\hfill
\begin{minipage}{0.49\textwidth}
\centering
\resizebox{\textwidth}{!}{%
\begin{tabular}{l c c c c c}
\toprule
\multicolumn{6}{c}{\textbf{MorphBench}} \\
\midrule
\textbf{Model name}
& FID$_{\text{local}}\downarrow$
& FID$_{\text{global}}\downarrow$
& LPIPS$\downarrow$
& PPL$\downarrow$
& GLCS$\uparrow$ \\
\midrule
IMPUS
& \textcolor{red}{\textbf{93.417}}
& \textcolor{red}{\textbf{44.287}}
& 1.296
& 0.216
& 89.426 \\

DiffMorpher
& 133.086
& 62.127
& \textcolor{blue}{\underline{1.044}}
& \textcolor{blue}{\underline{0.174}}
& \textcolor{blue}{\underline{91.887}} \\

FreeMorph
& 148.972
& 81.019
& 1.494
& 0.249
& 90.566 \\

\textbf{CHIMERA (Ours)}
& \textcolor{blue}{\underline{113.794}}
& \textcolor{blue}{\underline{61.970}}
& \textcolor{red}{\textbf{1.037}}
& \textcolor{red}{\textbf{0.173}}
& \textcolor{red}{\textbf{93.373}} \\
\bottomrule
\end{tabular}%
}
\end{minipage}

\endgroup
\label{table:main_experiment_frames_side_by_side}
\end{table*}

\section{Experiment}
\label{sec:experiment}

\subsection{Implementation Details} 
Our proposed model, CHIMERA, is based on Stable Diffusion 2.1~\cite{rombach2022high}. 
For ACI, we use $N_{inv}=40$ DDIM inversion timesteps and $N_{dng}=50$ denoising timesteps. 
The cached features are weighted with $\lambda_S=0.4$. For SAP, we use Qwen2.5-VL~\cite{bai2025qwen2} as the VLM.
Further implementation details are provided in the \textit{Suppl.} for reproducibility.

\subsection{Evaluation Datasets}
MorphBench~\cite{zhang2024diffmorpher} contains 90 object pairs covering object metamorphosis and animation-style continuous transformations. 
Morph4Data~\cite{cao2025freemorph} expands diversity with broader semantic and layout variations, including similar-layout or different-semantic pairs, aligned-semantic pairs (e.g., faces, cars), random ImageNet-1K~\cite{imagenet15russakovsky} pairs, and dog--cat pairs collected online.

\subsection{Evaluation Metrics} We conduct quantitative evaluation using the metrics adopted in prior methods~\cite{yang2024impus,zhang2024diffmorpher,cao2025freemorph}, including FID\(_{\text{local}}\)~\cite{heusel2017fid}, FID\(_{\text{global}}\)~\cite{heusel2017fid}, LPIPS~\cite{zhang2018lpips}, PPL~\cite{karras2020analyzing}, and our proposed GLCS. 
For detailed definitions and evaluation protocols, please refer to the \textit{Suppl}.

\subsection{Quantitative and Qualitative Evaluations}\label{exp:eval}

\subsubsection{Quantitative Evaluation}\label{exp:quant}
We evaluate CHIMERA using the standard metrics adopted in prior morphing methods~\cite{yang2024impus,zhang2024diffmorpher,cao2025freemorph}, including 
FID\(_{\text{global}}\), FID\(_{\text{local}}\), LPIPS, and PPL, together with our proposed GLCS. 
\cref{table:main_experiment_frames_side_by_side} reports results for 5-image morphing.
As shown in the table, existing methods exhibit a clear trade-off.
IMPUS~\cite{yang2024impus} achieves strong performance on FID\(_{\text{local}}\) and FID\(_{\text{global}}\), but records worse LPIPS, PPL, and GLCS scores, indicating abrupt transitions. 
DiffMorpher~\cite{zhang2024diffmorpher} achieves the second best LPIPS and PPL, yet its FID and GLCS scores are lower, suggesting weak domain consistency.
FreeMorph~\cite{cao2025freemorph} performs poorly on conventional metrics due to over-smoothing and excessive color saturation. However, because transitions between adjacent images are relatively smooth, it attains comparatively competitive GLCS results.
In contrast, CHIMERA achieves balanced performance across FID\(_{\text{local}}\), FID\(_{\text{global}}\), LPIPS, and PPL. 
It significantly outperforms the zero-shot baseline FreeMorph and achieves results competitive with tuning-based methods. 
Moreover, CHIMERA attains the highest GLCS and the lowest LPIPS and PPL scores on both Morph4Data and MorphBench, demonstrating strong \emph{domain consistency} and \emph{smoothness} simultaneously. Additional quantitative results for the 14-image setting are provided in the \textit{Suppl}.

\begin{figure}[t]
    \centering
    \includegraphics[width=\linewidth]{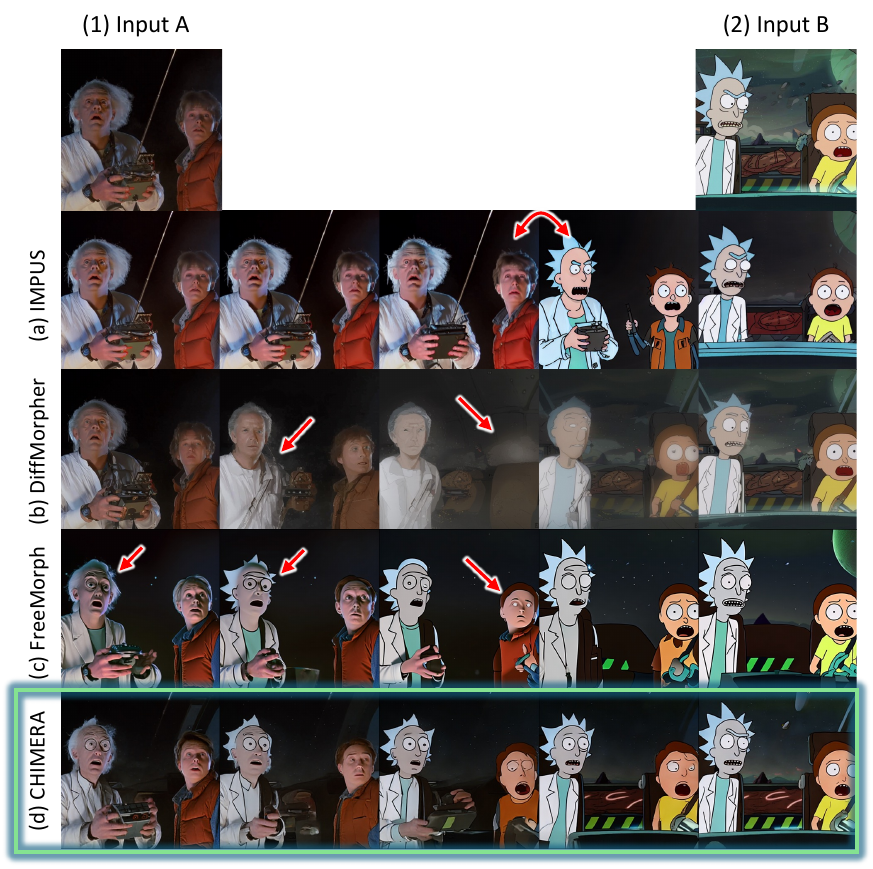}
    \vspace{-0.5em}
    \caption{Qualitative comparisons with existing methods. (1)–(2) denote the input image pairs. (a)–(d) present results on Morph4Data~\cite{cao2025freemorph}.}
    \label{fig:main_qual_1}
\end{figure}

\subsubsection{User Study}\label{exp:userstudy}
In a user study under four perceptual criteria, CHIMERA is consistently preferred over all baselines.
\cref{fig:teaser1}(b) shows the pairwise win--tie--loss outcomes, where CHIMERA wins against FreeMorph and latent \texttt{slerp} in most cases and records more wins than losses against DiffMorpher and IMPUS. Notably, this subjective trend is consistent with our GLCS-based ranking, suggesting that GLCS \textit{aligns well with human preference}. Detailed user study protocol, additional statistics, and further analyses are provided in the \textit{Suppl.}

\subsubsection{Qualitative Comparison}\label{exp:qual}
As shown by the red arrows in \cref{fig:main_qual_1}(a), IMPUS produces relatively natural transitions near the endpoints of the input images, but abrupt changes appear in the middle of the sequence. 
DiffMorpher tends to lose semantic information from both inputs when the input pair is highly dissimilar (red arrow). 
FreeMorph generates excessive over-smoothing and color saturation, producing visual characteristics that deviate from the original images (red arrow). 
In contrast, CHIMERA preserves the semantics of both \(A\) and \(B\) while maintaining smooth transitions and perceptually consistent textures. 
Additional qualitative results are provided in the \textit{Suppl.}, where we also include results for the 14-image setting.

\subsection{Ablation Studies}
\label{ablation}

\begin{table*}[t]
\centering
\caption{Ablation study on caching block and layer types. 
Left: ablation over the major diffusion blocks, namely the down, mid, and up blocks. 
Right: ablation study on the layers within each diffusion block. The layers are categorized into Key and Value features, Attention layers, and ResNet features.}
\footnotesize
\begin{minipage}{0.48\textwidth}
\centering
\resizebox{\textwidth}{!}{%
\begin{tabular}{c c c c c c c}
\toprule
\multicolumn{7}{c}{ACI Ablation - Block type} \\
\midrule
    & Block & $\mathrm{FID}_{\text{local}}\downarrow$ & $\mathrm{FID}_{\text{global}}\downarrow$ & LPIPS$\downarrow$ & PPL$\downarrow$ & GLCS$\uparrow$ \\
\midrule
(a) & $\{\textbf{D}\}$ 
& 181.560
& 92.174
& 1.802 
& 0.300
& 87.904 \\
(b) & $\{\textbf{D},\textbf{M}\}$ 
& \textcolor{blue}{\underline{178.256}}
& \textcolor{blue}{\underline{89.889}}
& \textcolor{blue}{\underline{1.740}}
& \textcolor{blue}{\underline{0.292}}
& \textcolor{blue}{\underline{87.886}} \\
(c) & $\{\textbf{D},\textbf{M},\textbf{U}\}$ \textbf{(Ours)} 
& \textcolor{red}{\textbf{161.054}} 
& \textcolor{red}{\textbf{82.308}} 
& \textcolor{red}{\textbf{1.576}} 
& \textcolor{red}{\textbf{0.263}} 
& \textcolor{red}{\textbf{88.079}} \\
\bottomrule
\end{tabular}%
}
\end{minipage}
\hfill
\begin{minipage}{0.48\textwidth}
\centering
\resizebox{\textwidth}{!}{%
\begin{tabular}{c l c c c c c}
\toprule
\multicolumn{7}{c}{ACI Ablation - Layer type} \\
\midrule
 & Layer 
& $\mathrm{FID}_{\text{local}}\downarrow$ 
& $\mathrm{FID}_{\text{global}}\downarrow$ 
& LPIPS$\downarrow$ 
& PPL$\downarrow$ 
& GLCS$\uparrow$ \\
\midrule
(a) & {KV} 
& 209.809
& 102.093 
& 1.908
& 0.318 
& 87.096 \\
(b) & {Attn} 
& \textcolor{blue}{\underline{183.486}} 
& \textcolor{blue}{\underline{92.829}} 
& \textcolor{blue}{\underline{1.785}} 
& \textcolor{blue}{\underline{0.298}} 
& \textcolor{blue}{\underline{87.159}} \\
(c) & {Attn, Res} \textbf{(Ours)} 
& \textcolor{red}{\textbf{161.054}}
& \textcolor{red}{\textbf{82.308}}
& \textcolor{red}{\textbf{1.576}}
& \textcolor{red}{\textbf{0.263}}
& \textcolor{red}{\textbf{88.079}} \\
\bottomrule
\end{tabular}%
}
\end{minipage}
\label{tab:ACI_ab_layer_block_type}
\end{table*}
\begin{table}[t]
\centering
\caption{Ablation studies on ACI. Left: impact of the injection weight $\lambda_S$. Right: effect of Layer- and Timestep-wise Frequency Matching (LTM).}
\begin{minipage}{0.67\columnwidth}
\centering
\small
\setlength{\tabcolsep}{3pt}
\resizebox{\linewidth}{!}{%
\begin{tabular}{c c c c c c c}
\toprule
\multicolumn{7}{c}{(a) ACI Ablation -- layer weight} \\
\midrule
 & $\lambda_S$ & $\mathrm{FID}_{\text{local}}\downarrow$ & $\mathrm{FID}_{\text{global}}\downarrow$ & LPIPS$\downarrow$ & PPL$\downarrow$ & GLCS$\uparrow$ \\
\midrule
(i) & 0.1 
& \textcolor{blue}{\underline{178.749}}
& \textcolor{blue}{\underline{90.525}}
& \textcolor{blue}{\underline{1.794}}
& \textcolor{blue}{\underline{0.299}}
& 86.236 \\
(ii) & 0.4 \textbf{(Ours)}  
& \textcolor{red}{\textbf{161.054}}
& \textcolor{red}{\textbf{82.308}}
& \textcolor{red}{\textbf{1.576}}
& \textcolor{red}{\textbf{0.263}}
& \textcolor{red}{\textbf{88.079}} \\
(iii) & 0.7 
& 186.378
& 92.275
& 1.830
& 0.353
& \textcolor{blue}{\underline{87.888}} \\
(iv) & 1.0 
& 222.817
& 117.846
& 1.952
& 0.394
& 87.599 \\
\bottomrule
\end{tabular}}
\end{minipage}
\begin{minipage}{0.3\columnwidth}
\centering
\small
\setlength{\tabcolsep}{3pt}

\resizebox{\linewidth}{!}{%
\begin{tabular}{c c c}
\toprule
\multicolumn{3}{c}{(b) ACI Ablation -- LTM} \\
\midrule
 LTM & \textcolor{red}{\ding{55}} & \textcolor{ForestGreen}{\ding{51}} \\
\midrule
$\mathrm{FID}_{\text{local}}\downarrow$
& 182.279
& \textcolor{red}{\textbf{161.054}} \\
$\mathrm{FID}_{\text{global}}\downarrow$
& 92.8293
& \textcolor{red}{\textbf{82.308}} \\
LPIPS$\downarrow$
& 1.617
& \textcolor{red}{\textbf{1.576}} \\
PPL$\downarrow$
& 0.270
& \textcolor{red}{\textbf{0.263}} \\
GLCS$\uparrow$
& 87.492
& \textcolor{red}{\textbf{88.079}} \\
\bottomrule
\end{tabular}}
\end{minipage}
\label{tab:aci_ab_weight_ltm}
\end{table}

\subsubsection{Caching Block and Layer Type on ACI} \label{exp:ACI_ab_1}
\begin{figure}
    \centering
    \includegraphics[width=\linewidth]{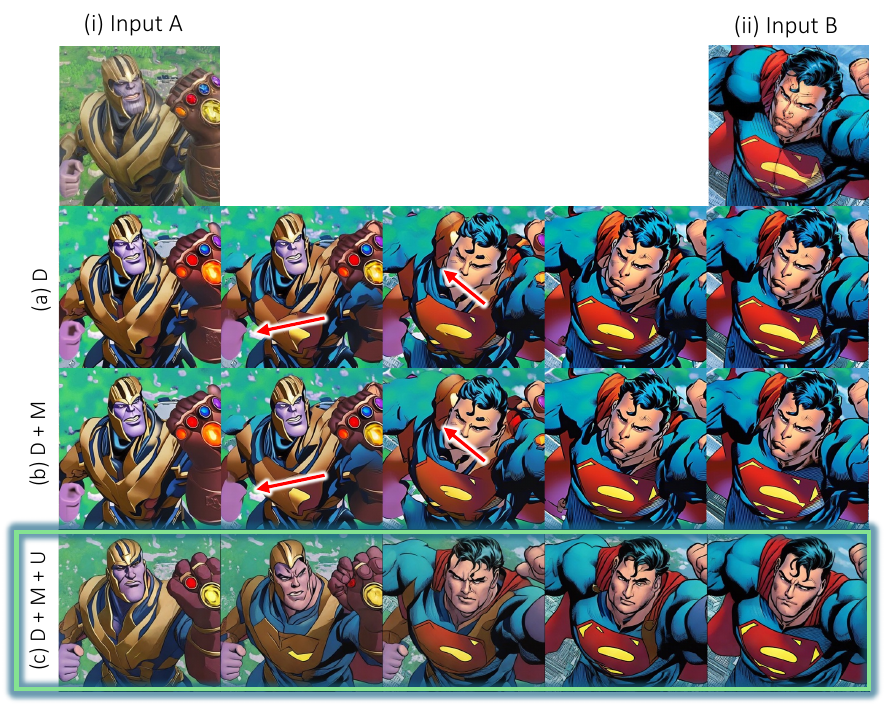}
    \caption{Qualitative results based on the types of cached block in ACI. (i) and (ii) represent the input image pair, while D, M, and U denote the down, mid, and up block, respectively.}
    \label{fig:ACI_ab_block_type}
\end{figure}

\begin{figure}
    \centering
    \includegraphics[width=\linewidth]{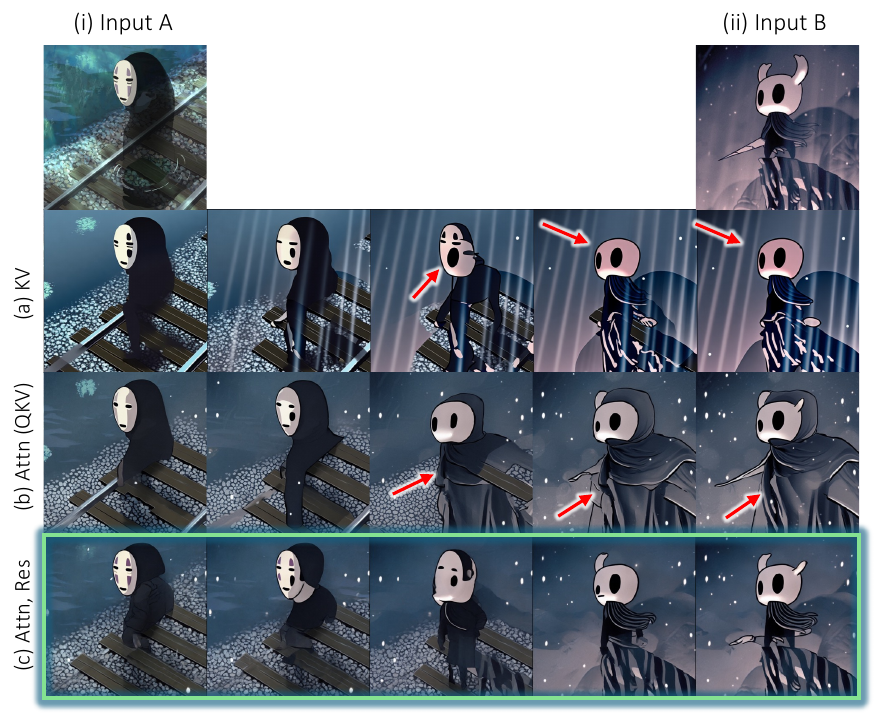}
    \caption{Qualitative results for different cached layer types in ACI. KV denotes the Key and Value features, Attn refers to the entire attention layer, and Res represents the ResNet layers.}
    \label{fig:ACI_ab_layer_type}
\end{figure}

To verify the effectiveness of leveraging all components of diffusion features in ACI, we provide both qualitative and quantitative results in \cref{tab:ACI_ab_layer_block_type}, \cref{fig:ACI_ab_block_type}, and \cref{fig:ACI_ab_layer_type}. 
As indicated by the red arrows in \cref{fig:ACI_ab_block_type}(a),(b), using only specific diffusion blocks leads to the loss of meaningful elements from the two inputs (e.g., an arm disappears in the second column) or degradation of fine details (e.g., the fist becomes over-smoothed in the first and second columns). 
In contrast, using all blocks in \cref{fig:ACI_ab_block_type}(c) preserves fine details while properly incorporating key elements from both images $A$ and $B$. 
This trend is clearly reflected in the left table of \cref{tab:ACI_ab_layer_block_type}. 
Similarly, as shown by the red arrows in \cref{fig:ACI_ab_layer_type}(a),(b), when only partial layers are used, insufficient guidance is provided. 
As a result, important details may disappear (e.g., the horn vanishes in (a)), or unintended objects may be generated (e.g., a cloak-like structure appears in (b)). 
In contrast, leveraging all layers in \cref{fig:ACI_ab_layer_type}(c) preserves the fine details of both input images and avoids generating unexpected objects. 
This is also supported by the quantitative results in the right table of \cref{tab:ACI_ab_layer_block_type}.

\begin{figure}[!t]
    \centering
    \includegraphics[width=\linewidth]{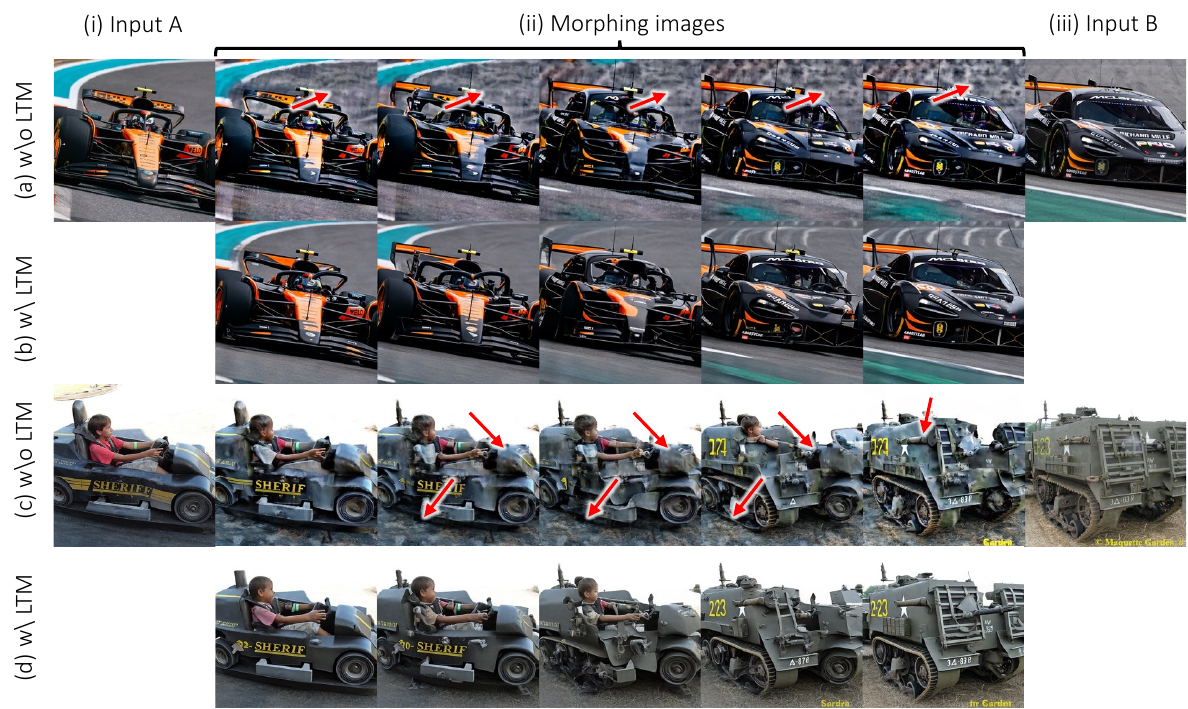}
    \caption{Qualitative comparison with and without LTM. Each row shows the (ii) intermediate frames generated between the two inputs (i) A and (iii) B. Rows (a) and (c) are obtained without LTM, and rows (b) and (d) with LTM.}
    \label{fig:ltm_ab}
\end{figure}

\subsubsection{Layer- and Timestep-wise Frequency Matching (LTM) of ACI} \label{exp_subsubsec:ltm}
To validate the effectiveness of LTM, which considers the layer-wise and timestep-wise characteristics of diffusion features, we provide quantitative evaluations in \cref{tab:aci_ab_weight_ltm}(b). 
As shown in \cref{tab:aci_ab_weight_ltm}(b), applying LTM consistently improves all evaluation metrics compared to the setting without LTM. 
Without LTM, excessive guidance often introduces noise or injects overly strong details, resulting in unnaturally sharp and unstable structures (\cref{fig:ltm_ab}(a),(b)). In contrast, incorporating LTM suppresses such noise and prevents overly emphasized details, leading to more stable and visually consistent results. 
These results support the motivation for applying LTM within ACI.

\subsubsection{Caching Injection Weight of ACI} 
\begin{figure}[t]
  \centering
  \includegraphics[width=1.0\linewidth]{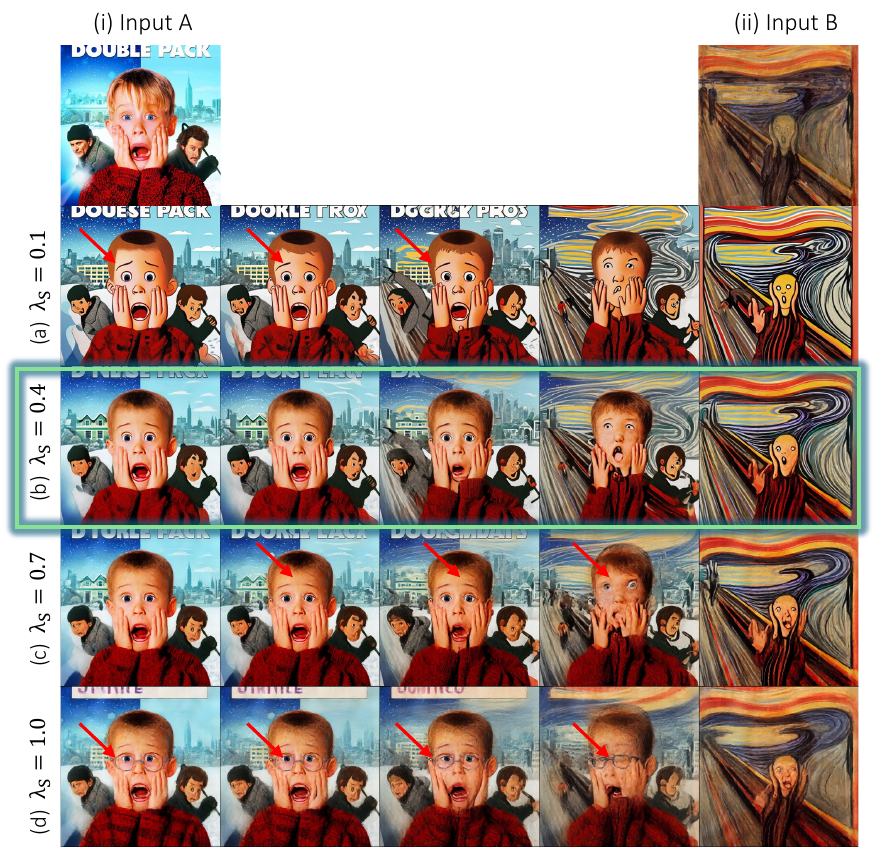}
  \caption{Qualitative results for different injection weights of the cached ACI features in the denoising process. (i) and (ii) denote the input image pair, and (a)--(d) show the results for $\lambda_S$ values of 0.1, 0.4, 0.7, and 1.0, respectively.}
  \label{fig:ACI_ab_layer_weight}
\end{figure}

In \cref{tab:aci_ab_weight_ltm}(a), we provide the quantitative evaluation of the injection weight. We set $\lambda_S$ to $0.4$, which achieves the best FID scores. Although the LPIPS and PPL values are relatively higher, we choose $\lambda_S = 0.4$ as the final weight because GLCS offers a more reliable assessment of smoothness. Additional qualitative results are provided in the \textit{Suppl}.
As shown in \cref{fig:ACI_ab_layer_weight}(a) (red arrow), when the injection weight in ACI is set too small, the results exhibit over-smoothing and saturated colors. 
This indicates that, without a sufficient ACI effect, the diffusion model tends to produce its characteristic artifacts. 
In addition, the 2nd, 3rd, and 4th column images in \cref{fig:ACI_ab_layer_weight}(c) (red arrow) become noticeably noisy, and the 1st–4th images in \cref{fig:ACI_ab_layer_weight}(d) (red arrow) generate glasses that do not exist in the input image, while the outputs also appear noisy and blurry. 
These observations show that when the ACI weight is overly large, the morphing trajectory is excessively constrained, causing high-frequency details that do not exist in the original images to be injected.

\begin{table}
    \centering
    \caption{Ablation on anchor-correlated prompt triplet.
    Independent endpoint prompts, anchor-conditioned endpoint prompts, and the full anchor-correlated prompt triplet are compared.}
    \label{tab:sap_prompt_ablation}
    \scriptsize
    \resizebox{\linewidth}{!}{%
    \begin{tabular}{l c c c c c}
        \toprule
        Method & FID$_{\text{local}}\downarrow$ & FID$_{\text{global}}\downarrow$ & LPIPS$\downarrow$ & PPL$\downarrow$ & GLCS$\uparrow$ \\
        \midrule
        (a) Base$+$per-image prompts & 227.558 & 110.977 & 1.997 & 0.316 & 84.610 \\
        (b) Base$+$paired prompts & \textcolor{blue}{\underline{205.799}} & \textcolor{blue}{\underline{101.291}} & 1.941 & 0.323 & 84.598 \\
        (c) Base$+$SAP & 209.862 & 102.401 & \textcolor{blue}{\underline{1.907}} & \textcolor{blue}{\underline{0.318}} & \textcolor{blue}{\underline{88.076}} \\
        (d) \textbf{CHIMERA (Ours)} & \textcolor{red}{\textbf{161.054}} & \textcolor{red}{\textbf{82.308}} & \textcolor{red}{\textbf{1.576}} & \textcolor{red}{\textbf{0.263}} & \textcolor{red}{\textbf{88.079}} \\
        \bottomrule
    \end{tabular}%
    }
\end{table}
\subsubsection{Anchor-correlated Prompt Triplet of SAP}
\cref{tab:sap_prompt_ablation} shows the impact of prompt construction for morphing.
Row (a) uses independently captioned endpoint prompts following prior work~\cite{cao2025freemorph, liu2024llavanext}, which can yield incompatible conditioning for heterogeneous pairs.
Replacing them with our anchor-conditioned endpoint prompts (b) improves most conventional metrics, while GLCS remains comparable, highlighting the benefit of anchor-conditioned prompt formation.
Activating SAP with the shared anchor and the full triplet $(text_{\mathrm{anc}}, text_A, text_B)$ (c) further improves transition smoothness and morphing quality, as reflected by LPIPS, PPL, and GLCS, indicating that an explicit shared anchor-prompt stabilizes intermediate semantics. Finally, CHIMERA (d) combines SAP with ACI and achieves the best overall performance, confirming that semantic anchoring and adaptive cache reuse are complementary. The full VLM instruction template for SAP is provided as follows.

\begin{center}
\scalebox{0.85}{%
    \begin{tcolorbox}[colback=gray!4, colframe=gray!40, boxrule=0.35pt]
    \small
    \textbf{Prompt template used for Qwen2.5-VL.}
    
    \medskip
    You are given two correlated images.\\
    1. Describe their shared visual or semantic theme in one short phrase.\\
    2. Then describe each image separately, but ensure that both captions naturally include that shared theme.\\[0.4em]
    Use this exact format strictly:\\
    \hspace*{1.5ex}\texttt{Shared theme:} [X]\\
    \hspace*{1.5ex}\texttt{Caption A:} [short description of image A including X]\\
    \hspace*{1.5ex}\texttt{Caption B:} [short description of image B including X]\\[0.3em]
    Avoid artistic or stylistic adjectives.
    
    \medskip
    \textbf{Output format.}\\
    \texttt{Shared theme:} [X]\\
    \texttt{Caption A:} [short description of image A including X]\\
    \texttt{Caption B:} [short description of image B including X]
    \end{tcolorbox}
}
\vspace{-0.3cm}
\end{center}

\subsection{Computational Cost, VLM Robustness, and Fairness of SAP}
\label{subsec:sap_cost}
 
\subsubsection{Computational cost and VLM dependency}
Because SAP queries a VLM, one might worry that it introduces an unfair or unpredictable overhead relative to VLM-free morphing methods. We clarify that this overhead is small and bounded. First, the comparison is on equal footing: the inversion- and tuning-based baselines already obtain their endpoint
prompts from a VLM~\cite{liu2024llavanext}, so SAP does not introduce a VLM where none existed it only adds construction of the anchor-correlated triplet and the \textit{cossim}-based reliability re-query introduced in the \emph{Suppl}. Second, this increment is light: as reported above, the re-query terminates in $1.62$ calls per pair ($0.62$ re-queries on average), adding only $+7.1$\,s and $+45.79$\,TFLOPs over Morph4Data and MorphBench ($n{=}167$). Including this cost, CHIMERA still requires only $19.5$\,s and $258.81$\,TFLOPs end-to-end and remains the most efficient method as shown in \cref{fig:teaser1}(c), below the tuning-based IMPUS and DiffMorpher. The dominant cost therefore remains the diffusion sampling shared by all methods.

\begin{table}
  \centering
  \caption{Robustness of SAP to the VLM. Morphing quality is stable across VLMs and
  exceeds all baselines (see \cref{table:main_experiment_frames_side_by_side}).
  GLCS is averaged on Morph4Data and MorphBench.}
  \label{tab:sap_vlm}
  \setlength{\tabcolsep}{4pt}
  \scriptsize
  \begin{tabular}{lc}
    \toprule
    VLM used in SAP & GLCS\,$\uparrow$ \\
    \midrule
    LLaVA                       & 88.274 \\
    InternVL2                   & 88.084 \\
    Qwen2.5-VL \textbf{(Ours)}  & 88.079 \\
    \bottomrule
  \end{tabular}
\end{table}
\subsubsection{Robustness to the VLM choice}
SAP does not hinge on a single, particularly capable VLM. As shown in \cref{tab:sap_vlm}, replacing Qwen2.5-VL~\cite{bai2025qwen2} with LLaVA~\cite{liu2024llavanext} or InternVL2~\cite{chen2024expanding} leaves morphing quality essentially unchanged: all three VLMs yield GLCS within $0.2$ of one another and exceed every baseline in \cref{table:main_experiment_frames_side_by_side}. This indicates that the improvement originates from the shared-anchor formulation itself rather than from a specific VLM.

\subsection{Further Analysis}\label{exp:further_analysis}
\begin{table}
    \centering
    \caption{Performance comparison across different diffusion backbones. 
    }
    \label{tab:backbone_comparison}
    \small
    \setlength{\tabcolsep}{5pt}
    \resizebox{\linewidth}{!}{%
    \begin{tabular}{l c c c c c}
        \toprule
        \textbf{Exp name} 
        & $\mathrm{FID}_{\text{local}}\downarrow$ 
        & $\mathrm{FID}_{\text{global}}\downarrow$ 
        & LPIPS$\downarrow$ 
        & PPL$\downarrow$ 
        & GLCS$\uparrow$ \\
        \midrule
        SD1.4 & 180.237 & 91.003 & 1.782 & 0.300 & 86.924 \\
        SD1.4 + \textbf{CHIMERA} & \textbf{176.673} & \textbf{90.593} & \textbf{1.699} & \textbf{0.283} & \textbf{87.311} \\
        SD1.5 & 174.803 & 89.842 & 1.724 & 0.297 & 86.929 \\
        SD1.5 + \textbf{CHIMERA} & \textbf{172.954} & \textbf{88.914} & \textbf{1.697} & \textbf{0.283} & \textbf{87.037} \\
        SD2 & 187.790 & 97.053 & 1.667 & 0.280 & 87.741 \\
        SD2 + \textbf{CHIMERA} & \textbf{185.239} & \textbf{95.104} & \textbf{1.659} & \textbf{0.277} & \textbf{88.001} \\
        \midrule
        SDXL & 209.974 & 101.715 & 2.218 & 0.370 & 84.809 \\
        SDXL + \textbf{CHIMERA} & \textbf{203.880} & \textbf{100.345} & \textbf{2.156} & \textbf{0.359} & \textbf{84.959} \\
        \midrule
        FLUX & 167.256 & 85.326 & 1.566 & 0.273 & 88.253 \\
        FLUX + \textbf{CHIMERA} & \textbf{157.780} & \textbf{80.826} & \textbf{1.552} & \textbf{0.259} & \textbf{89.983} \\
        \bottomrule
    \end{tabular}%
    }
\end{table}

\begin{figure}
    \centering
    \includegraphics[scale=0.55]{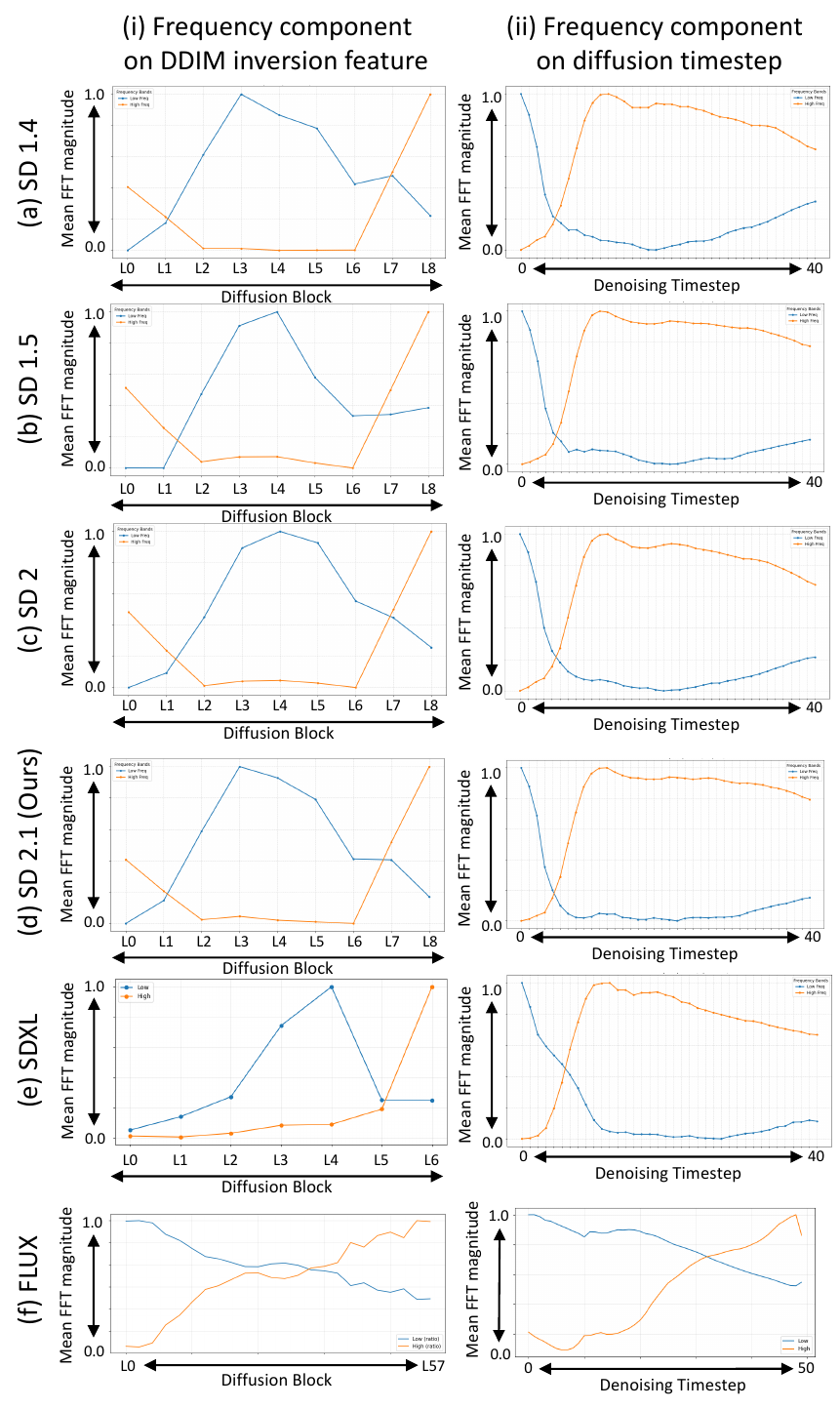}
    \caption{Frequency analysis on another backbone. (i) shows the layer-wise frequency characteristics of a given diffusion model, while (ii) presents the timestep-wise frequency characteristics. In each plot, the orange line represents high-frequency components and the blue line represents low-frequency components.}
    \label{fig:another_backbone_freq}
\end{figure}

\subsubsection{Efficiency}
To demonstrate the efficiency of CHIMERA, we report computational cost (runtime, TFLOPs, and learnable parameters) in \cref{fig:teaser1}(c). As shown in \cref{fig:teaser1}(c), CHIMERA requires significantly lower runtime and TFLOPs than tuning-based models~\cite{zhang2024diffmorpher,yang2024impus}. Notably, it is even more efficient than the training-free method~\cite{cao2025freemorph}.

\subsubsection{Another Backbone}
We provide additional analyses for the experiments presented in \cref{Sec:PreNObs}. 
In \cref{Sec:PreNObs}, we analyzed diffusion features from a frequency perspective and presented observations that motivate adaptive feature injection based on layer-wise and timestep-wise characteristics. 
Here, we extend the same frequency-based analysis to other diffusion models to demonstrate that CHIMERA can generalize to different architectures.

The analysis in \cref{fig:freq_analysis} focuses on layer- and timestep-wise characteristics of SD~2.1~\cite{rombach2022high}. 
In this section, we extend the analysis to other U-Net-based diffusion models, including SD~1.4, SD~1.5, and SD~2, as well as larger models such as SDXL~\cite{podell2023sdxl} and the DiT-based model FLUX~\cite{peebles2023scalable,labs2025flux}. 
As shown in \cref{fig:another_backbone_freq}(a)-(c), U-Net-based diffusion models share very similar layer structures and therefore exhibit layer- and timestep-wise characteristics that are largely consistent with those observed in SD~2.1 (\cref{fig:another_backbone_freq}(d)). 
Based on this observation, applying CHIMERA leads to consistent performance improvements across these models, as reported in \cref{tab:backbone_comparison}.

As shown in \cref{fig:another_backbone_freq}(e), SDXL exhibits frequency characteristics similar to those of SD~2.1, although the number of layers differs from that of SD~2.1. 
To further demonstrate that CHIMERA is not limited to U-Net-based architectures, we extend the analysis to the diffusion transformer model FLUX (\cref{fig:another_backbone_freq}(f)). 
As noted in prior DiT-based studies~\cite{chang2025sparsedit, yu2024representation}, early layers of DiT mainly capture low-frequency components, while later layers focus on high-frequency details; similarly, early timesteps tend to generate low-frequency structures, while later timesteps refine high-frequency details. 
These properties are also reflected in \cref{fig:another_backbone_freq}(f). Based on these observations, applying CHIMERA to both SDXL and FLUX results in consistent performance improvements, as shown in \cref{tab:backbone_comparison}. 
These results demonstrate that CHIMERA is not limited to a specific backbone and can be effectively extended to a wide range of diffusion-based models.

\subsubsection{Backbone-agnostic Applicability}
LTM constructs the layer- and timestep-wise correspondence from the frequency characteristics of each layer group rather than from absolute layer positions, so the correspondence is derived from the backbone itself rather than specified by hand. The same procedure therefore transfers directly to a new backbone. We apply LTM, which accounts for layer- and timestep-wise characteristics, to the per-backbone frequency analysis presented in \cref{exp:further_analysis}, obtaining the correspondence for each backbone in the same manner. Accordingly, the ACI and SAP configurations are kept identical to the SD~2.1 setting across every backbone we evaluate, and we follow only the sampler native to each model. Under this single shared configuration, \cref{tab:backbone_comparison} shows that CHIMERA yields consistent improvements across both U-Net-based and DiT-based models.
\section{Conclusion}

We have presented CHIMERA, a zero-shot diffusion-based framework for smooth, semantically coherent, and domain-consistent image morphing. CHIMERA combines Adaptive Cache Injection (ACI) and Semantic Anchor Prompting (SAP) to guide denoising with multi-scale inversion features and VLM-derived anchors, alleviating over-smoothing, over-saturation, and semantic inconsistency in prior methods. 
We also introduced GLCS, a morphing-oriented metric that correlates with human judgment. 
Experiments and a user study show consistent improvements over existing approaches, establishing strong performance for zero-shot diffusion morphing.
\section{Acknowledgment}
The authors used Claude Code (Claude Opus 5, Anthropic) to assist with debugging and correcting code errors in the implementation described in Section IV, and with configuring the execution environments of the comparison models in Section V-TABLE II, VII. The tool was used only to support debugging and environment setup for author-written code.





\bibliographystyle{IEEEtran}
\bibliography{ref}


\setcounter{page}{1}
\setcounter{section}{0}
\setcounter{subsection}{0}
\setcounter{subsubsection}{0}
\setcounter{figure}{0}
\setcounter{table}{0}
\setcounter{equation}{0}

\renewcommand{\thepage}{S\arabic{page}}
\renewcommand{\thesection}{S\arabic{section}}
\renewcommand{\thesubsection}{\thesection-\Alph{subsection}}
\renewcommand{\thesubsubsection}{\thesubsection.\arabic{subsubsection}}

\renewcommand{\thefigure}{S\arabic{figure}}
\renewcommand{\thetable}{S\arabic{table}}
\renewcommand{\theequation}{S\arabic{equation}}


\twocolumn[
\begin{center}

{\fontsize{24}{28}\selectfont
Supplementary Material for\\[5pt]
CHIMERA: Adaptive Cache Injection and\\
Semantic Anchor Prompting for Zero-shot\\
Image Morphing with Morphing-oriented Metrics
\par}

\vspace{1.25em}

{\fontsize{12}{14}\selectfont
Dahyeon~Kye$^{*}$,
Jeahun~Sung$^{*}$,
Minkyu~Jeon,
and~Jihyong~Oh$^{\dagger}$
\par}

\vspace{1.8em}

\end{center}
]


\section*{Supplementary Contents}
\begin{itemize}
    \item \cref{sp_sec:idm_ablation}~Ablation Study on Inversion-Denoising Timestep Mapping (IDM)
    \item \cref{sec:sap_supp}~Detailed Analyses of SAP
    \item \cref{sec:hetero_pairs}~Generalization Experiments on Heterogeneous Pairs
    \item \cref{sp_sec:eval_met_trad}~Evaluation Metric
    \item \cref{sp_sec:glcs_further_analysis}~Further Analysis on GLCS
    \item \cref{sec:userstudy_supp}~User Study: Subjective Preference Analysis
    \item \cref{sp_sec:application}~Application
    \item \cref{sp_sec:extend_exp}~Extended Experiment Results
    \item \cref{sp_sec:imple_detail}~Additional Implementation Detail
    \item \cref{sec:add-qual}~Additional Qualitative Result
    \item \cref{sec:limitation}~Limitations and Failure Cases
\end{itemize}

\section{Ablation Study on Inversion-Denoising Timestep Mapping (IDM)}
\label{sp_sec:idm_ablation}

\begin{table}[h]
\centering
\caption{Quantitative evaluation with respect to the IDM. We fix the inversion timesteps while performing injections at multiple denoising timesteps.}
\label{table:sp_idm_1}

\resizebox{0.9\columnwidth}{!}{%
\begin{tabular}{l c c c c c}
\toprule
\multicolumn{6}{c}{IDM Ablation - Fixed Inversion Timestep} \\
\midrule
 & $\mathrm{FID}_{\text{local}}\downarrow$ 
 & $\mathrm{FID}_{\text{global}}\downarrow$ 
 & LPIPS$\downarrow$ 
 & PPL$\downarrow$ 
 & GLCS$\uparrow$ \\
\midrule
(a) \textbf{Ours}
    & \textcolor{red}{\textbf{161.054}} 
    & \textcolor{red}{\textbf{82.308}} 
    & \textcolor{red}{\textbf{1.576}}
    & \textcolor{red}{\textbf{0.263}}
    & \textcolor{red}{\textbf{88.079}} \\
(b) Early
    & \textcolor{blue}{\underline{186.030}} 
    & \textcolor{blue}{\underline{93.002}} 
    & 1.813 
    & 0.300 
    & 87.784 \\
(c) Mid
    & 196.437 
    & 97.327 
    & 1.764
    & 0.303
    & \textcolor{blue}{\underline{87.929}} \\
(d) Late
    & 205.031
    & 101.490
    & \textcolor{blue}{\underline{1.726}} 
    & \textcolor{blue}{\underline{0.295}} 
    & 87.124 \\
\bottomrule
\end{tabular}%
}
\vspace{-0.2cm}
\end{table}

In this section, we present additional experimental results on the effectiveness of the Inversion-Denoising Timestep Mapping (IDM) described in \cref{pm:aci}. To validate the benefit of IDM, we compare the case where the mapping function is used (Ours) with the case where it is not used, and we report both qualitative and quantitative results. We divide the non-mapping cases into two configurations:  
(i) the inversion timesteps are fixed to early, mid, or late regions, and the denoising process injects the corresponding fixed cached layers for each timestep;  
(ii) the inversion timesteps are extracted at all timesteps as in the original setting, but the denoising process injects the cached features only within one fixed region (early, mid, late).  
For clarity, we unify the interpretation of early, mid, and late as follows: early denotes the state with the least injected noise, mid denotes a medium noise level, and late denotes the highest noise level (although, in practice, early denoising timesteps have high noise levels, while late timesteps have low noise levels).

\begin{figure}[!t]
    \centering
    \includegraphics[width=\linewidth]{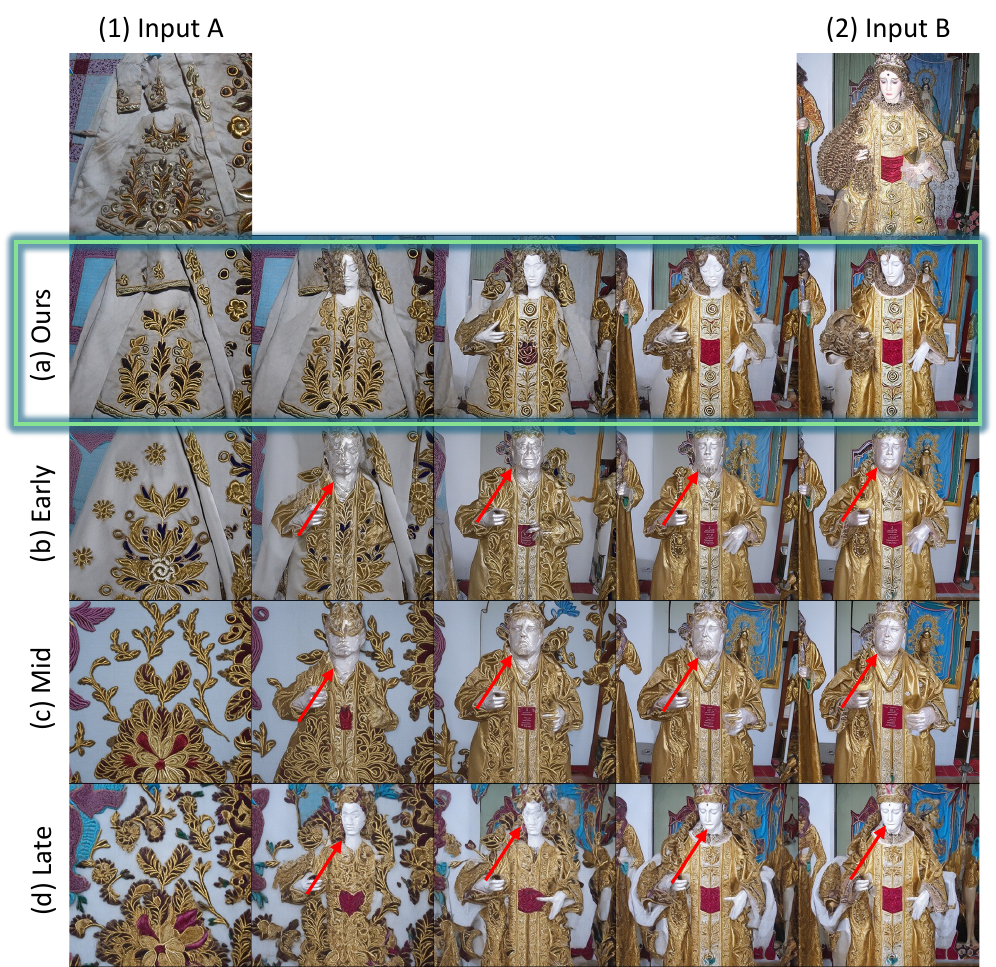}
    \caption{Qualitative results when the inversion timesteps are fixed. Panels (b) Early, (c) Mid, and (d) Late correspond to states with high noise, medium noise, and very low noise, respectively. Panel (a) represents our model with the IDM applied.}
    \label{ab:idm1}
\end{figure}

\begin{table}[!t]
\centering
\caption{Quantitative evaluation with respect to the IDM.
We fix the denoising timesteps while extracting multiple inversion timesteps.}
\label{table:sp_idm_2}

\resizebox{0.9\columnwidth}{!}{%
\begin{tabular}{l c c c c c}
\toprule
\multicolumn{6}{c}{IDM Ablation - Fixed Denoising Timestep} \\
\midrule
 & $\mathrm{FID}_{\text{local}}\downarrow$ 
 & $\mathrm{FID}_{\text{global}}\downarrow$ 
 & LPIPS$\downarrow$ 
 & PPL$\downarrow$ 
 & GLCS$\uparrow$ \\
\midrule
(a) \textbf{Ours}
    & \textcolor{red}{\textbf{161.054}} 
    & \textcolor{red}{\textbf{82.308}} 
    & \textcolor{red}{\textbf{1.576}} 
    & \textcolor{red}{\textbf{0.263}} 
    & \textcolor{red}{\textbf{88.079}} \\
(b) Early
    & \textcolor{blue}{\underline{195.827}} 
    & \textcolor{blue}{\underline{99.192}} 
    & 1.858 
    & 0.310 
    & \textcolor{blue}{\underline{86.036}} \\
(c) Mid
    & 204.923 
    & 102.100 
    & 1.691
    & 0.271 
    & 85.306 \\
(d) Late
    & 206.036 
    & 102.629 
    & \textcolor{blue}{\underline{1.690}} 
    & \textcolor{blue}{\underline{0.269}} 
    & 85.176 \\
\bottomrule
\end{tabular}%
}
\vspace{-0.2cm}
\end{table}

\begin{figure}[!t]
    \centering
    \includegraphics[width=\linewidth]{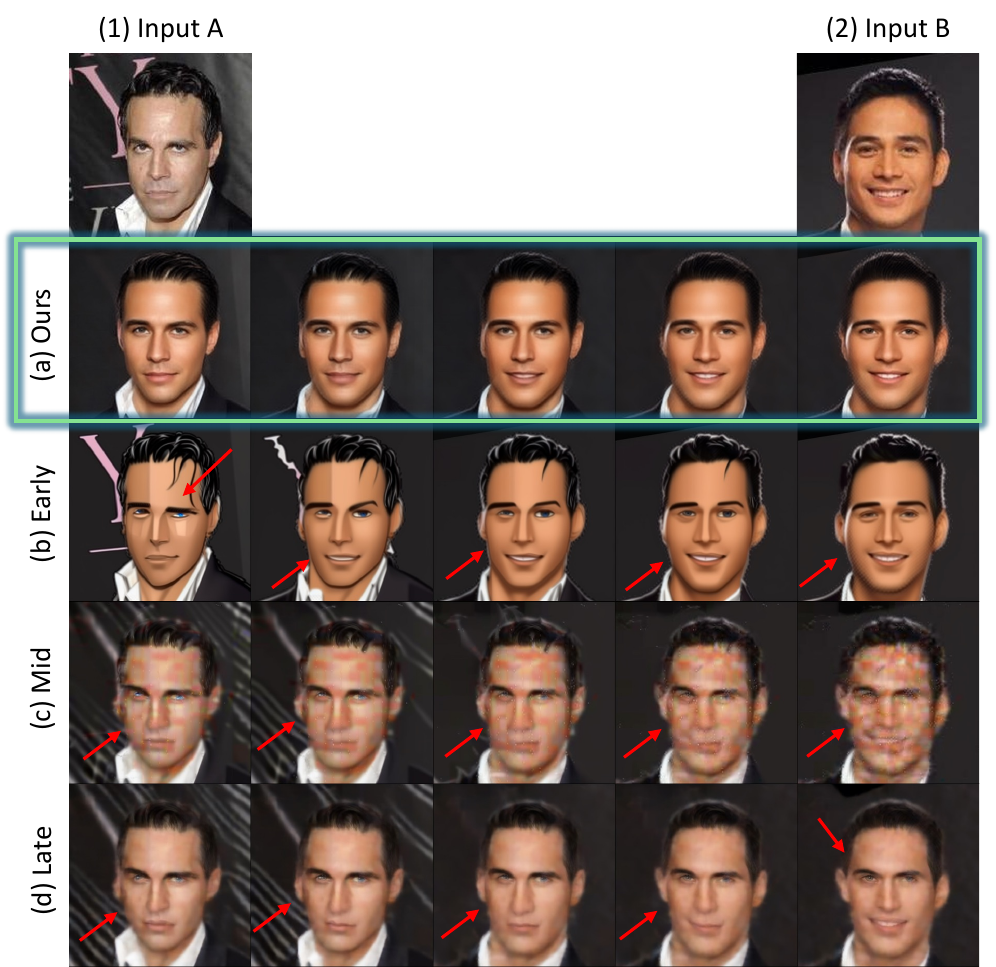}
    \caption{Qualitative results when the denoising timesteps are fixed. Panels (b) Early, (c) Mid, and (d) Late correspond to states with high noise, medium noise, and no noise, respectively. Panel (a) represents our model with the IDM applied.}
    \label{ab:idm2}
    \vspace{-0.2cm}
\end{figure}

When the inversion timesteps are fixed, \cref{table:sp_idm_1} shows that our IDM-based model (a) achieves the best quantitative performance. As shown in \cref{ab:idm1}, fixing inversion to early, mid, or late produces undesired results: the model generates images that deviate from the input images \(A\) and \(B\) (\cref{ab:idm1}(b), (c)), or produces structurally unstable results with severe artifacts (\cref{ab:idm1}(d)). In each case, the red arrows in the figure explicitly indicate the regions where these degradations occur.

When the denoising timesteps are fixed, \cref{table:sp_idm_2} again shows that the IDM-based model (a) provides the best overall quantitative performance. As shown in \cref{ab:idm2}, injecting cached features only at early, mid, or late denoising timesteps leads to several issues: overly saturated images (\cref{ab:idm2}(b)) or images that are noticeably blurry or noisy (\cref{ab:idm2}(c), (d)). In these cases, the red arrows explicitly indicate the regions corresponding to the undesired artifacts and noise.
\newcommand{\best}[1]{\textbf{\textcolor{red}{#1}}}
\newcommand{\second}[1]{\underline{\textcolor{blue}{#1}}}

\section{Detailed Analyses of SAP}
\label{sec:sap_supp}

\subsection{VLM Instruction Strategy}
\begin{figure*}
    \centering
    \includegraphics[width=\textwidth]{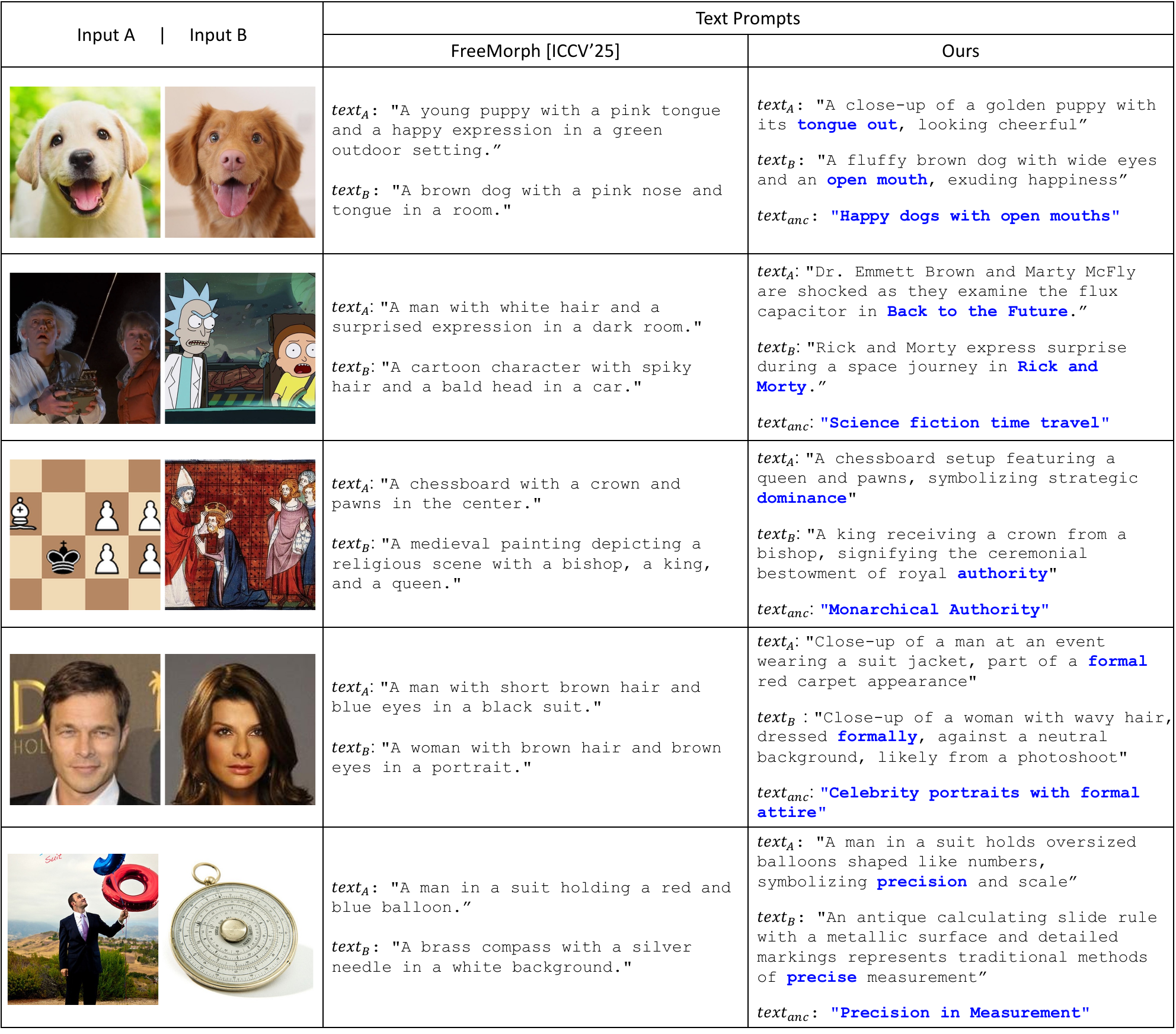}
    \caption{Examples of VLM-generated anchor-correlated prompt triplet. Given two endpoint images, VLM produces a shared anchor-prompt (\(text_{anc}\)) and the anchor-conditioned endpoint prompts (\(text_A\), \(text_B\)).}
    \label{fig:prompts} 
\end{figure*}

Following the SAP formulation in \cref{sec:sap}, given two images we query Qwen2.5-VL~\cite{bai2025qwen2} with a structured instruction that first asks for a short shared concept (\textit{anchor}) and then asks for endpoint prompts that naturally preserve that shared concept. This anchor-correlated prompt triplet $(text_{\mathrm{anc}}, text_A, text_B)$ formulation encourages all prompts to remain mutually correlated, allowing SAP to provide stable semantic guidance during denoising. As shown in \cref{fig:prompts}, the resulting prompt triplet is mutually correlated rather than independently formed. In many cases, this correlation appears through explicit keyword sharing, while in other cases it emerges through higher-level semantic compatibility. This property is important for SAP, since the shared anchor-prompt is intended to guide the denoising process toward semantically coherent intermediate states rather than introducing incompatible endpoint-specific semantics.

\subsection{Anchor Reliability}
\label{subsec:sap_reliab}
SAP relies on the assumption that the anchor-prompt $text_{\mathrm{anc}}$ captures a concept shared by both endpoints.
We quantify this using the cosine similarity (\textit{cossim}) between the anchor-prompt and each endpoint prompt, namely $\textit{cossim}(text_{\mathrm{anc}}, text_A)$ and $\textit{cossim}(text_{\mathrm{anc}}, text_B)$.
As shown in \cref{tab:anchor_similarity}, shared anchors consistently yield higher similarities to both endpoint prompts than deliberately unshared anchors obtained by reversing the correspondence.
This observation suggests that \textit{cossim} serves as a practical proxy for anchor reliability, as semantically relevant anchors remain better aligned with both endpoint prompts.

Motivated by this, we adopt a \textit{cossim}-based anchor-reliability criterion. After generating the triplet, we encode all prompts with a CLIP text
encoder and compute the similarity between the anchor-prompt and each endpoint prompt; if either falls below a predefined threshold ($0.45$ in all experiments),
we re-query the VLM until both similarities satisfy the criterion (\cref{alg:chimera_updated_compact}, lines~7--11). In practice, this loop terminates after only a few iterations:
averaged over Morph4Data~\cite{cao2025freemorph} and MorphBench~\cite{zhang2024diffmorpher} ($n{=}167$ pairs), SAP issues $1.62$ VLM~\cite{bai2025qwen2} calls per pair, i.e., only $0.62$ re-queries on average. The criterion is thus
met after a small, bounded number of re-queries rather than an unbounded search, so constructing reliable anchors does not introduce unpredictable latency. This
validation step ensures that SAP operates on reliable shared anchors and strongly correlated prompt triplets.

\begin{figure*}[t]
    \centerline{\includegraphics[width=\linewidth]{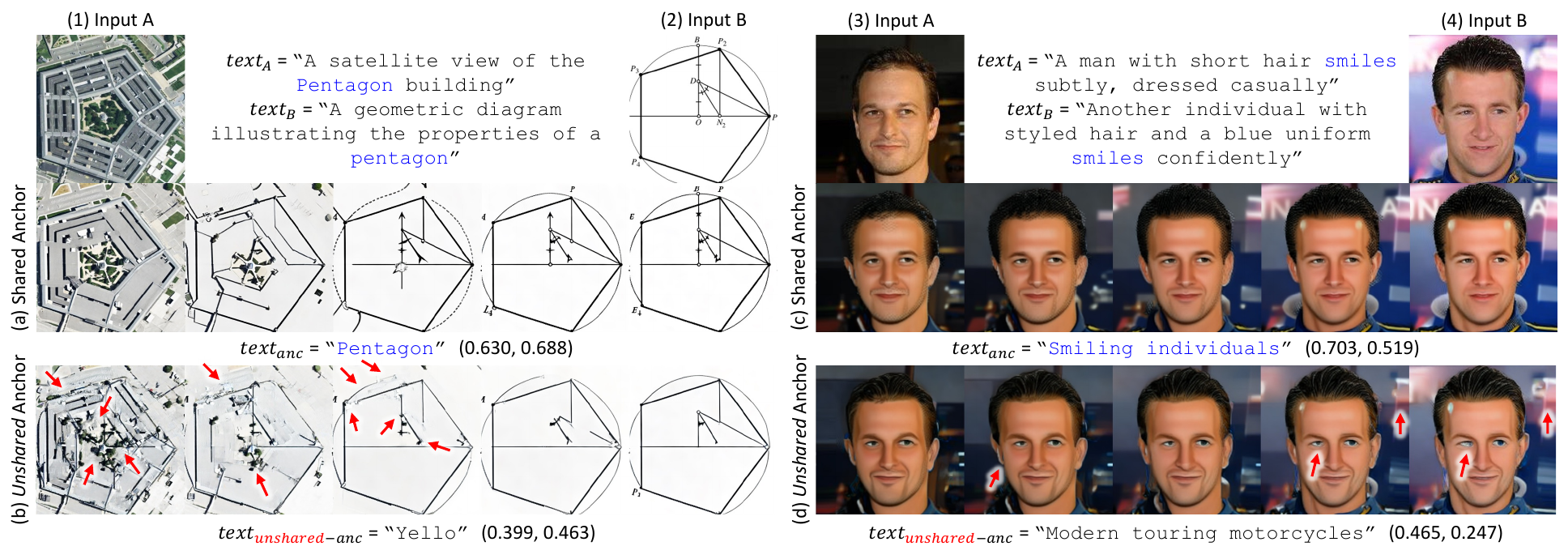}}
    \caption{Additional qualitative comparisons of shared and unshared anchors. For each input pair, the upper row ((a), (c)) uses a shared anchor-prompt, while the lower row ((b), (d)) uses an intentionally unshared anchor-prompt.
    The numbers in parentheses denote the cosine similarity between the anchor-prompt and each endpoint prompt, respectively.
    Using shared anchors produces smoother morphing results, whereas using unshared anchors leads to distorted structures or localized artifacts.}
    \vspace{-0.5cm}
    \label{fig:sap_more_cases}
\end{figure*}

\subsection{Additional Qualitative Comparisons of Shared vs. Unshared Anchor-prompts}
We provide additional qualitative comparisons of shared and intentionally unshared anchors in \cref{fig:sap_more_cases}.
Consistent with the observation in \cref{fig:anchor_glcs}(a), shared anchors produce more stable and plausible morphing results, whereas unshared anchors often introduce artifacts.
In particular, the red arrows highlight representative failure regions under unshared anchors, such as broken geometric structures, spurious local patterns, and facial artifacts.
These examples show how unshared anchor guidance can disrupt the semantic compatibility of the morphing even when the endpoint prompts remain individually plausible.
This observation is also consistent with the \textit{cossim} analysis.
In particular, unshared anchors with low \textit{cossim} tend to produce lower-quality morphing results.
Together, these results support our use of \textit{cossim}-based anchor-reliability criterion to filter out unshared anchors or weakly correlated prompt triplets before constructing the text prompts used by SAP.

\begin{figure}[t]
    \centerline{\includegraphics[width=\columnwidth]{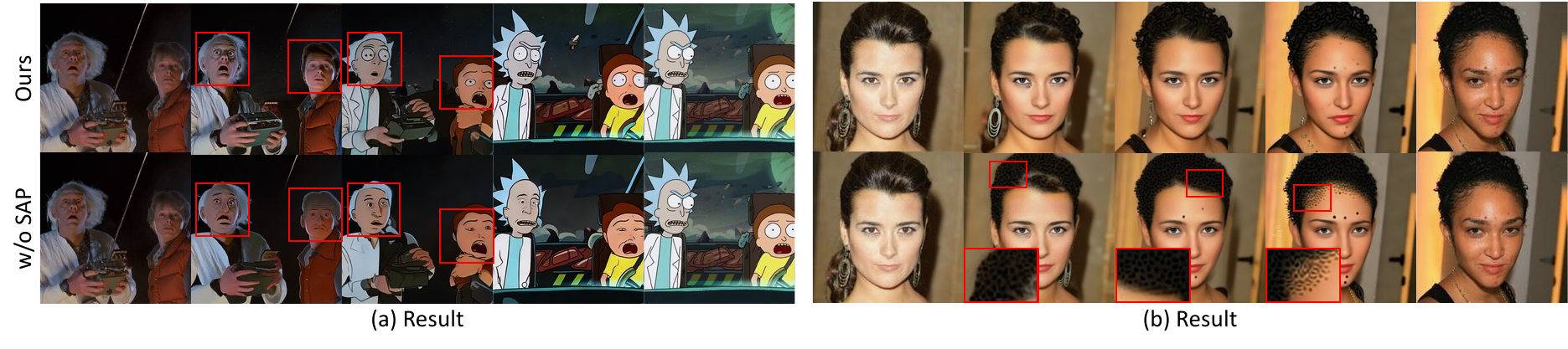}}
    \caption{Qualitative effect of SAP. Morphing results with SAP (top row) and without SAP (bottom row) for the same input pairs. Without SAP, the intermediate images exhibit distorted faces and unnatural dotted-hair artifacts; enabling SAP removes these artifacts and yields more plausible, semantically coherent transitions.}
    \label{fig:wo_sap}
\end{figure}

\begin{table}[t]
    \centering
        \caption{Ablation on SAP activation timestep. We compare different SAP schedules, including disabling SAP, applying it only in the early stage (\texttt{stage1}), only in the later stage (\texttt{stage2}), and throughout both stages. Activating SAP only in the early stage yields the best overall balance across metrics.}
    \resizebox{\linewidth}{!}{
    \begin{tabular}{l c c c c c}
        \toprule
        Method & FID$_{\text{local}}\downarrow$ & FID$_{\text{global}}\downarrow$ & LPIPS$\downarrow$ & PPL$\downarrow$ & GLCS$\uparrow$ \\
        \midrule
        
        (a) \texttt{w/o SAP} 
         & \textcolor{blue}{\underline{163.521}}
         & \textcolor{blue}{\underline{84.669}}
         & 1.597
         & 0.270
         & \textcolor{blue}{\underline{87.811}} \\
        
        (b) \texttt{stage1} \textbf{(Ours)} 
         & \textcolor{red}{\textbf{161.054}}
         & \textcolor{red}{\textbf{82.308}}
         & 1.576
         & 0.263
         & \textcolor{red}{\textbf{88.079}} \\
        
        (c) \texttt{stage2} 
         & 193.621
         & 98.564
         & \textcolor{red}{\textbf{1.516}}
         & \textcolor{red}{\textbf{0.253}}
         & 87.624 \\
        
        (d) \texttt{stage1+stage2} 
         & 207.917
         & 105.107
         & \textcolor{blue}{\underline{1.571}}
         & \textcolor{blue}{\underline{0.262}}
         &  87.191 \\
        \bottomrule
    \end{tabular}
    }
    \label{table:SAP_ab_timestep}
    \vspace{-0.2cm}
\end{table}

\subsection{SAP Activation Schedule}

\cref{table:SAP_ab_timestep} shows that restricting SAP to the early stage achieves the best overall performance.
Following FreeMorph~\cite{cao2025freemorph}, we divide the denoising timestep into two stages and set the SAP activation schedule accordingly.
We define \texttt{stage1} as the early timesteps up to 0.2 of the full denoising timestep and \texttt{stage2} as the subsequent timesteps from 0.2 to 0.6.
Applying SAP only in \texttt{stage1} achieves the best results across all metrics, consistently improving over \texttt{w/o SAP}. As shown in \cref{fig:wo_sap} the qualitative effect of removing SAP is equally clear, the \texttt{w/o SAP} variant produces distorted faces and unnatural dotted-hair artifacts in the intermediate frames, whereas enabling SAP yields more plausible and semantically coherent transitions mirroring the quantitative gap between rows~(a) and~(b) of \cref{table:SAP_ab_timestep}.
In contrast, enabling SAP only in the later stage, or throughout both stages, degrades the overall balance across metrics.
These results suggest that SAP is most effective when it provides semantic guidance at the early denoising stage, 
where the diffusion model establishes the coarse global structure and semantic layout, whereas applying it later can over-constrain the generation process
and harm overall morphing quality~\cite{liu2024drag, pardo2025matchdiffusion, kim2025early}.
We therefore adopt an early-stage-only SAP schedule throughout all experiments.

\begin{table}[t]
  \centering
  \caption{GLCS on 20 heterogeneous pairs. Each cell reports the mean/std over the 20 evaluation pairs. CHIMERA achieves both the highest mean GLCS and the lowest variance, indicating consistent and stable morphing across heterogeneous inputs. Best in \textcolor{red}{red}, second best in \textcolor{blue}{blue}.}
  \label{tab:gen_qual}

  \setlength{\tabcolsep}{10pt}
  \renewcommand{\arraystretch}{1.2}

  \resizebox{\linewidth}{!}{%
  \begin{tabular}{l cccc}
    \toprule
     & IMPUS & DiffMorpher & FreeMorph & CHIMERA \\
    \midrule
    GLCS (mean/std)$\uparrow$
      & 79.30\,/\,6.68
      & 80.54\,/\,6.57
      & \textcolor{blue}{\underline{81.93\,/\,8.77}}
      & \textcolor{red}{\textbf{87.34\,/\,5.35}} \\
    \bottomrule
  \end{tabular}%
  }
\end{table}

\begin{figure*}[t]
  \centering
  \includegraphics[width=\linewidth]{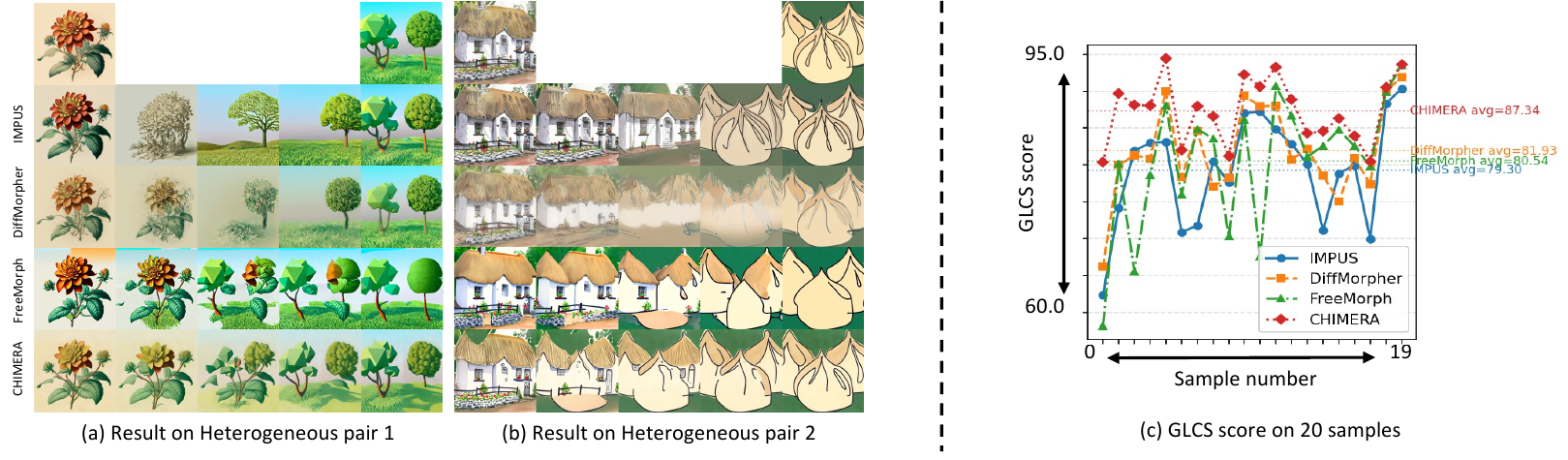}
  \caption{Generalization across sample pairs. (a),(b) Qualitative morphing results on two heterogeneous input pairs. (c) Per-sample GLCS scores over the 20 evaluation pairs; CHIMERA (red)
  consistently ranks near the top with the smallest variance across the baselines.}
  \label{fig:gen_qual}
  \vspace{-0.2cm}
\end{figure*}


\section{Generalization Experiments on Heterogeneous Pairs}
\label{sec:hetero_pairs}
Inference-based does not imply uncontrollable. We additionally evaluate on
20 pairs from 40 publicly available images, providing fully unseen inputs
including extreme heterogeneous cases with minimal semantic or structural
overlap. To collect these evaluation pairs, we prompted Qwen3~\cite{yang2025qwen3}
to generate a set of mutually diverse text prompts, and used the generated
prompts as search queries to gather 20 copyright-free image pairs from web,
ensuring that the evaluation inputs span a wide range of semantic and structural
combinations rather than being biased toward a particular domain.
As shown in \cref{tab:gen_qual}, CHIMERA achieves both the highest mean GLCS
and the lowest standard deviation. Moreover, the per-sample plot in
\cref{fig:gen_qual}(c) shows that CHIMERA consistently ranks near the top
across every pair with substantially smaller variance than the baselines.
The qualitative comparisons in \cref{fig:gen_qual}(a), (b) further confirm that
CHIMERA produces stable and coherent morphing results.

\section{Evaluation Metric}
\label{sp_sec:eval_met_trad}

This section provides detailed explanations of the metrics introduced in \cref{sec:experiment}. The motivation, significance, and limitations of these metrics are further discussed in \cref{pm:glcs}.

\subsection{Fréchet Inception Distance (FID)-Based Metrics}

\subsubsection{Local FID}
We use a local variant, \(\mathrm{FID}_{\text{local}}\), to measure distribution gaps between the input image pair \(\{A,B\}\) and the morphing images \(\{I_k\}_{k=1}^{K}\) on a per-pair basis. For an image pair \(j\), the input images \(\{A_j,B_j\}\) serve as the real domain, and the morphing images \(\{I^{(j)}_k\}_{k=1}^{K_j}\) serve as the generated domain. Let

\begin{equation}
X^{(j)}_{\text{real}} = \{f(A_j), f(B_j)\}, \qquad
X^{(j)}_{\text{gen}} = \{f(I^{(j)}_k)\}_{k=1}^{K_j}.
\end{equation}
The local FID for pair \(j\) is defined as:

\begin{equation}
\mathrm{FID}_{\text{local}}^{(j)}
= \mathrm{FID}\big(\{A_j,B_j\}, \{I^{(j)}_k\}_{k=1}^{K_j}\big),
\end{equation}
which measures how well the morphing images align with the endpoint distribution for each pair. At the dataset level, we compute:

\begin{equation}
\overline{\mathrm{FID}}_{\text{local}}
= \frac{1}{N}\sum_{j=1}^{N} \mathrm{FID}_{\text{local}}^{(j)},
\end{equation}

to summarize pair-wise domain consistency.

\subsubsection{Global FID}
In contrast, \(\mathrm{FID}_{\text{global}}\) evaluates the distribution gap at the dataset level. Let

\begin{equation}
\mathcal{X}_{\text{real}}
= \bigcup_{j=1}^{N} \{A_j,B_j\}, \qquad
\mathcal{X}_{\text{gen}}
= \bigcup_{j=1}^{N} \{I^{(j)}_k\}_{k=1}^{K_j}.
\end{equation}

We estimate the mean and covariance of each set and apply the standard FID formula:

\begin{equation}
\mathrm{FID}_{\text{global}}
= \mathrm{FID}\Big(
\bigcup_{j=1}^{N} \{A_j,B_j\},
\bigcup_{j=1}^{N} \{I^{(j)}_k\}_{k=1}^{K_j}
\Big),
\end{equation}

Thus, \(\mathrm{FID}_{\text{local}}\) measures pair-wise domain alignment, while \(\mathrm{FID}_{\text{global}}\) captures how well the model preserves the distribution of the input images at the dataset level.

\subsubsection{Learned Perceptual Image Patch Similarity (LPIPS)-Based Metrics}

For each image pair \(j\), we define an ordered path

\begin{equation}
J^{(j)}_0 = A_j,\quad
J^{(j)}_k = I^{(j)}_k\ (k=1,\dots,K_j),\quad
J^{(j)}_{K_j+1} = B_j.
\end{equation}

We compute pairwise LPIPS distances using \(L(\cdot,\cdot)\):

\begin{equation}
d^{(j)}_n
= L\big(J^{(j)}_{n-1}, J^{(j)}_{n}\big),
\qquad n = 1,\dots,K_j+1.
\end{equation}
The path-based LPIPS metric is then defined as:

\begin{equation}
\mathrm{LPIPS}^{(j)}
= \sum_{n=1}^{K_j+1} d^{(j)}_n,
\end{equation}
and its dataset-level average is

\begin{equation}
\overline{\mathrm{LPIPS}}
= \frac{1}{N}\sum_{j=1}^{N} \mathrm{LPIPS}^{(j)}.
\end{equation}

\subsubsection{Perceptual Path Length (PPL)}
The Perceptual Path Length (PPL)~\cite{karras2020analyzing} measures the smoothness of the generator mapping by quantifying how sensitively the generated image changes under small perturbations in the latent space. Given a generator \( g: \mathcal{W} \to \mathcal{Y} \) and two nearby latent codes \( \mathbf{w}, \mathbf{w}' \in \mathcal{W} \) sampled along a linear interpolation, the PPL is defined as the expected perceptual distance between the corresponding images, normalized by the squared step size in latent space:
\[
\mathrm{PPL} = \mathbb{E}_{\mathbf{w},\,\mathbf{w}'} \left[ \frac{ d_{\mathrm{LPIPS}}\big( g(\mathbf{w}),\, g(\mathbf{w}') \big) }{ \| \mathbf{w} - \mathbf{w}' \|_2^2 } \right],
\]
where \( d_{\mathrm{LPIPS}}(\cdot,\cdot) \) denotes the LPIPS perceptual distance computed in a deep feature space. This metric approximates the local curvature of the generator manifold, and lower PPL values indicate a smoother, more semantically consistent latent-to-image mapping.

\begin{figure*}[!t]
    \centering
    \includegraphics[width=0.8\linewidth]{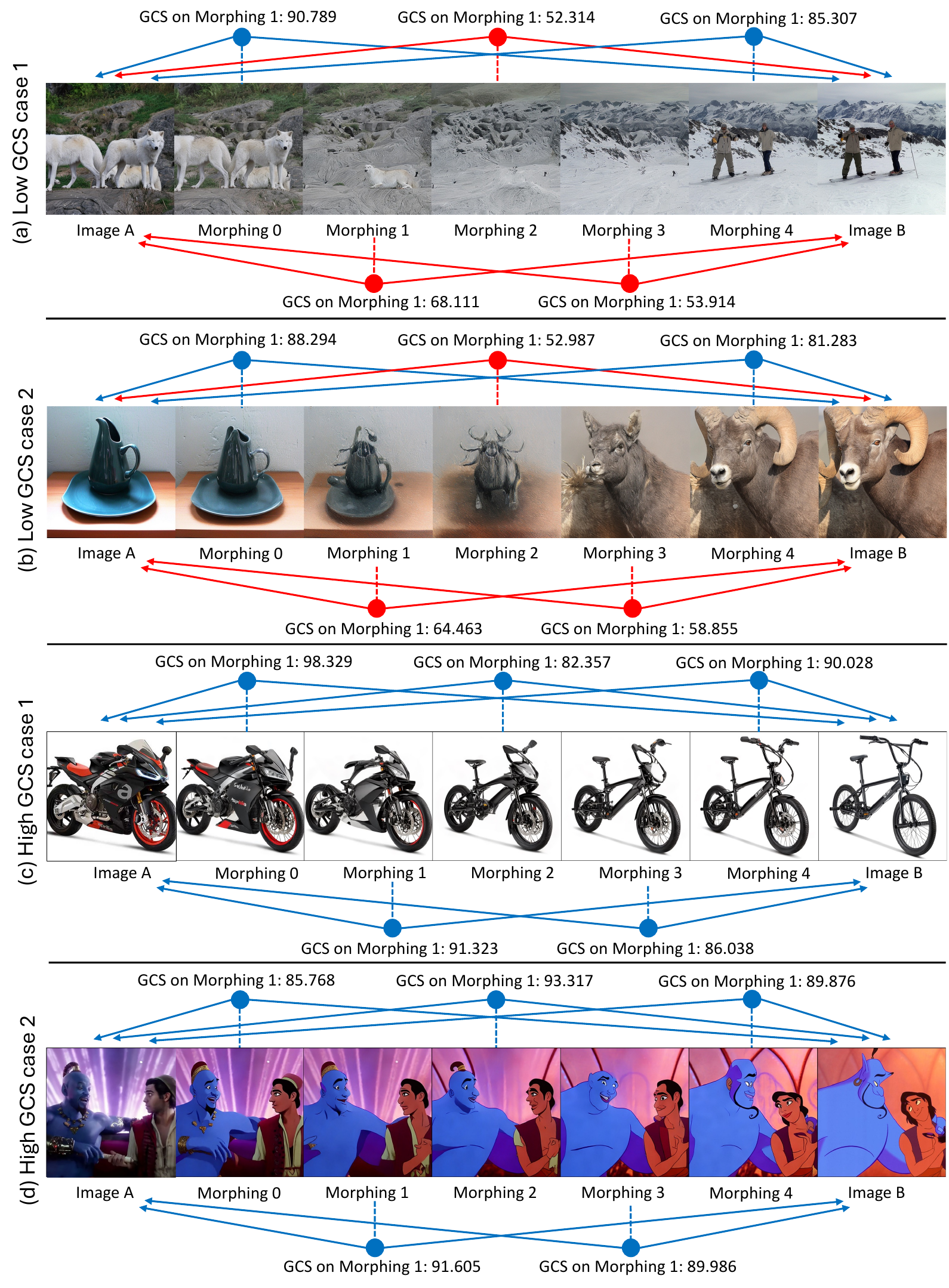}
    \caption{Qualitative examples showing how the GCS, a component of GLCS, aligns with human perception. Blue arrows indicate images where the domains of \(A\) and \(B\) are properly mixed according to the perceived interpolation ratio, while red arrows indicate images where the two domain cues are not well reflected given the same interpolation ratio.}
    \label{fig:effect_of_gcs}
\end{figure*}

\begin{figure*}[!t]
    \centering
    \includegraphics[width=0.8\textwidth]{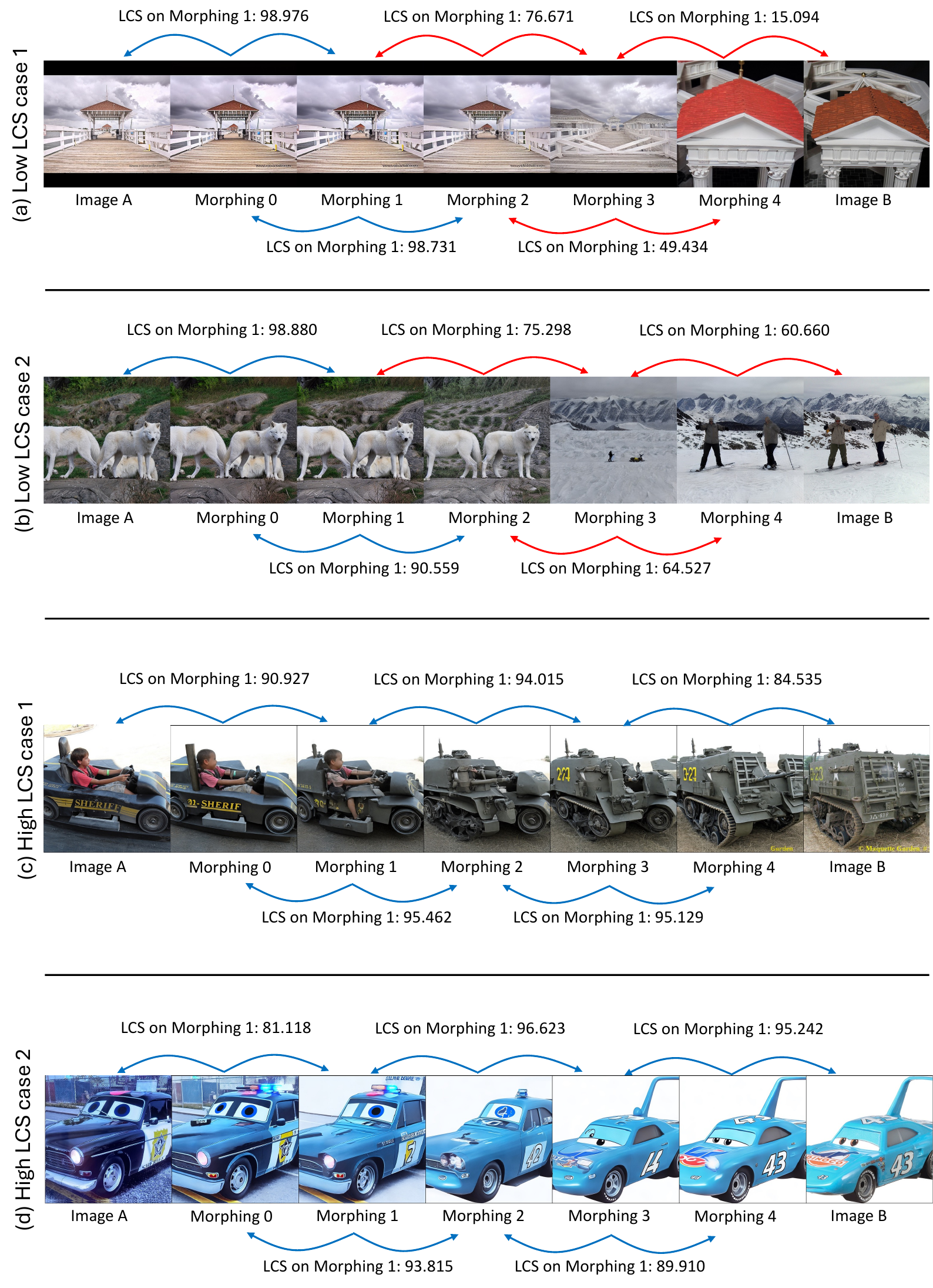}
    \caption{Qualitative examples showing how the LCS, a component of GLCS, aligns with human perception. Blue arrows indicate cases that are judged as similar by human observers, while red arrows indicate cases with abrupt perceptual changes.}
    \label{fig:effect_of_lcs}
\end{figure*}

\section{Further Analysis on GLCS}
\label{sp_sec:glcs_further_analysis}

\subsection{Effects of GCS and LCS}
\label{sup:eff_gcs_lcs}

\cref{fig:effect_of_gcs} reports the effect of GCS on selected morphing images. In \cref{fig:effect_of_gcs}(a) and (b), the red lines and dots indicate cases with low GCS scores, while the blue lines and dots indicate cases with high GCS scores. In \cref{fig:effect_of_gcs}(a), the Morphing-0 image is highly similar to image \(A\) and also shares a similar background with Morphing-1, resulting in a high GCS score of 90.789. In contrast, Morphing-1 should strongly reflect the wolf and moderately reflect the human from image \(A\), but it fails to do so, leading to a low GCS score. Moreover, Morphing-2 does not properly reflect either the wolf or the human, and thus shows the lowest score among the morphing images (\cref{fig:effect_of_gcs}(b) shows a similar case). Unlike (a) and (b), panels (c) and (d) exhibit consistently high GCS values across the morphing sequence, and human observers also perceive strong domain consistency that includes both domains of \(A\) and \(B\). This indicates that (c) and (d) have higher domain consistency than (a) and (b). These results demonstrate that the proposed GCS can evaluate domain consistency in a manner that aligns well with human perception.

\cref{fig:effect_of_lcs} reports the effect of LCS on selected morphing images. In \cref{fig:effect_of_lcs}(a) and (b), the red arrows indicate images with low perceptual smoothness, while the blue lines indicate images with high perceptual smoothness. We observe that the LCS score decreases as the difference between adjacent images increases. In \cref{fig:effect_of_lcs}(c) and (d), we report transitions where the LCS values are consistently high across the morphing sequence. Human observers also perceive the transitions in (c) and (d) as smoother than those in (a) and (b), and our metric assigns higher scores to these transitions. These results show that the proposed LCS can evaluate perceptual smoothness in a way that is consistent with human judgment.

\begin{figure*}[!t]
    \centering
    \includegraphics[width=0.8\textwidth]{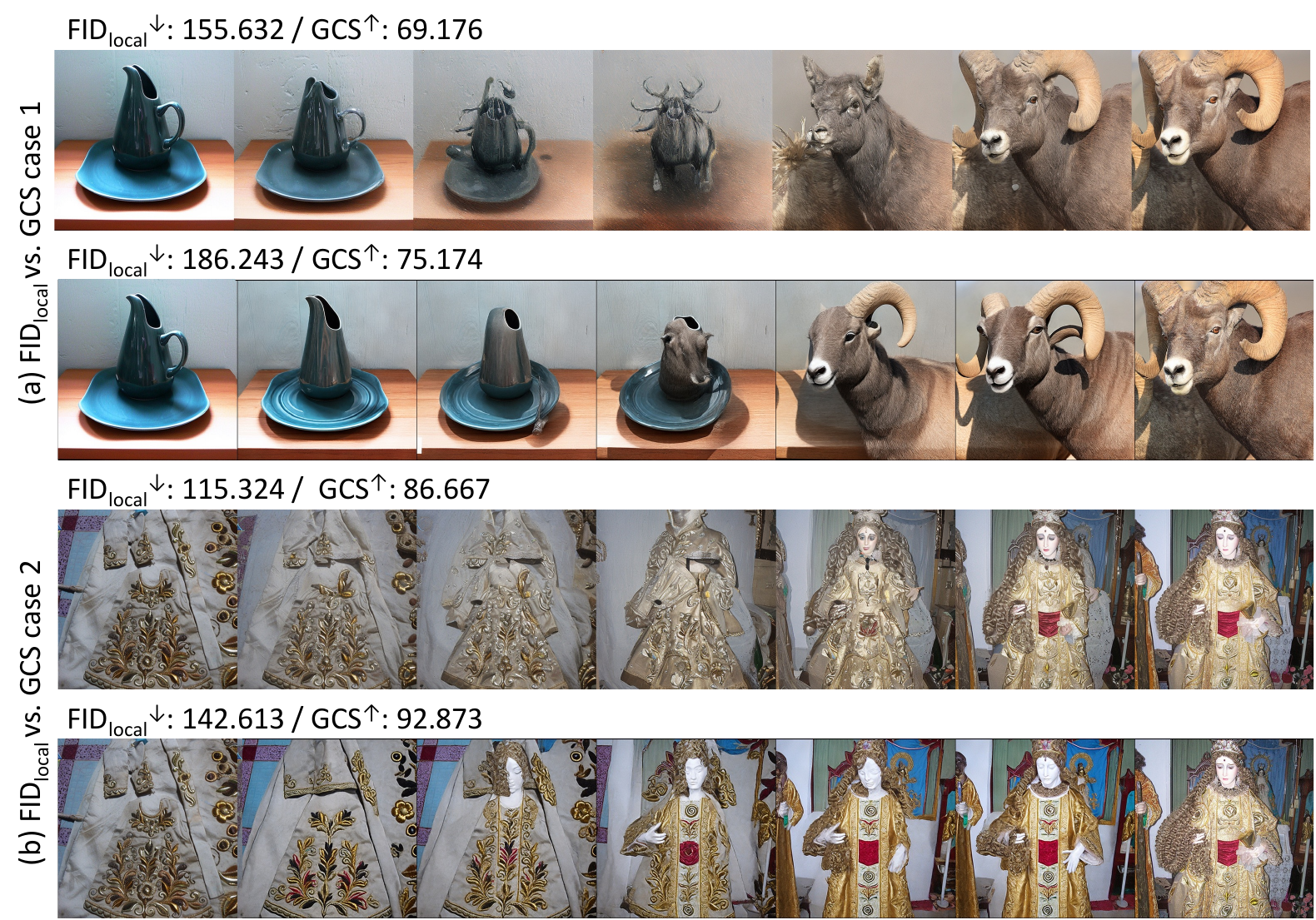}
    \caption{Qualitative comparisons between $\mathrm{FID}_{\text{local}}$ and GCS, which is a component of our proposed metric. Panels (a) and (b) present qualitative results for two different cases.}
    \label{fig:fid_vs_gcs}
\end{figure*}

\begin{figure*}[!b]
    \centering
    \includegraphics[width=0.8\textwidth]{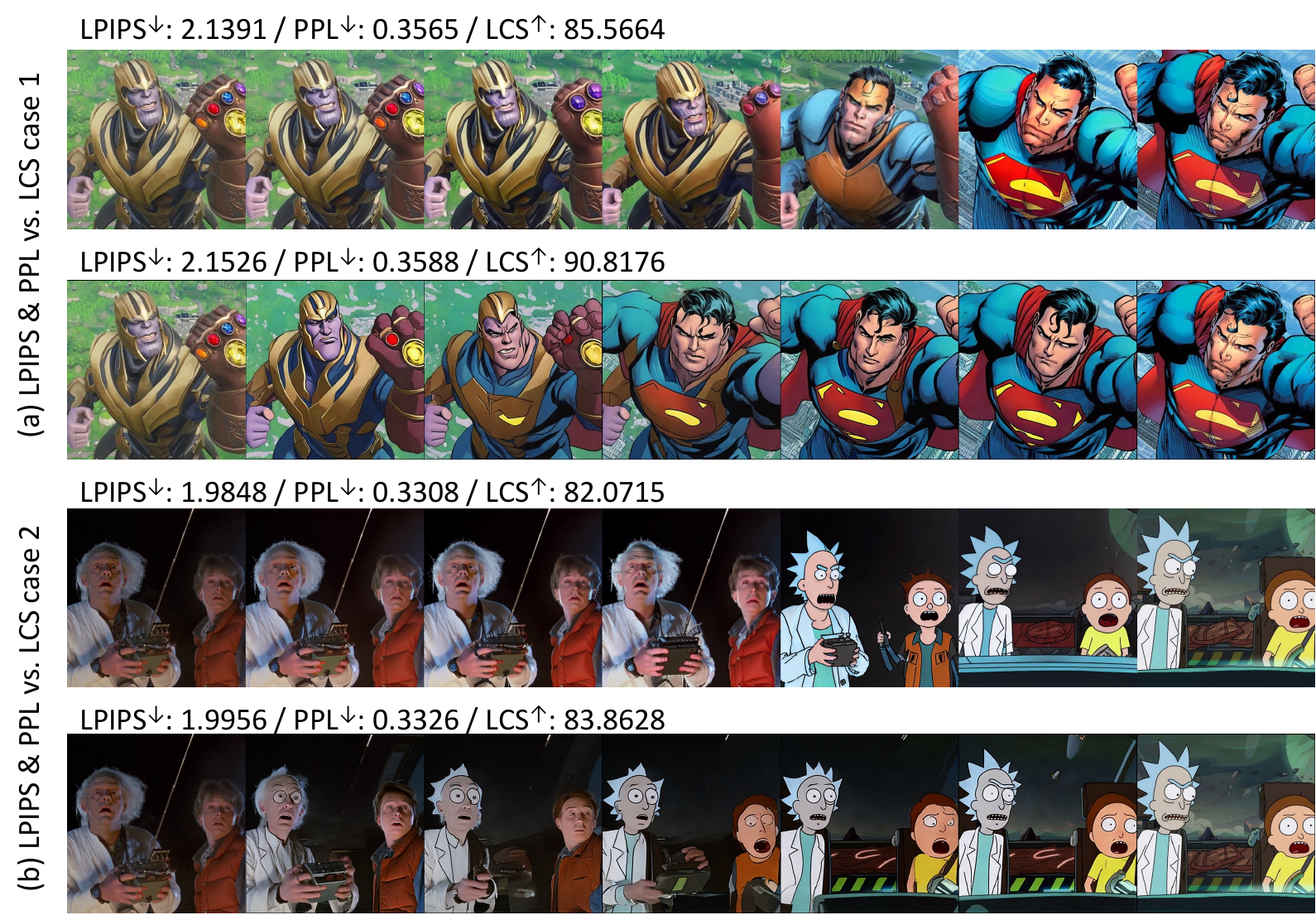}
    \caption{Qualitative comparisons between LPIPS, PPL, and LCS, which is a component of our proposed metric. Panels (a) and (b) present qualitative results for two different cases.}
    \label{fig:lpips_ppl_vs_lcs}
\end{figure*}

\subsection{Comparison between Traditional Metrics and GLCS}
\label{sup:trad_vs_glcs}

\cref{fig:fid_vs_gcs} provides a qualitative comparison between $\mathrm{FID}_{\text{local}}$ and GCS. As shown in \cref{fig:fid_vs_gcs}, the first rows of (a) and (b) achieve better $\mathrm{FID}_{\text{local}}$ scores than the second rows. However, visual inspection reveals that the third image in the first row of (a) does not properly include both domains of \(A\) and \(B\), and the fourth image even produces a result that is unrelated to image \(B\). Similarly, in the first row of (b), the third and fourth images contain almost no elements from image \(B\). These observations indicate that $\mathrm{FID}_{\text{local}}$ does not align well with human perception when evaluating domain consistency, since it only compares the overall distributions of \(A,B\) and the morphing images. 

In contrast, the proposed GCS evaluates whether each image properly reflects both domains of \(A\) and \(B\) according to the interpolation ratio. As a result, the second rows of (a) and (b), which better preserve domain consistency, are assigned higher quality scores than the first rows. This demonstrates that GCS provides a more human-aligned assessment of domain consistency.

\cref{fig:lpips_ppl_vs_lcs} presents a qualitative comparison between LPIPS, PPL, and LCS. As shown in \cref{fig:lpips_ppl_vs_lcs}, the first rows of (a) and (b) obtain higher LPIPS and PPL scores than the second rows. However, visual inspection shows that the second rows exhibit smoother transitions than the first rows. This indicates that LPIPS and PPL do not align well with human perception when evaluating smoothness, as they rely on VGG- and GAN-based networks. 

In contrast, the proposed LCS leverages DiffSim~\cite{song2025diffsim}, which measures diffusion-based similarity and benefits from diffusion priors to better match human perception. As a result, LCS assigns higher scores to the second rows in \cref{fig:lpips_ppl_vs_lcs}(a) and (b), which are perceived as smoother by human observers. These results demonstrate that the proposed LCS provides a perceptually aligned measure of transition smoothness.

\begin{table}[t]
  \centering
  \caption{GLCS with different similarity functions. Agreement denotes
  the ranking consistency between the GLCS-induced ranking and user study scores
  across the four morphing methods. All variants achieve 67--85\% agreement,
  showing that GLCS's reliability is not bottlenecked by a DiffSim-specific bias.}
  \label{tab:sim_ablation}
  \setlength{\tabcolsep}{12pt}
  \renewcommand{\arraystretch}{1.2}
  \begin{tabular}{l c}
    \toprule
    Similarity & User study agr. \\
    \midrule
    DreamSim~\cite{fu2023dreamsim}      & 67.66\% \\
    LipSim~\cite{ghazanfari2024lipsim}  & 70.00\% \\
    {\textbf{DiffSim (Ours)}}             & \textbf{80.00\%} \\
    DINO~\cite{oquab2023dinov2}         & 85.00\% \\
    \bottomrule
  \end{tabular}
  \vspace{-0.2cm}
\end{table}

\subsection{Analysis on the Similarity Function of GLCS.}
In this section, we provide an additional experiment on the similarity
function $s(\cdot,\cdot)$ used in GLCS. While all experiments in this paper use a
DiffSim-based~\cite{song2025diffsim} $s(\cdot,\cdot)$ (\cref{eq:sim_def}), we
measure its impact by replacing $s(\cdot,\cdot)$ with three alternative
similarity functions: DreamSim~\cite{fu2023dreamsim},
LipSim~\cite{ghazanfari2024lipsim}, and DINO~\cite{oquab2023dinov2}, to analyze
whether GLCS depends on a specific similarity function. As shown in
\cref{tab:sim_ablation}, we measure the ranking consistency between the
GLCS-induced ranking and the user study scores across the four methods
(IMPUS, DiffMorpher, FreeMorph, and CHIMERA); all variants achieve 67--85\%
agreement. This indicates that the reliability of GLCS stems from the GLCS
formulation itself rather than from any DiffSim-specific property.
Interestingly, we observe that using a DINO-based similarity yields the highest
agreement with the user study (85.00\%), suggesting that the choice of similarity
function in GLCS may offer additional room for improvement. Since the primary
focus of this work is the GLCS formulation itself for morphing evaluation, we
leave a more systematic analysis of the similarity function and the corresponding
additional experiments as future work.
The full algorithm for GLCS is provided in \cref{alg:glcs}.

\begin{figure}[h]
  \centering
    \includegraphics[width=0.9\linewidth]{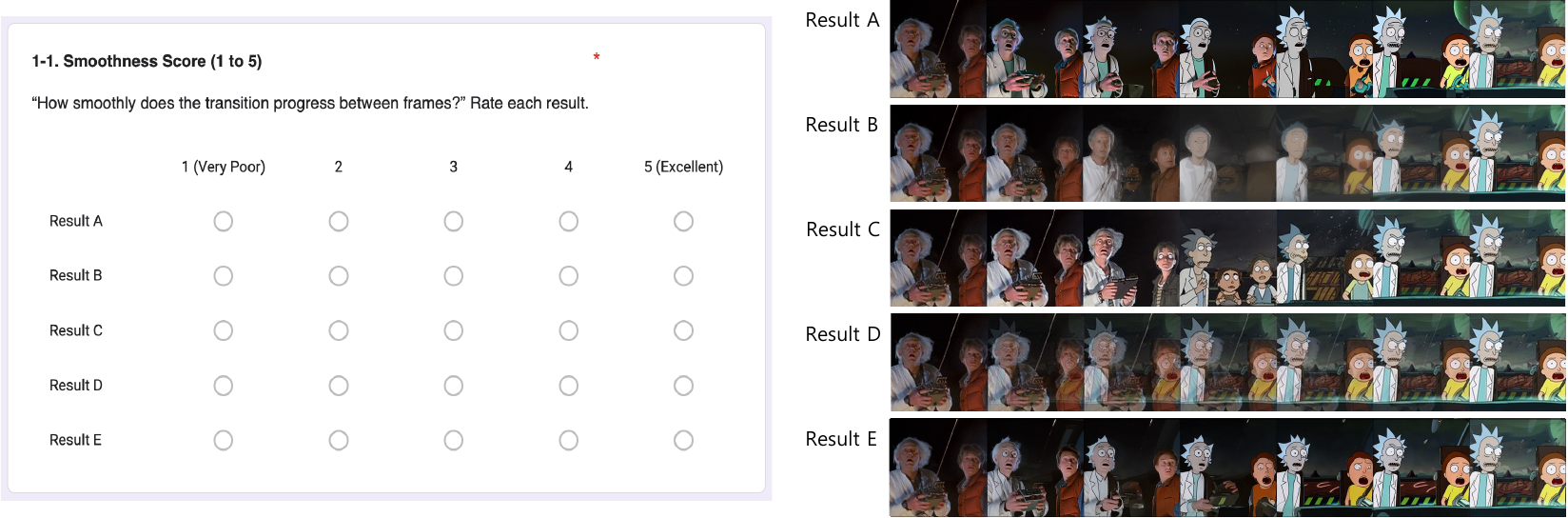} 
    \caption{\textbf{User study interface and questionnaire form.}}
    \label{fig:user_study_interface}
    \vspace{-0.2cm}
\end{figure}

\section{User Study: Subjective Preference Analysis}
\label{sec:userstudy_supp}

\subsection{Protocol}
We conduct a user study on 15 morphing sequences to assess how well each method aligns with human perception.
For each sequence, 36 participants are shown five anonymized results
(\(A\)--\(E\)) generated by CHIMERA, FreeMorph~\cite{cao2025freemorph},
DiffMorpher~\cite{zhang2024diffmorpher}, IMPUS~\cite{yang2024impus}, and latent \texttt{slerp} (see~\cref{fig:user_study_interface}).
The mapping between \(\{A,\dots,E\}\) and the underlying methods is randomized per sequence and participant.
Participants rate each result on a 5-point Likert scale (1–5) for four criteria:
Smoothness, Domain Consistency, Perceptual Quality, and Overall Quality.

\begin{table}[t!]
\centering
\caption{Mean opinion scores (MOS), mean rank, and Borda score of each method in the user study.
CHIMERA consistently achieves the highest MOS, the best (lowest) mean rank, and the highest Borda score across all four criteria, indicating a strong overall user preference over existing morphing methods.}
\resizebox{\columnwidth}{!}{%
\begin{tabular}{@{}>{\raggedright\arraybackslash}p{2.0cm} l c c c@{}}
\toprule
Criteria & Method & MOS $\uparrow$ & Mean rank $\downarrow$ & Borda score $\uparrow$ \\
\midrule
\multirow{5}{*}{Smoothness}
  & \textbf{CHIMERA (Ours)} 
    & \best{3.828} $\pm$ 0.501 
    & \best{1.486} 
    & \best{4.514} \\
  & FreeMorph [ICCV'25]        
    & 2.998 $\pm$ 0.472 
    & 3.222 
    & 2.778 \\
  & DiffMorpher [CVPR'24]      
    & \second{3.574} $\pm$ 0.481 
    & \second{2.014} 
    & \second{3.986} \\
  & IMPUS [ICLR'24]            
    & 2.815 $\pm$ 0.514 
    & 3.819 
    & 2.181 \\
  & \texttt{slerp}   
    & 2.243 $\pm$ 0.936 
    & 4.458 
    & 1.542 \\
\midrule
\multirow{5}{*}{\makecell[l]{Domain\\Consistency}}
  & \textbf{CHIMERA (Ours)} 
    & \best{3.661} $\pm$ 0.537 
    & \best{1.847} 
    & \best{4.153} \\
  & FreeMorph [ICCV'25]        
    & 2.728 $\pm$ 0.612 
    & 3.736 
    & 2.264 \\
  & DiffMorpher [CVPR'24]      
    & \second{3.489} $\pm$ 0.511 
    & \second{2.208} 
    & \second{3.792} \\
  & IMPUS [ICLR'24]            
    & 3.172 $\pm$ 0.547 
    & 2.792 
    & 3.208 \\
  & \texttt{slerp}   
    & 2.326 $\pm$ 0.880 
    & 4.417 
    & 1.583 \\
\midrule
\multirow{5}{*}{\makecell[l]{Perceptual\\Quality}}
  & \textbf{CHIMERA (Ours)} 
    & \best{3.635} $\pm$ 0.594 
    & \best{1.819} 
    & \best{4.181} \\
  & FreeMorph [ICCV'25]        
    & 2.957 $\pm$ 0.597 
    & 3.153 
    & 2.847 \\
  & DiffMorpher [CVPR'24]      
    & \second{3.383} $\pm$ 0.459 
    & \second{2.472} 
    & \second{3.528} \\
  & IMPUS [ICLR'24]            
    & 3.270 $\pm$ 0.472 
    & 2.694 
    & 3.306 \\
  & \texttt{slerp}   
    & 1.826 $\pm$ 0.824 
    & 4.861 
    & 1.139 \\
\midrule
\multirow{5}{*}{\makecell[l]{Overall\\Quality}}
  & \textbf{CHIMERA (Ours)} 
    & \best{3.639} $\pm$ 0.607 
    & \best{1.625} 
    & \best{4.375} \\
  & FreeMorph [ICCV'25]        
    & 2.906 $\pm$ 0.519 
    & 3.431 
    & 2.569 \\
  & DiffMorpher [CVPR'24]      
    & \second{3.404} $\pm$ 0.448 
    & \second{2.194} 
    & \second{3.806} \\
  & IMPUS [ICLR'24]            
    & 3.067 $\pm$ 0.495 
    & 2.972 
    & 3.028 \\
  & \texttt{slerp}   
    & 1.913 $\pm$ 0.822 
    & 4.778 
    & 1.222 \\
\bottomrule
\end{tabular}
}
\label{tab:userstudy_mos}
\end{table}

\subsection{Statistical Analyses}
\subsubsection{Mean Opinion Scores}
From the resulting user--sequence--method score matrix, we first aggregate the scores over the 15 sequences for each participant and method, and then compute the mean opinion score (MOS), standard deviation, and average rank (lower is better) for each method and criterion.
These statistics are summarized in \cref{tab:userstudy_mos}.
CHIMERA achieves the highest MOS, the lowest mean rank, and the highest Borda score across all four criteria.
While DiffMorpher shows competitive performance, particularly in Domain Consistency where the MOS gap is the narrowest, CHIMERA still maintains a clear and consistent superiority across all metrics including Smoothness, Perceptual Quality, and Overall Quality.
Meanwhile, IMPUS and FreeMorph generally receive lower MOS and worse mean ranks, with \texttt{slerp} consistently performing the worst.

\begin{table}[!t]
\centering
\caption{
Friedman test over the five methods for each subjective criterion.
In all cases, the null hypothesis that all methods are equivalent is rejected
($p \ll 0.05$), confirming statistically significant differences in user ratings.}
\footnotesize
\begin{tabular}{lcc}
\toprule
Criteria & Friedman $\chi^2$ & $p$-value \\
\midrule
Smoothness          & 89.251 & $1.900\times 10^{-18}$ \\
Domain Consistency  & 66.225 & $1.420\times 10^{-13}$ \\
Perceptual Quality  & 76.168 & $1.128\times 10^{-15}$ \\
Overall Quality     & 85.354 & $1.276\times 10^{-17}$ \\
\bottomrule
\end{tabular}
\label{tab:userstudy_friedman}
\end{table}
\subsubsection{Significance Test}
To test whether the observed differences are statistically meaningful, we apply
a Friedman test over the five methods for each criterion, treating each participant
as a block. \cref{tab:userstudy_friedman} reports the resulting test statistics and $p$–values.
For all criteria, the null hypothesis that all methods are equivalent is rejected
with $p \ll 0.05$, indicating that the differences in participant-level preferences across methods are statistically significant.

\subsubsection{Pairwise Preferences}
We further analyze pairwise preferences between our CHIMERA and each baseline.
For each participant, the scores of CHIMERA and a baseline are first aggregated over the 15 sequences for a given criterion and then compared to count wins (CHIMERA $>$ baseline), ties, and losses.
The win--tie--loss statistics in \cref{tab:userstudy_wtl}, also visualized in ~\cref{fig:teaser1}~(b), show that CHIMERA wins the majority of comparisons across all four criteria.
Among the baselines, DiffMorpher is the strongest competitor, especially in domain consistency, but CHIMERA still maintains a consistent overall advantage across all compared methods.

\begin{table}[t]
\centering
\caption{
Win--tie--loss statistics of CHIMERA against each baseline.
For each participant, we aggregate the scores over the 15 sequences for a given criterion and compare the resulting score of CHIMERA with that of each baseline to count wins (CHIMERA $>$ baseline), ties, and losses.
CHIMERA wins in the majority of cases, showing consistent subjective superiority. This is also visualized in \cref{fig:teaser1}~(b)
}
\resizebox{\columnwidth}{!}{%
\begin{tabular}{l l c}
\toprule
Criteria & Baseline & W / T / L vs.\ \textbf{CHIMERA (Ours)} \\
\midrule
\multirow{4}{*}{\shortstack[l]{Smoothness}}
  & FreeMorph [ICCV'25]   & \textcolor[HTML]{039C6E}{\textbf{36}} / $0$ / $0$ \\
  & DiffMorpher [CVPR'24] & \textcolor[HTML]{039C6E}{\textbf{21}} / $2$ / $13$ \\
  & IMPUS [ICLR'24]       & \textcolor[HTML]{039C6E}{\textbf{34}} / $1$ / $1$ \\
  & \texttt{slerp}        & \textcolor[HTML]{039C6E}{\textbf{34}} / $0$ / $2$ \\
\midrule
\multirow{4}{*}{\shortstack[l]{Domain\\Consistency}}
  & FreeMorph [ICCV'25]   & \textcolor[HTML]{039C6E}{\textbf{36}} / $0$ / $0$ \\
  & DiffMorpher [CVPR'24] & \textcolor[HTML]{039C6E}{\textbf{18}} / $2$ / $16$ \\
  & IMPUS [ICLR'24]       & \textcolor[HTML]{039C6E}{\textbf{26}} / $1$ / $9$ \\
  & \texttt{slerp}        & \textcolor[HTML]{039C6E}{\textbf{32}} / $0$ / $4$ \\
\midrule
\multirow{4}{*}{\shortstack[l]{Perceptual\\Quality}}
  & FreeMorph [ICCV'25]   & \textcolor[HTML]{039C6E}{\textbf{31}} / $0$ / $5$ \\
  & DiffMorpher [CVPR'24] & \textcolor[HTML]{039C6E}{\textbf{23}} / $0$ / $13$ \\
  & IMPUS [ICLR'24]       & \textcolor[HTML]{039C6E}{\textbf{25}} / $1$ / $10$ \\
  & \texttt{slerp}        & \textcolor[HTML]{039C6E}{\textbf{35}} / $0$ / $1$ \\
\midrule
\multirow{4}{*}{\shortstack[l]{Overall\\Quality}}
  & FreeMorph [ICCV'25]   & \textcolor[HTML]{039C6E}{\textbf{35}} / $0$ / $1$ \\
  & DiffMorpher [CVPR'24] & \textcolor[HTML]{039C6E}{\textbf{21}} / $2$ / $13$ \\
  & IMPUS [ICLR'24]       & \textcolor[HTML]{039C6E}{\textbf{29}} / $1$ / $6$ \\
  & \texttt{slerp}        & \textcolor[HTML]{039C6E}{\textbf{35}} / $0$ / $1$ \\
\bottomrule
\end{tabular}
}
\label{tab:userstudy_wtl}
\end{table}

\subsubsection{Relation to GLCS}
We further compare the user study outcomes with our GLCS-based quantitative evaluation.
Among the four methods for which GLCS is defined (CHIMERA, FreeMorph, DiffMorpher, and IMPUS),
CHIMERA attains the highest GLCS on both MorphBench and Morph4Data and, at the same time, achieves the highest Overall Quality MOS, the best mean rank, and the highest Borda score in \cref{tab:userstudy_mos}.
Methods with lower GLCS values also tend to receive lower MOS and worse ranks in the user study,
indicating that GLCS is aligned with human preference at the method level.
Given this agreement between human judgments and dataset-level scores, we regard GLCS as a
promising reference metric for future image morphing research, providing a principled quantitative
measure that jointly reflects temporal smoothness and semantic consistency.
\section{Application}
\label{sp_sec:application}

\begin{figure}
    \centering
    \includegraphics[width=0.9\linewidth]{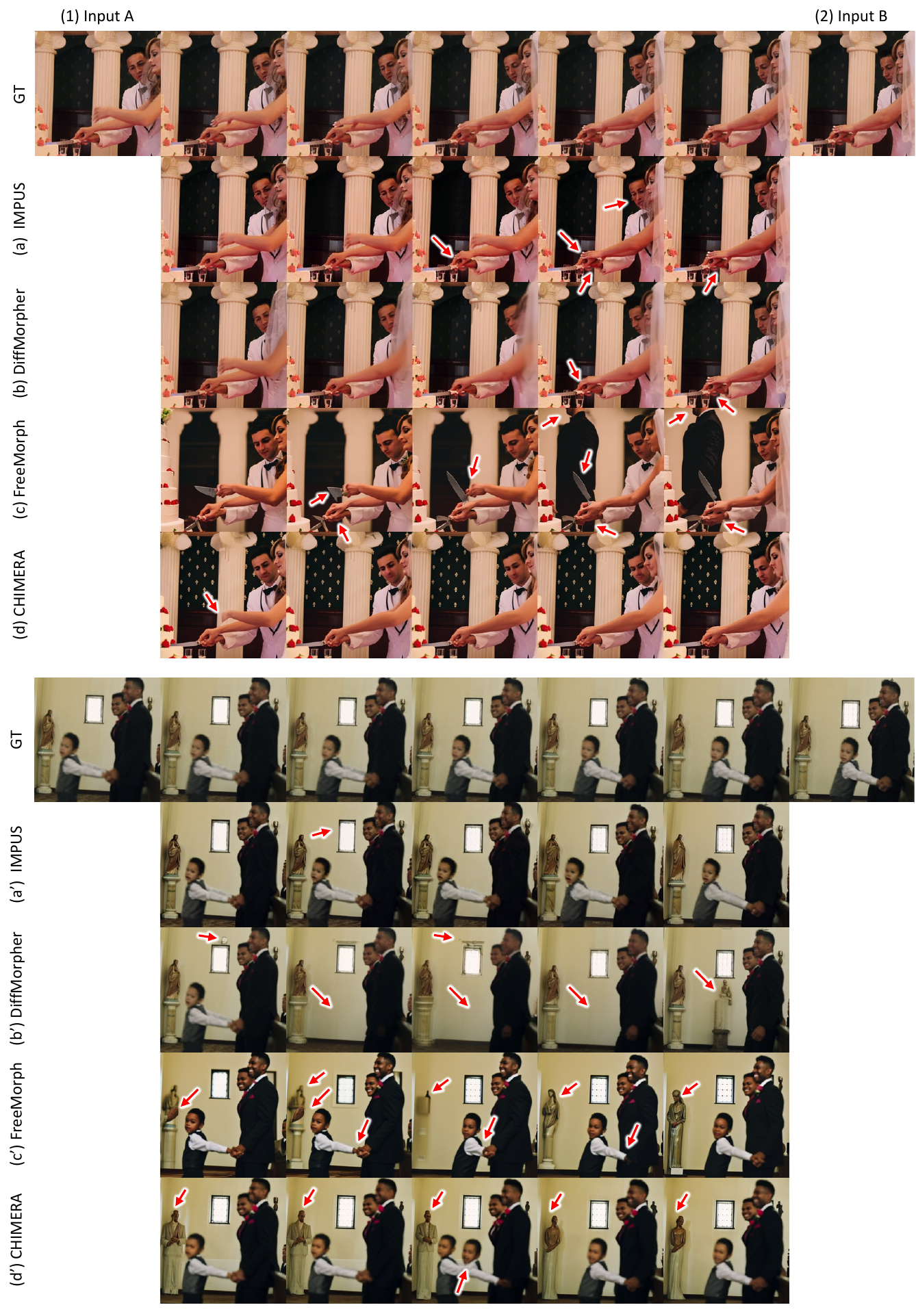}
    \caption{Qualitative VFI results on Vimeo90K-septuplet.
    Panels (a)–(d) and (a')-(d') correspond to IMPUS, DiffMorpher, FreeMorph, and CHIMERA (Ours), respectively. 
    For each sequence, red arrows mark representative artifacts such as unrealistic limb configurations or duplicated local structures in the interpolated frames.}
    \label{fig:vimeo_qual}
    \vspace{-0.3cm}
\end{figure}

\begin{figure}
    \centering
    \includegraphics[width=0.9\linewidth]{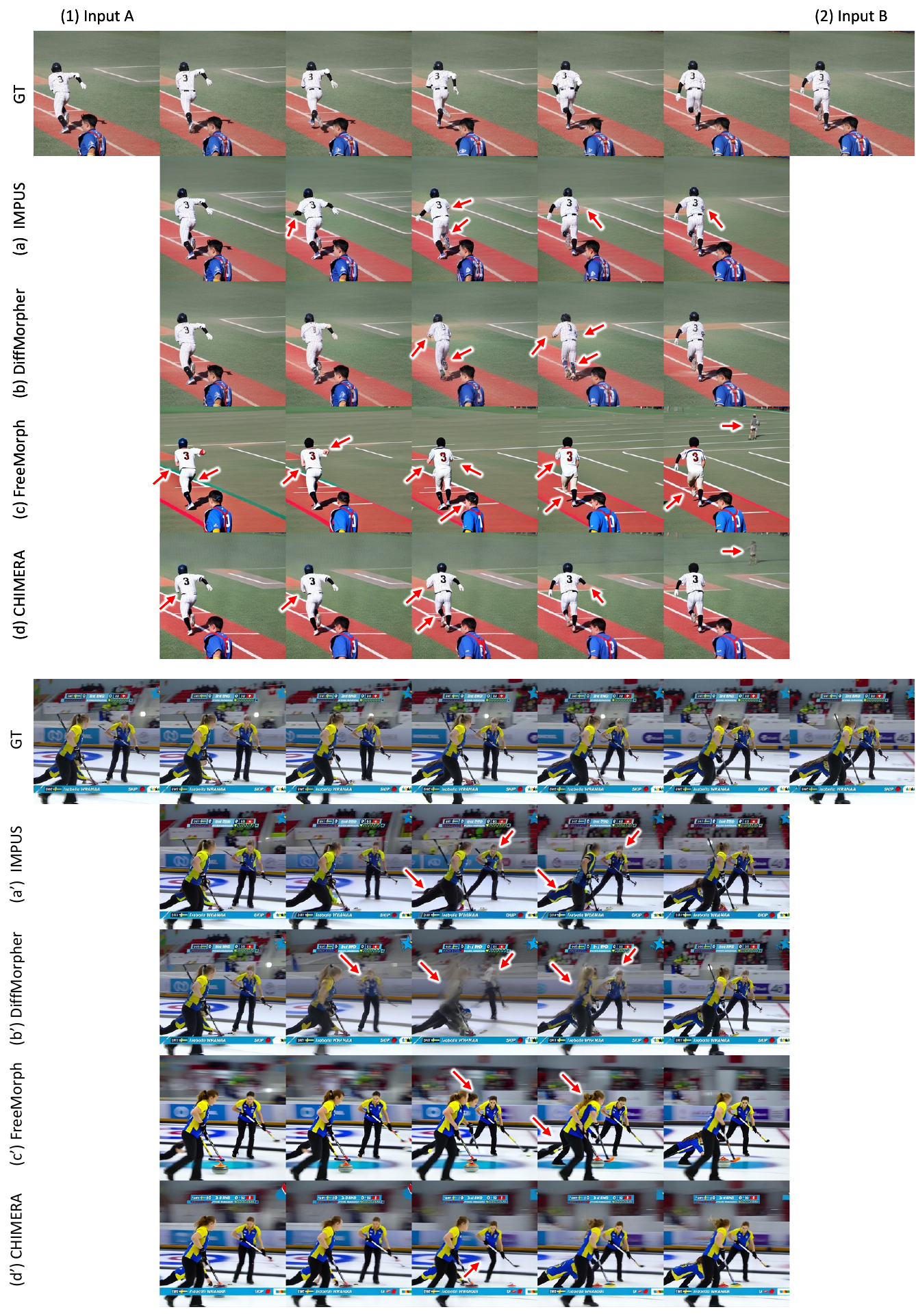}
    \caption{Qualitative VFI results on DAVIS.
    Panels (a)–(d) and (a')-(d') correspond to IMPUS, DiffMorpher, FreeMorph, and CHIMERA (Ours), respectively. 
    The red arrows highlight severe failure cases where the interpolated results exhibit non-physical human bodies, including truncated or distorted arms and legs.}
    \label{fig:davis_qual}
    \vspace{-0.3cm}
\end{figure}

\subsection{Video Frame Interpolation}
Although CHIMERA is designed for still-image morphing, its capability to generate temporally dense sequences naturally suggests an application to video frame interpolation (VFI). To probe this connection, frames from VFI benchmark datasets~\cite{xue2019video, pont20172017} are used as input, where two frames separated by a fixed temporal offset are treated as endpoints and the intermediate outputs of CHIMERA are interpreted as interpolated results. 
As shown in~\cref{fig:vimeo_qual} on Vimeo90K-septuplet~\cite{xue2019video} some frames visually resemble reasonable interpolation, but noticeable artifacts remain. In the CHIMERA rows ((d), (d')), the red arrows highlight typical failure modes such as truncated or duplicated limbs and locally distorted arm configurations. Similar issues are also observed in the other morphing baselines. IMPUS~\cite{yang2024impus} (rows (a), (a')) often produces implausible hand shapes or causes arms to partially disappear mid sequence. DiffMorpher~\cite{zhang2024diffmorpher} (rows (b), (b') yields over-smoothed and blurry frames consistent with its morphing behavior, and FreeMorph~\cite{cao2025freemorph} (rows (c), (c')) hallucinates content absent from both inputs (e.g., transforming a statue into a realistic human). 

On the DAVIS dataset~\cite{pont20172017} in~\cref{fig:davis_qual}, where human motion and occlusions are more complex, all morphing methods exhibit pronounced non-physical deformations. CHIMERA (rows (d), (d')) generates unrealistic human bodies with truncated or severely warped arms and legs, and sometimes stretches limbs into unnatural shapes across frames. IMPUS (rows (a), (a')) produces broken silhouettes with missing or dislocated arms, while DiffMorpher (rows (b), (b')) shows similar limb truncation together with strong motion blur that obscures fine details. FreeMorph (rows (c), (c')) suffers from distorted body shapes and over-saturated colors and, like CHIMERA, sometimes hallucinates entirely new objects in the background. Overall, these observations indicate that such failures are not specific to our method but are inherent to morphing methods when applied to VFI data.

We conjecture that this stems from a fundamental mismatch between the objectives of morphing and VFI. Unlike VFI methods that establish explicit correspondences between input frames and reconstruct the motion trajectory connecting them through optical flow~\cite{niklaus2020softmax}, deformable kernels~\cite{cheng2021multiple}, or learned spatiotemporal representations~\cite{zhang2025eden, kye2025acevfi}, morphing models operate as generative processes that synthesize plausible in between states without being constrained to follow the true motion path. CHIMERA has no motion specific modules and receives no supervision from real videos; it is optimized for smooth transitions between two inputs rather than faithful reconstruction of motion trajectories. Moreover, CHIMERA is applied to VFI datasets in a purely zero-shot setting without domain specific fine-tuning, further widening the gap relative to VFI models. As a result, intermediate frames can traverse “imagined” states in latent space that do not correspond to physically realizable frames, which is acceptable or even desirable in morphing contexts but manifests as artifacts in VFI benchmarks.

Overall, these observations indicate that CHIMERA is distinct from reconstruction-driven VFI methods. They also suggest a natural extension: augmenting the cache and prompt-based design with explicit motion priors~\cite{wang2024generative, liu2024sparse, seo2025bim} and video-driven objectives~\cite{wu2024perception, chen2025repurposing} could evolve the framework toward a VFI model that better satisfies the physical and temporal requirements of standard benchmarks.

\subsection{Creative Content Creation and Animation}
CHIMERA directly supports applications in film, game, and animation production, where artists often require smooth transitions between disparate visual concepts~\cite{sun2026morphany3d}. Given two images that serve as keyframes, the framework generates a temporally dense sequence of structurally consistent and semantically coherent intermediate frames without manual correspondence annotation or model fine-tuning. This capability aligns with the growing demand for engaging transitions in short-form video platforms (e.g., TikTok, Kuaishou), where visually distinctive morphing effects contribute to viewer engagement and content memorability. By providing zero-shot generation of high-quality metamorphic transitions, CHIMERA lowers the barrier for both professional creators and non-experts to prototype and deploy production-ready visual effects, ranging from character evolution and object transformations to stylized scene changes tailored for short-form content.

\begin{table}[t!]
  \centering
  \caption{Comparison with video-prior methods on Morph4Data.
  Each method is applied to morphing inputs under its native task. CHIMERA
  achieves the highest GLCS, while video-prior methods degrade due to endpoint
  loss or structural/ghosting artifacts. Best in \textcolor{red}{red},
  second best in \textcolor{blue}{blue}.}
  \label{tab:i2v_qual}

  \vspace{2pt}
  \small
  \setlength{\tabcolsep}{2pt}
  \renewcommand{\arraystretch}{1.05}

  \begin{tabular}{l c c}
    \toprule
    Method & Task & GLCS$\uparrow$ \\
    \midrule
    LTX~\cite{hacohen2024ltxvideo}
      & Image to Video
      & 76.428 \\
    TLB-VFI~\cite{lyu2025tlbvfi}
      & Frame Interpolation
      & 79.425 \\
    FCVG~\cite{zhu2025fcvg}
      & Inbetweening
      & \textcolor{blue}{\underline{83.001}} \\
    \textbf{CHIMERA (Ours)}
      & Morphing
      & \textbf{\textcolor{red}{88.079}} \\
    \bottomrule
  \end{tabular}
\end{table}

\begin{figure}[t]
  \centering
  \includegraphics[width=0.7\linewidth]{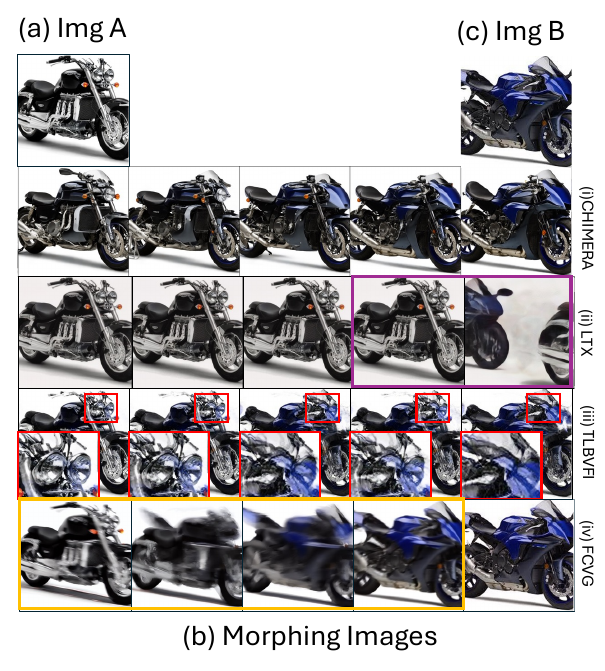}
  \caption{Qualitative comparison with video-prior methods on a
  heterogeneous morphing pair. (a) Input A, (c) Input B, and (b) the morphing
  sequences produced by (i) CHIMERA, (ii) LTX, (iii) TLB-VFI, and (iv) FCVG.
  \textcolor{violet}{Purple}: loss of endpoint information;
  \textcolor{red}{red}: structural inconsistency;
  \textcolor{orange}{yellow}: ghosting/blurring artifacts.}
  \label{fig:i2v_qual}
\end{figure}



\subsection{Comparison with Image to Video / Video Frame Interpolation / Inbetweening Methods}
\label{sec:supp_i2v}
Morphing is fundamentally different from video-based tasks in that it requires preserving heterogeneous endpoints rather than temporal/motion continuity. To verify this, we apply three recent video-prior methods, LTX~\cite{hacohen2024ltxvideo} (Image to Video), TLB-VFI~\cite{lyu2025tlbvfi} (Frame Interpolation), and FCVG~\cite{zhu2025fcvg} (Inbetweening), to Morph4Data and compare them with CHIMERA. These methods are designed to prioritize smooth motion transitions between adjacent frames, making them ill-suited for morphing inputs whose endpoints differ substantially in semantics and structure. As shown in \cref{tab:i2v_qual}, CHIMERA attains the highest GLCS ($88.079$), whereas all three video-prior methods yield substantially lower GLCS. This gap is
qualitatively evident in \cref{fig:i2v_qual}: LTX loses endpoint information in the intermediate frames (purple box), TLB-VFI exhibits structural inconsistency (red box), and FCVG suffers from ghosting or blurring artifacts (yellow box). In contrast, CHIMERA preserves the identity of both endpoints while producing structurally consistent and smooth transitions. This demonstrates that methods designed for video tasks do not directly transfer to heterogeneous morphing, and that morphing requires a dedicated formulation.
\section{Extended Experiment Results}
\label{sp_sec:extend_exp}

\begin{table*}[t!]
\centering
\begingroup
\small
\setlength{\tabcolsep}{3pt}
\renewcommand{\arraystretch}{1.05}
\caption{Quantitative results of 14-image morphing on Morph4Data and MorphBench datasets. The best scores are marked in bold, while the second best are underlined.}
\resizebox{\linewidth}{!}{%
\begin{tabular}{l c c c c c @{\qquad} l c c c c c}

\toprule
\multicolumn{6}{c}{\textbf{Morph4Data}} 
&
\multicolumn{6}{c}{\textbf{MorphBench}} \\
\cmidrule(r){1-6} \cmidrule(l){7-12}

Model name 
& $\mathrm{FID}_{\text{local}}\downarrow$ 
& $\mathrm{FID}_{\text{global}}\downarrow$ 
& LPIPS$\downarrow$ 
& PPL$\downarrow$ 
& GLCS$\uparrow$
& Model name 
& $\mathrm{FID}_{\text{local}}\downarrow$ 
& $\mathrm{FID}_{\text{global}}\downarrow$ 
& LPIPS$\downarrow$ 
& PPL$\downarrow$ 
& GLCS$\uparrow$ \\
\midrule

IMPUS
& \textbf{\textcolor{red}{120.815}}
& \textbf{\textcolor{red}{60.046}}
& 2.737
& 0.183
& 88.944
& IMPUS
& \textbf{\textcolor{red}{78.944}} 
& \textbf{\textcolor{red}{40.892}}
& 1.587
& 0.106
& 93.679 \\

DiffMorpher
& 175.409
& 89.870 
& \underline{\textcolor{blue}{1.875}}
& \underline{\textcolor{blue}{0.125}}
& 89.212
& DiffMorpher
& \textcolor{blue}{\underline{90.739}}
& \textcolor{blue}{\underline{46.176}}
& \textbf{\textcolor{red}{1.051}}
& \textbf{\textcolor{red}{0.070}}
& \textcolor{blue}{\underline{94.814}} \\

FreeMorph
& 178.792
& 94.062
& 2.538
& 0.169
& \textcolor{blue}{\underline{90.151}}
& FreeMorph
& 141.727
& 79.178
& 1.776
& 0.118
& 92.412 \\

\textbf{CHIMERA (Ours)}
& \textcolor{blue}{\underline{151.359}}
& \textcolor{blue}{\underline{78.569}}
& \textcolor{red}{\textbf{1.846}}
& \textcolor{red}{\textbf{0.123}}
& \textbf{\textcolor{red}{91.888}}
& \textbf{CHIMERA (Ours)}
& 111.496
& 61.401
& \textcolor{blue}{\underline{1.141}}
& \textcolor{blue}{\underline{0.076}}
& \textbf{\textcolor{red}{95.593}} \\
\bottomrule
\end{tabular}
} 

\label{tab:main_quan_16f}
\endgroup
\end{table*} 

This section provides additional quantitative and qualitative evaluations that supplement \cref{exp:eval}.  
In \cref{sp_sec:extend_quan}, we present qualitative and quantitative results for the setting where 14 morphing images are generated between input images \(A\) and \(B\). Unlike \cref{exp:eval}, which reports quantitative results for the 5-image morphing setting, this section evaluates CHIMERA under a longer morphing transition to assess the general applicability of the proposed method.  
In addition, \cref{sp_sec:extend_qual} provides further qualitative results for the 5-image morphing scenario discussed in~\cref{exp:qual}.

\subsection{Extended Evaluation on Challenging 14-Image Morphing}
\label{sp_sec:extend_quan}

\begin{figure*}[!t]
    \centering
    \includegraphics[width=\textwidth]{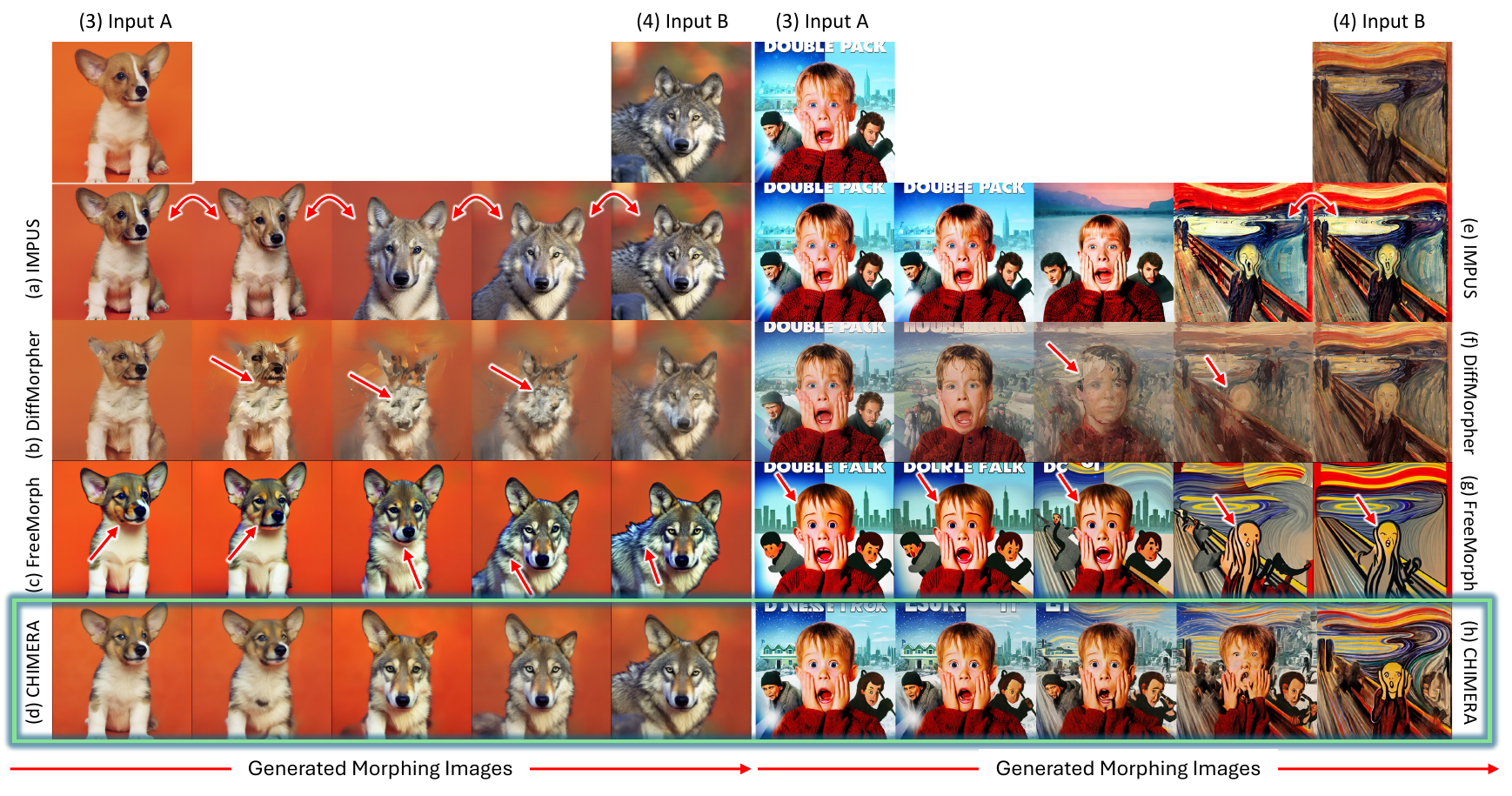}
    \caption{First qualitative comparison showing the results of generating five morphing images between input images \(A\) and \(B\). Panels (1)–(4) denote the input images, and panels (a)–(d) correspond to IMPUS, DiffMorpher, FreeMorph, and CHIMERA (Ours), respectively. The same convention applies to panels (e)–(h).}
    \label{fig:7f_qual3}
\end{figure*}

\begin{figure*}[!b]
    \centering
    \includegraphics[width=\textwidth]{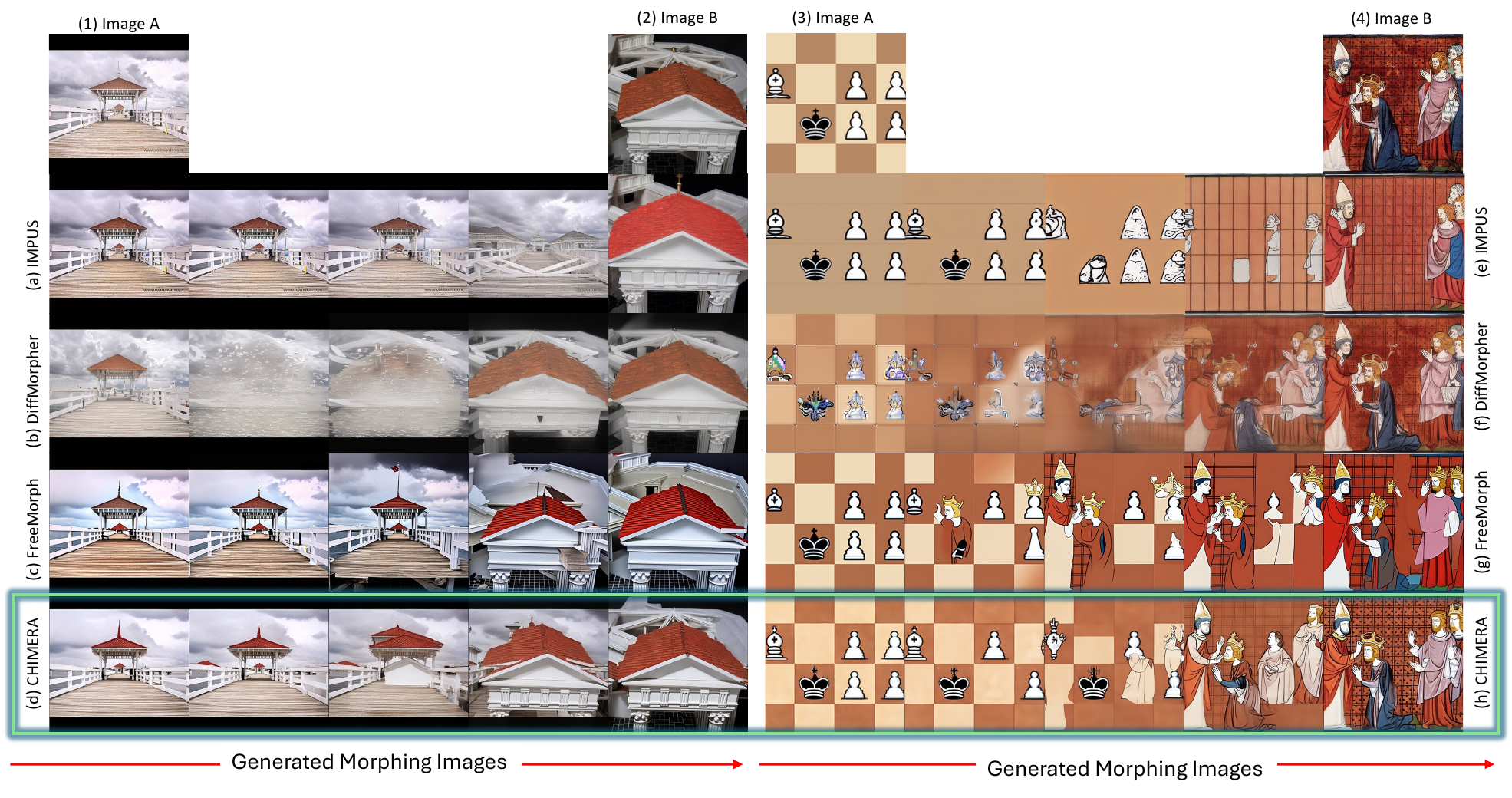}
    \caption{Second qualitative comparison showing the results of generating five morphing images between input images \(A\) and \(B\). Panels (1)–(4) denote the input images, and panels (a)–(d) correspond to IMPUS, DiffMorpher, FreeMorph, and CHIMERA (Ours), respectively. The same convention applies to panels (e)–(h).}
    \label{fig:7f_qual4}
\end{figure*}

\cref{tab:main_quan_16f} reports the quantitative results for the setting where 14 morphing images are generated between each input image pair.
Similar to the observations in~\cref{exp:quant}, IMPUS achieves the best scores in $\mathrm{FID}_{\text{local}}$ and $\mathrm{FID}_{\text{global}}$ on both datasets, but shows weaker performance in LPIPS, PPL, and GLCS. DiffMorpher obtains the best LPIPS and PPL scores, yet its performance in $\mathrm{FID}_{\text{local}}$, $\mathrm{FID}{\text{global}}$, and GLCS is relatively lower. FreeMorph shows degraded performance in all metrics except GLCS.

In contrast, the proposed CHIMERA demonstrates performance comparable to the fine-tuning-based models IMPUS and DiffMorpher across $\mathrm{FID}_{\text{local}}$, $\mathrm{FID}_{\text{global}}$, LPIPS, and PPL, while achieving a significantly higher GLCS. Furthermore, compared to FreeMorph, which is also a zero-shot model, CHIMERA outperforms it by a large margin across all metrics.

Qualitatively, IMPUS maintains strong domain consistency in each generated image but lacks smooth transitions between frames. DiffMorpher produces smooth transitions but often introduces severe artifacts, leading to poor domain consistency. FreeMorph provides visually smooth transitions but suffers from overly saturated colors, which also reduces domain consistency. In contrast, CHIMERA achieves both smooth frame-to-frame transitions and strong domain consistency, making it superior across both qualitative and quantitative evaluations.

\begin{figure*}[t]
    \centering
    \includegraphics[width=0.85\textwidth]{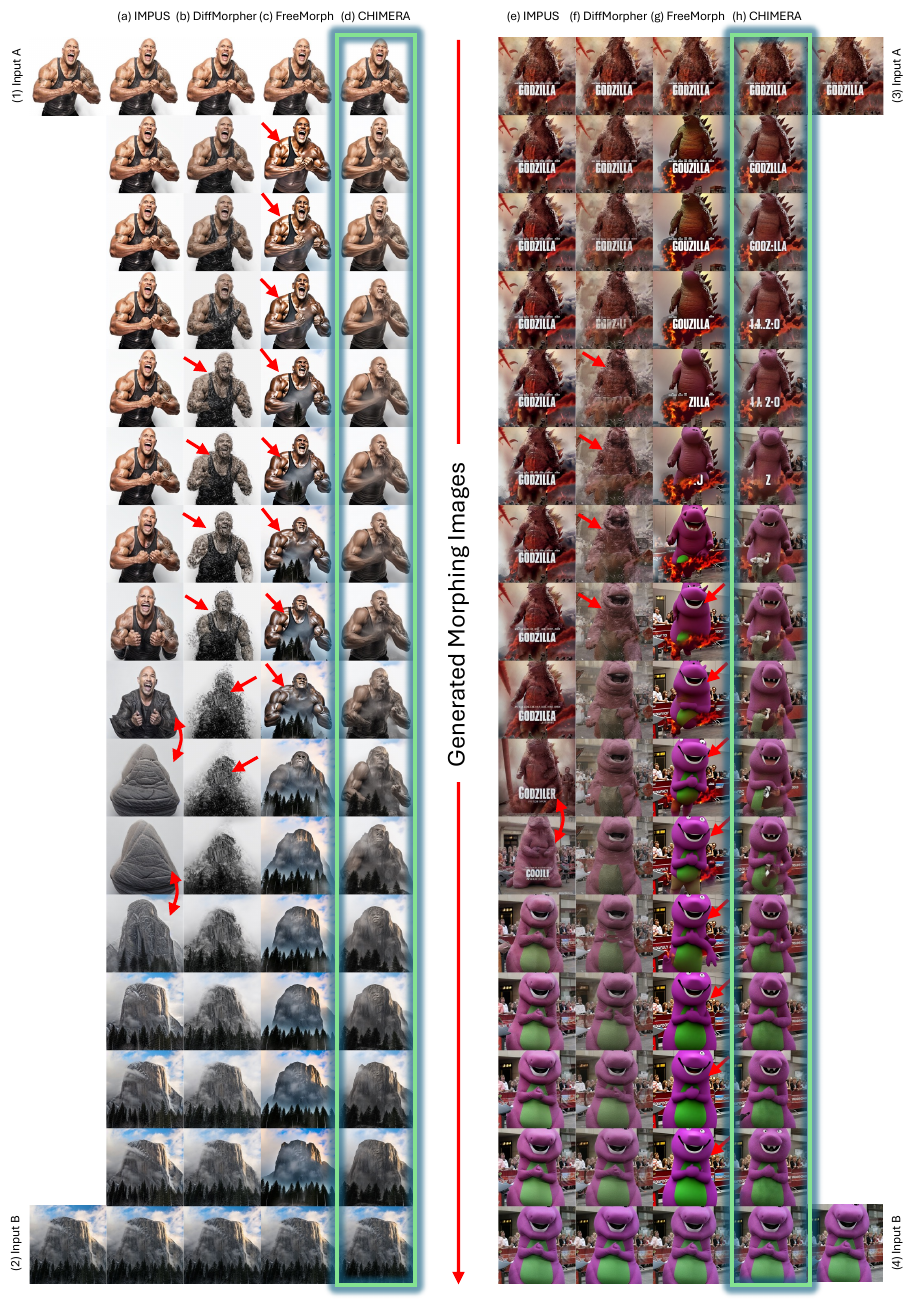}
    \caption{First qualitative comparison showing the results of challenging 14-image morphing (compared to 5-image morphing) between input images \(A\) and \(B\). Panels (1)–(4) denote the input images, and panels (a)–(d) correspond to IMPUS, DiffMorpher, FreeMorph, and CHIMERA (Ours), respectively. The same convention applies to panels (e)–(h). Please zoom in for better visualization.}
    \label{fig:16f_qual1}
\end{figure*}

\begin{figure*}[t]
    \centering
    \includegraphics[width=0.85\textwidth]{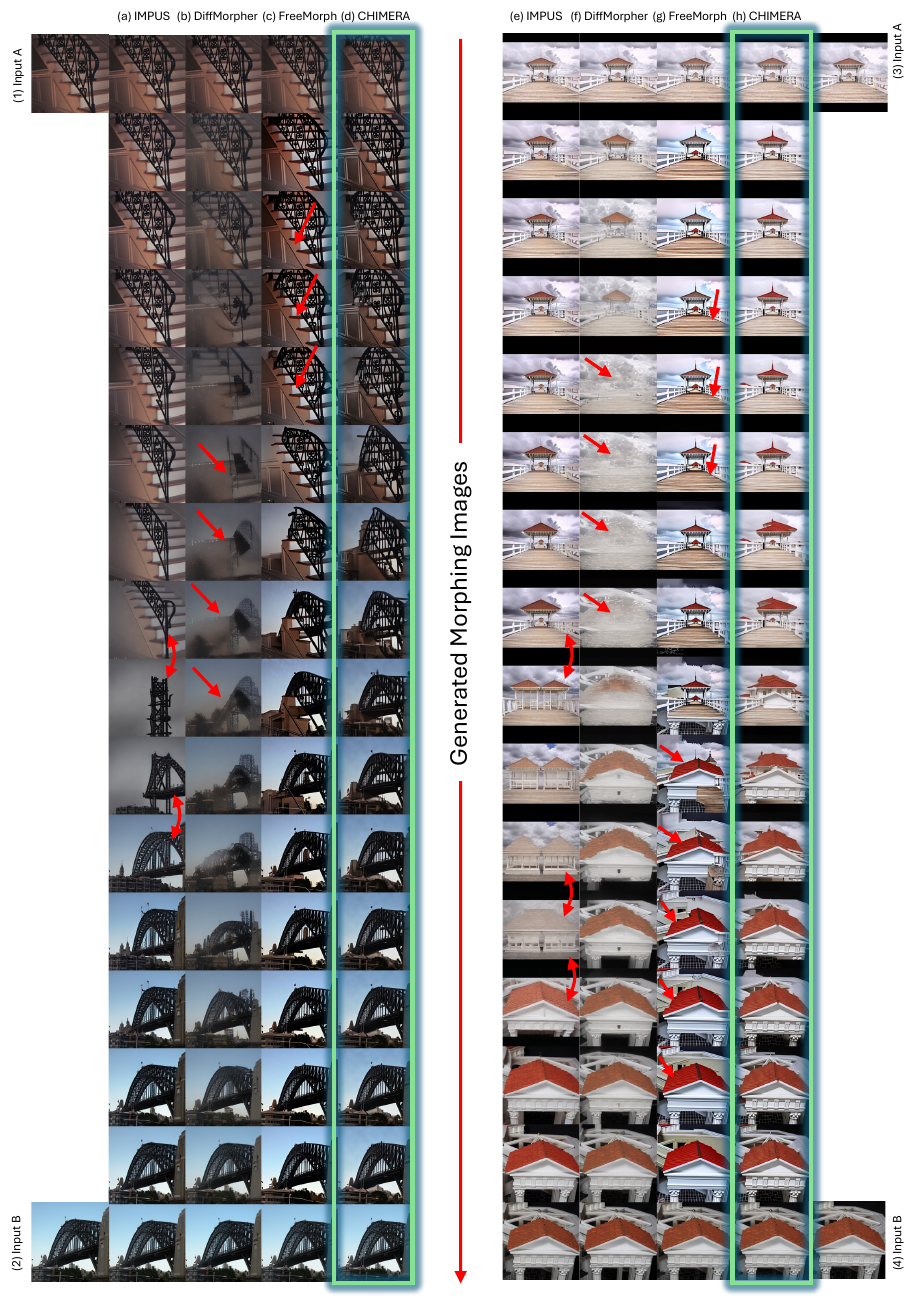}
    \caption{Second qualitative comparison showing the results of challenging 14-image morphing (compared to 5-image morphing) between input images \(A\) and \(B\). Panels (1)–(4) denote the input images, and panels (a)–(d) correspond to IMPUS, DiffMorpher, FreeMorph, and CHIMERA (Ours), respectively. The same convention applies to panels (e)–(h). Please zoom in for better visualization.}
    \label{fig:16f_qual2}
\end{figure*}

We additionally provide qualitative results for the setting with 14 morphing images in \cref{fig:16f_qual1} and \cref{fig:16f_qual2}. Similar to \cref{fig:7f_qual3} and \cref{fig:7f_qual4}, IMPUS shows transitions with insufficient smoothness, while DiffMorpher contains many frames where the structure collapses. FreeMorph also produces images with overly saturated colors. In contrast, as shown in panels (d) and (h) of \cref{fig:16f_qual1} and \cref{fig:16f_qual2}, CHIMERA consistently maintains both smooth transitions and strong domain consistency.

These qualitative results are consistent with the quantitative evaluations presented earlier. For example, IMPUS achieves high scores in $\mathrm{FID}_{\text{local}}$ and $\mathrm{FID}_{\text{global}}$, which measure domain consistency, but shows lower performance in LPIPS and PPL, which assess smoothness. Conversely, DiffMorpher performs well in terms of smoothness but exhibits lower domain consistency.

\subsection{Additional Qualitative Result on 5-Image Morphing}
\label{sp_sec:extend_qual}

\cref{fig:7f_qual3} and \cref{fig:7f_qual4} present qualitative results for the setting where five morphing images are generated between \(A\) and \(B\). As shown in panels (a) and (e) of \cref{fig:7f_qual3} and \cref{fig:7f_qual4} (red arrows), IMPUS produces images with abrupt transitions. In panels (b) and (f), the morphing images exhibit good smoothness, but the red arrows highlight collapsed structures or noticeable artifacts. In panels (c) and (g), the transitions remain smooth, yet the red arrows indicate a tendency toward excessively saturated colors. In contrast, panels (d) and (h) of \cref{fig:7f_qual3} and \cref{fig:7f_qual4} show that the proposed CHIMERA preserves both domain consistency and smoothness.

\clearpage
\section{Additional Implementation Detail}\label{sp_sec:imple_detail}

All experiments on Morph4Data~\cite{cao2025freemorph} and MorphBench~\cite{zhang2024diffmorpher} are conducted at a resolution of \(768 \times 768\). 
For SD~1.4, SD~1.5, SD~2, and SDXL~\cite{podell2023sdxl}, we use the same DDIM sampler as in the SD~2.1 setting of CHIMERA. 
For FLUX~\cite{labs2025flux}, we use the default Euler-based ODE sampler provided by the Flow Matching framework. For SD~1.4, SD~1.5, SD~2, and SDXL, we use a total of 40 timesteps, while for FLUX we follow the commonly used default setting of 50 steps. 
All configurations related to ACI and SAP remain identical to those used in the SD~2.1 setting of CHIMERA. All experiments are conducted on a single NVIDIA RTX 5090 GPU, except for FLUX, which is evaluated on a single NVIDIA RTX Pro 6000 GPU due to its larger model size.
\section{Additional Qualitative Result}
\label{sec:add-qual}

\begin{figure*}[!t]
    \centering
    \includegraphics[width=0.8\textwidth]{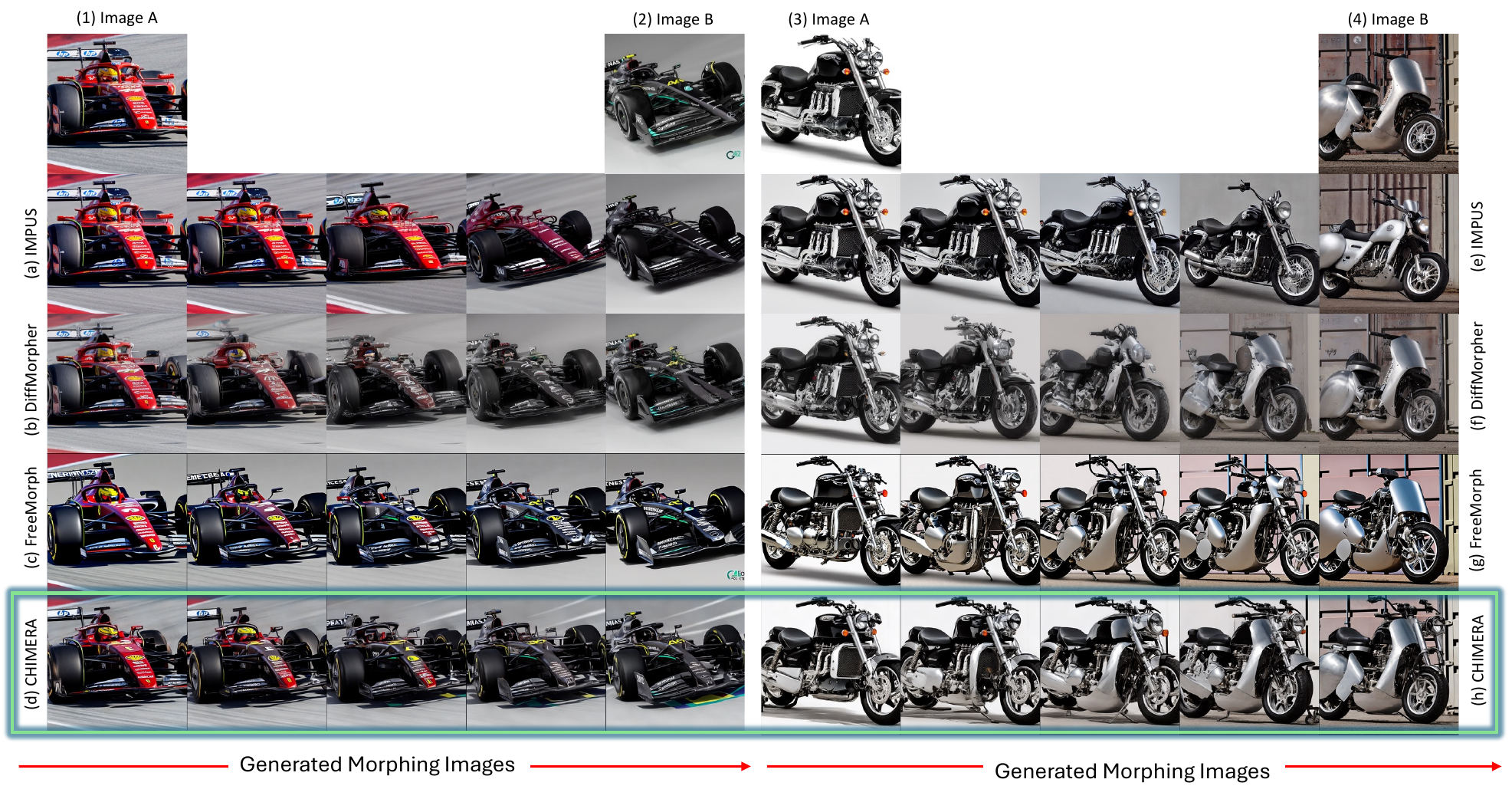}
    \caption{Additional qualitative results for 5-image morphing.}
    \label{fig:add1}
\end{figure*}

\begin{figure*}[t]
    \centering
    \includegraphics[width=0.8\textwidth]{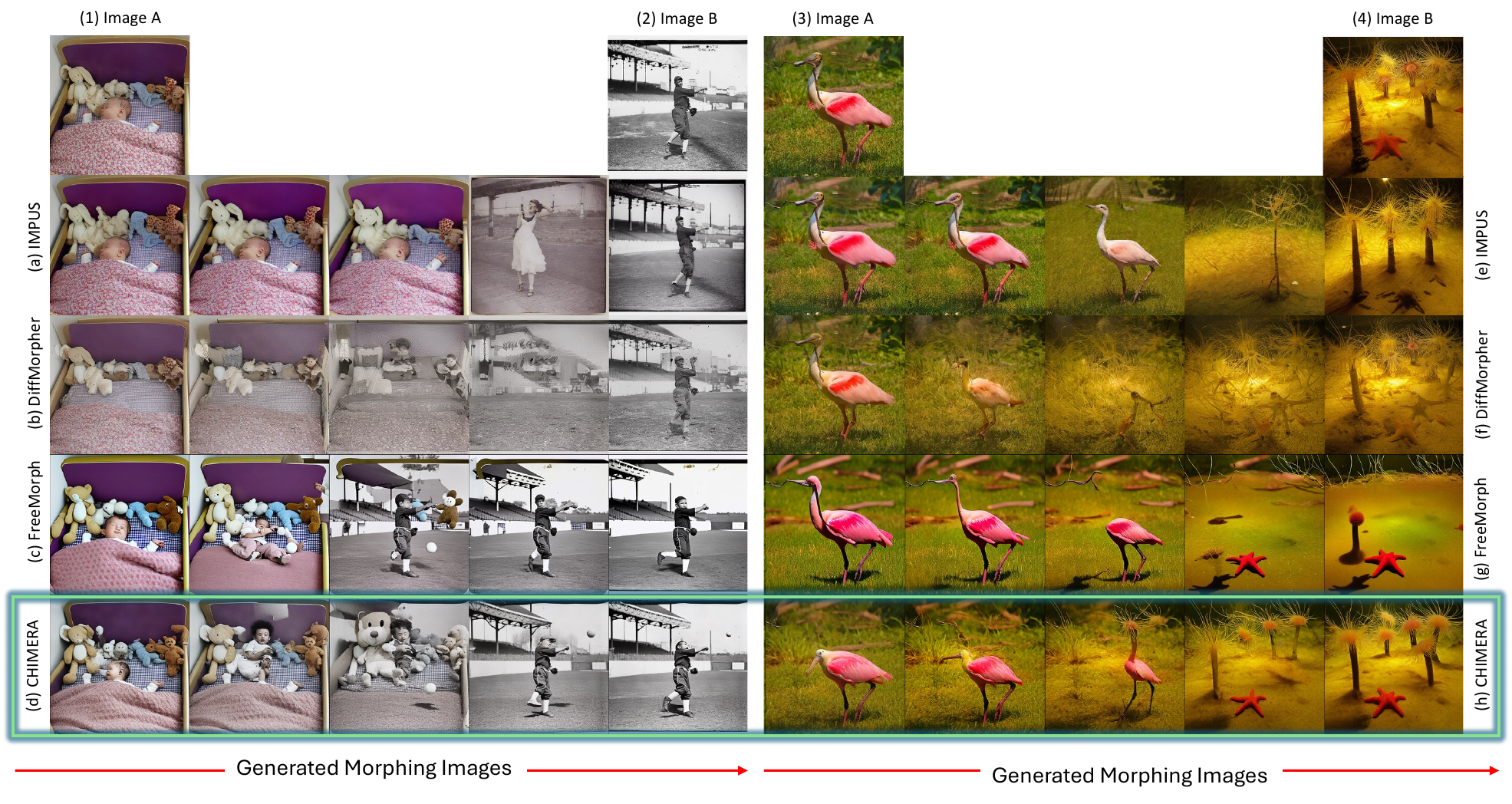}
    \caption{Additional qualitative results for 5-image morphing.}
    \label{fig:add2}
\end{figure*}

\begin{figure*}[t]
    \centering
    \includegraphics[width=1.0\textwidth]{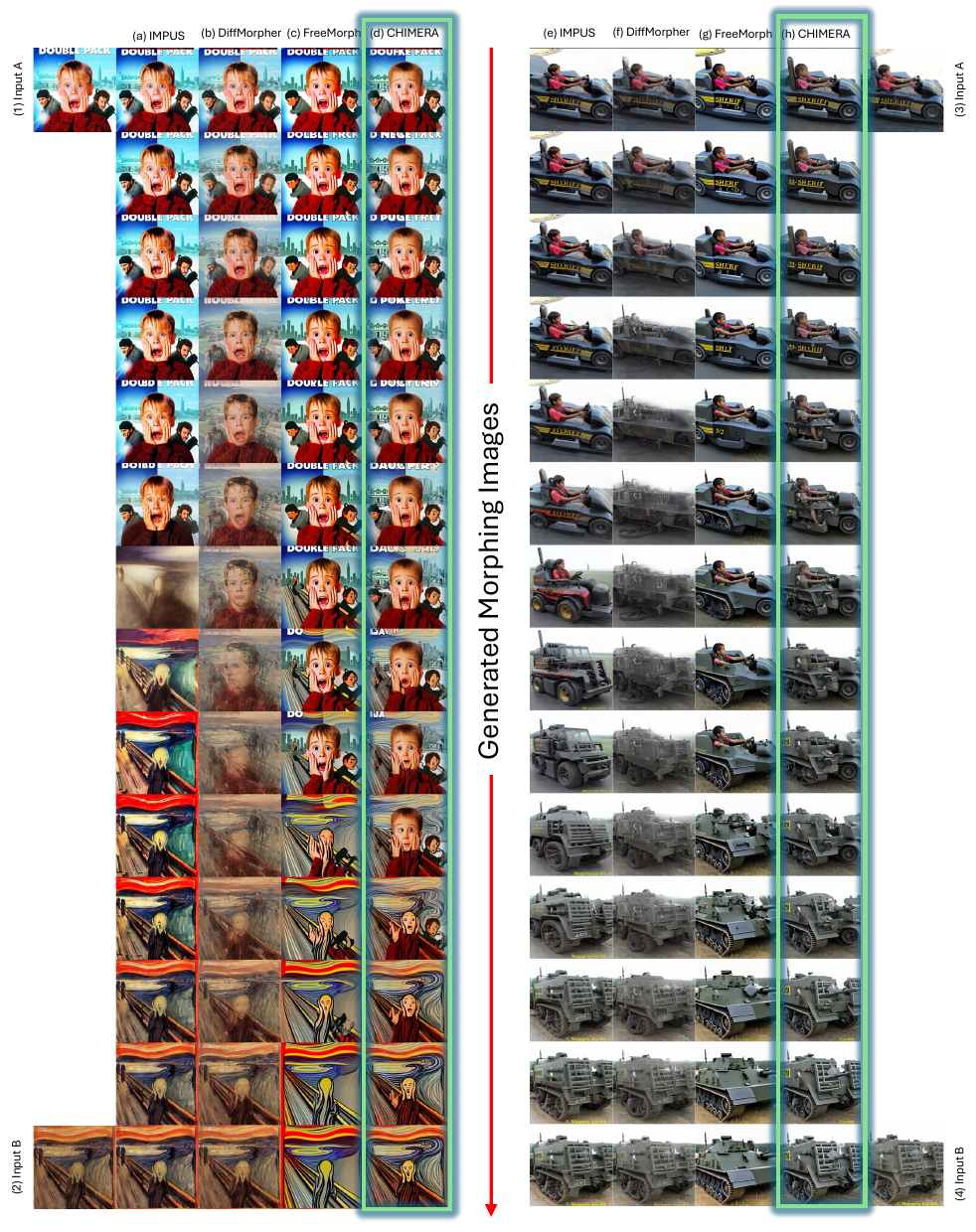}
    \caption{Additional qualitative results for 14-image morphing.}
    \label{fig:add3}
\end{figure*}

\begin{figure*}[t]
    \centering
    \includegraphics[width=1.0\textwidth]{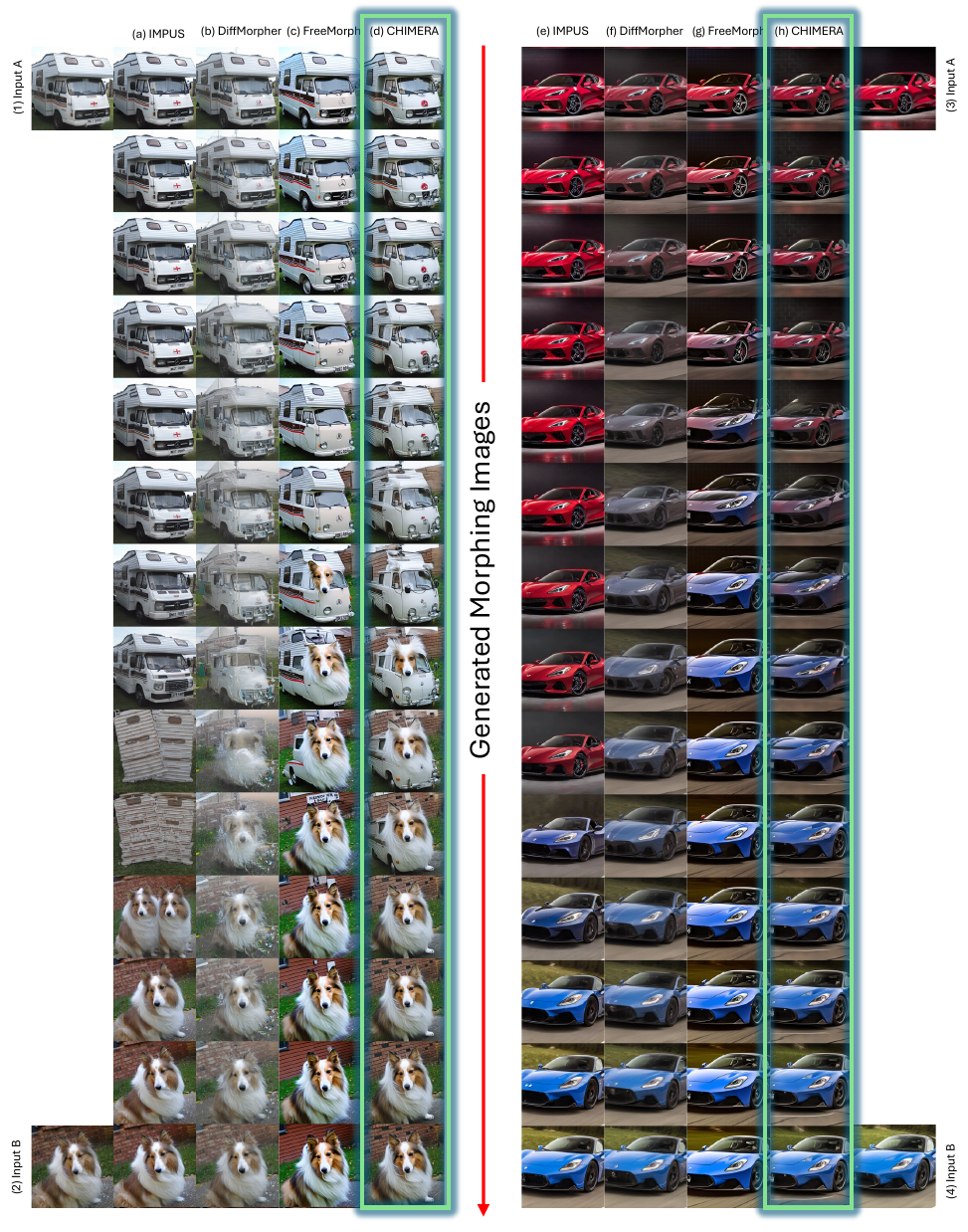} 
    \caption{Additional qualitative results for 14-image morphing.}
    \label{fig:add4}
\end{figure*}

In this section, we present additional qualitative comparisons
for the 5-image (Fig.~\ref{fig:add1} and Fig.~\ref{fig:add2}) and 14-image (Fig.~\ref{fig:add3} and Fig.~\ref{fig:add4}) morphing scenarios.

\clearpage

\begin{figure*}[t]
    \centerline{\includegraphics[width=0.85\textwidth]{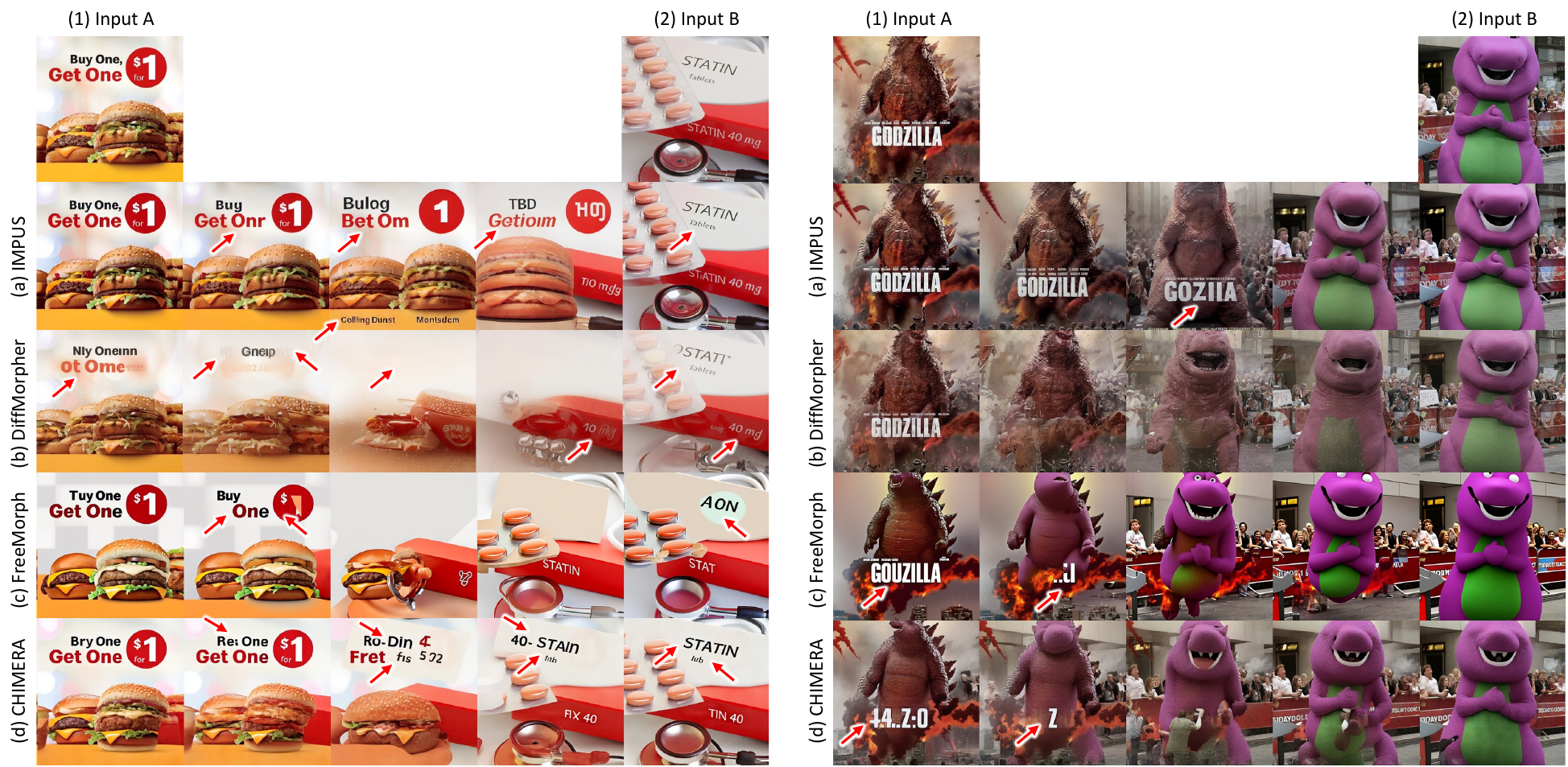}}
    \caption{
    Failure cases on images with prominent text. When the endpoint images contain different words or textual layouts, all compared methods, including CHIMERA, often produce broken or unreadable characters and occasional abrupt changes in the rendered text.
    }
    \label{fig:failure}
\end{figure*}

\section{Limitations and Failure Cases}
\label{sec:limitation}

\subsubsection{Text Rendering and Typography} Although CHIMERA produces smooth and semantically coherent transitions in general scenes, it remains limited when the endpoint images contain prominent textual elements, such as logos, signage, or dense typography (see Fig.~\ref{fig:failure}). In such cases, the generated intermediates often exhibit temporally inconsistent, partially broken, or illegible characters, even when the surrounding scene structure remains relatively coherent.
Importantly, this limitation is not unique to CHIMERA, but is consistently observed across other diffusion-based morphing methods as well~\cite{yang2024impus, cao2025freemorph, zhang2024diffmorpher}.

This failure is largely due to the inherent limitations of the underlying pre-trained diffusion backbones rather than to the morphing mechanism itself~\cite{rombach2022high, chen2023textdiffuser}.
Existing text-to-image diffusion models are well known to treat text as a high-frequency visual pattern rather than as a structured symbolic entity, and therefore often lack the fine-grained control required for accurate glyph generation~\cite{zhangli2024layout, zeng2024textctrl, gunawan2025omnitext}.
Since CHIMERA operates within such a pre-trained latent space, it inevitably inherits these weaknesses in text rendering, which explains why typographic failures persist even when the global morphing remains smooth.

\subsubsection{Future Direction: Glyph-Aware Morphing}
We view this limitation as a meaningful direction for future research.
Addressing textual inconsistency likely requires going beyond standard attention-based composition and incorporating explicit text-control mechanisms developed for recent text generation and editing frameworks, such as layout-guided generation~\cite{zhangli2024layout} or OCR-consistency objectives~\cite{chen2023textdiffuser}.
A promising next step is a \textit{glyph-aware morphing} framework that explicitly separates textual content from appearance, enabling smooth interpolation of character geometry while preserving legibility.
Extending our current attention composition strategy to better preserve glyph structure may provide an effective path toward bridging semantic morphing and precise typographic control.

\clearpage
\begin{algorithm}[t]
\scriptsize
\caption{CHIMERA}
\label{alg:chimera_updated_compact}

\begin{minipage}{0.92\linewidth}
\textbf{Input:} input image pair $A,B$; number of morphing images $K$; DDIM inversion steps $N_{\mathrm{inv}}$; denoising steps $N_{\mathrm{dng}}$; layer groups $\mathcal{S}=\{D,M,U\}$; ACI weights $\{\lambda_S\}_{S\in\mathcal{S}}$; SAP active ratio $\eta$; anchor-reliability threshold $\rho_\text{thr}$; \\
\textbf{Output:} morphing sequence $\{I_k\}_{k=0}^{K-1}$.\\[3pt]

\textcolor{Cerulean}{\textbf{Step 1: DDIM inversion and cache collection.}}\\
1:\;
\parbox[t]{0.86\linewidth}{For each $X \in \{A,B\}$, run DDIM inversion to obtain the inverted latent $z_X$ and cached multi-scale U-Net features $H_S(X,t)$ for $S \in \mathcal{S}$ and $t \in \mathcal{T}_{\mathrm{inv}}$ as in Eq.~(1).}\\[3pt]

\textcolor{Cerulean}{\textbf{Step 2: Layer- and Timestep-wise Frequency Matching (LTM).}}\\
2:\;
\parbox[t]{0.86\linewidth}{For each cached feature tensor $Z$, compute its FFT-magnitude descriptor $r(Z)=\mathrm{Pool}\!\left(\frac{1}{C}\sum_{c=1}^{C} |\mathcal{F}(Z_c)|\right)$.}\\
3:\;
\parbox[t]{0.86\linewidth}{Construct layer-group prototypes $\bar r_S$ and timestep prototypes $\bar r_t$, compute $D_{S,t}=d(\bar r_S,\bar r_t)$, and assign the matched group $S^*(t)=\arg\min_{S\in\mathcal{S}} D_{S,t}$.}\\
4:\;
\parbox[t]{0.86\linewidth}{Re-organize cached features using the matched group, i.e., use $H_{S^*}(X,t)$ for subsequent interpolation and injection.}\\[3pt]

\textcolor{Cerulean}{\textbf{Step 3: Morphing latent construction and cache interpolation.}}\\
5:\;
\parbox[t]{0.86\linewidth}{For $k=0,\dots,K-1$, compute $\alpha_k$ and construct the morphing latent $z_k=\mathrm{slerp}(z_A,z_B;\alpha_k)$.}\\
6:\;
\parbox[t]{0.86\linewidth}{For each $k$ and $t \in \mathcal{T}_{\mathrm{inv}}$, construct the interpolated cache $\widehat{C}_{S^*}(k,t)=\mathrm{slerp}\!\bigl(H_{S^*}(A,t),H_{S^*}(B,t);\alpha_k\bigr)$ as in Eq.~(2).}\\[3pt]

\textcolor{Cerulean}{\textbf{Step 4: Anchor-correlated prompt triplet construction for SAP.}}\\
7:\;
\parbox[t]{0.86\linewidth}{Initialize $\rho \leftarrow 0$.}\\
8:\;
\parbox[t]{0.86\linewidth}{\textbf{while} $\rho < \rho_\text{thr}$ \textbf{do}}\\
9:\;\;\;
\parbox[t]{0.86\linewidth}{Query the VLM with $(A,B)$ to obtain $(text_{\mathrm{anc}}, text_A, text_B)$, and encode them into $(e_{\mathrm{anc}}, e_A, e_B)$.}\\
10:\;\;\;
\parbox[t]{0.86\linewidth}{Compute anchor reliability $\rho=\min\bigl(\mathrm{cossim}(e_{\mathrm{anc}},e_A), \mathrm{cossim}(e_{\mathrm{anc}},e_B)\bigr)$.}\\
11:\;
\parbox[t]{0.86\linewidth}{\textbf{end while}}\\[3pt]

\textcolor{Cerulean}{\textbf{Step 5: Denoising with IDM, ACI, and SAP.}}\\
12:\;
\parbox[t]{0.86\linewidth}{For each morphing index $k=0,\dots,K-1$, initialize $x_{\tau_0}^{(k)} \leftarrow z_k$.}\\
13:\;
\parbox[t]{0.86\linewidth}{\textbf{for} each denoising timestep $\tau \in \mathcal{T}_{\mathrm{dng}}$ \textbf{do}}\\
14:\;\;\;
\parbox[t]{0.86\linewidth}{Map $\tau$ to the inversion timestep $t \leftarrow \phi(\tau)$ via IDM, and run the diffusion U-Net on $x_{\tau}^{(k)}$ to obtain $\{F_S^{(\tau)}\}_{S\in\mathcal{S}}$.}\\
15:\;\;\;
\parbox[t]{0.86\linewidth}{For each $S \in \mathcal{S}$, retrieve $\widehat{C}_{S^*}(k,\phi(\tau))$ and compute the ACI feature $\widetilde{F}_S^{(\tau)} = F_S^{(\tau)} + \lambda_S \cdot \widehat{C}_{S^*}(k,\phi(\tau))$.}\\
16:\;\;\;
\parbox[t]{0.86\linewidth}{\textbf{if} $\tau \in \mathcal{T}_{\mathrm{dng}}^{\mathrm{early}}(\eta)$ \textbf{then} apply SAP-guided cross-attention as in Eq.~(4); \textbf{else} use the original cross-attention.}\\
17:\;\;\;
\parbox[t]{0.86\linewidth}{Update the latent by one denoising step to obtain $x_{\tau+1}^{(k)}$.}\\
18:\;
\parbox[t]{0.86\linewidth}{\textbf{end for}}\\
19:\;
\parbox[t]{0.86\linewidth}{Decode $x_{\tau_{\mathrm{final}}}^{(k)}$ with the VAE decoder to obtain $I_k = \mathrm{VAE}^{-1}(x_{\tau_{\mathrm{final}}}^{(k)})$.}\\
20:\;
\parbox[t]{0.86\linewidth}{\textbf{end for}}\\[3pt]

\textcolor{Cerulean}{\textbf{Step 6: Return.}}\\
21:\;
\parbox[t]{0.86\linewidth}{\textbf{return} morphing sequence $\{I_k\}_{k=0}^{K-1}$.}
\end{minipage}
\end{algorithm}
\begin{algorithm}[t]
\scriptsize
\caption{Global--Local Consistency Score (GLCS)}
\label{alg:glcs}
\begin{minipage}{0.92\linewidth}
\textbf{Input:} endpoint images $A,B$; morphing images $\{I_k\}_{k=1}^{K}$; DiffSim-based bounded similarity $s(\cdot,\cdot)\in[-1,1]$.\\
\textbf{Output:} Global Consistency Score $\mathrm{GCS}$, Local Consistency Score $\mathrm{LCS}$, and Global--Local Consistency Score $\mathrm{GLCS}$.\\[3pt]

\textcolor{Cerulean}{\textbf{Step 1: Similarity computation.}}\\
1:\;%
\parbox[t]{0.86\linewidth}{\textbf{for} each endpoint $X \in \{A,B\}$ \textbf{do}}\\
2:\;\;\;%
\parbox[t]{0.86\linewidth}{\textbf{for} $k = 1,\dots,K$ \textbf{do} compute the per-frame similarity $s_X(k)=s(X,I_k)$.}\\
3:\;\;\;%
\parbox[t]{0.86\linewidth}{\textbf{end for}}\\
4:\;%
\parbox[t]{0.86\linewidth}{\textbf{end for}}\\
5:\;%
\parbox[t]{0.86\linewidth}{Compute the four endpoint similarities $s(A,A)$, $s(A,B)$, $s(B,A)$, and $s(B,B)$.}\\[3pt]

\textcolor{Cerulean}{\textbf{Step 2: Global Consistency Score (GCS).}}\\
6:\;%
\parbox[t]{0.86\linewidth}{\textbf{for} $k = 1,\dots,K$ \textbf{do}}\\
7:\;\;\;%
\parbox[t]{0.86\linewidth}{Compute the normalized interpolation ratio $\alpha_k=\frac{k}{K+1}$.}\\
8:\;\;\;%
\parbox[t]{0.86\linewidth}{Estimate the globally expected similarities by spherical interpolation in similarity space:
$\bar{s}_A(k)=\operatorname{slerp}(s(A,A),s(A,B);\alpha_k)$ and
$\bar{s}_B(k)=\operatorname{slerp}(s(B,A),s(B,B);\alpha_k)$.}\\
9:\;\;\;%
\parbox[t]{0.86\linewidth}{Compute the per-frame global consistency
$
g_k=
[1-|s_A(k)-\bar{s}_A(k)|]_0^1
\cdot
[1-|s_B(k)-\bar{s}_B(k)|]_0^1,
$
where $[x]_0^1=\min(1,\max(0,x))$.}\\
10:\;%
\parbox[t]{0.86\linewidth}{\textbf{end for}}\\
11:\;%
\parbox[t]{0.86\linewidth}{Aggregate all sharpened terms:
$\mathrm{GCS}=\frac{1}{K}\sum_{k=1}^{K} g_k$.}\\[3pt]

\textcolor{Cerulean}{\textbf{Step 3: Local Consistency Score (LCS).}}\\
12:\;%
\parbox[t]{0.86\linewidth}{\textbf{for} $k = 1,\dots,K$ \textbf{do}}\\
13:\;\;\;%
\parbox[t]{0.86\linewidth}{For each endpoint $X\in\{A,B\}$, estimate the locally expected similarity $\tilde s_X(k)$ from neighboring frames: use the single adjacent frame at the boundaries and the average of the previous and next frames otherwise.}\\
14:\;\;\;%
\parbox[t]{0.86\linewidth}{Compute the per-frame local consistency
$
\ell_k=
[1-|s_A(k)-\tilde{s}_A(k)|]_0^1
\cdot
[1-|s_B(k)-\tilde{s}_B(k)|]_0^1.
$
}\\
15:\;%
\parbox[t]{0.86\linewidth}{\textbf{end for}}\\
16:\;%
\parbox[t]{0.86\linewidth}{Aggregate all local consistency terms:
$\mathrm{LCS}=\frac{1}{K}\sum_{k=1}^{K}\ell_k$.}\\[3pt]

\textcolor{Cerulean}{\textbf{Step 4: Global--Local Consistency Score (GLCS).}}\\
17:\;%
\parbox[t]{0.86\linewidth}{Combine the two scores by geometric mean:
$\mathrm{GLCS}=\sqrt{\mathrm{GCS}\cdot\mathrm{LCS}}$.}\\[3pt]

18:\;%
\parbox[t]{0.86\linewidth}{\textbf{return} $\mathrm{GCS}, \mathrm{LCS}, \mathrm{GLCS}$.}
\end{minipage}
\end{algorithm}

\clearpage



\end{document}